\documentclass{article}

    \usepackage[nonatbib, final]{neurips_data_2022}

\usepackage[utf8]{inputenc} 
\usepackage[T1]{fontenc}    
\usepackage[pagebackref=true,breaklinks=true,colorlinks,bookmarks=false]{hyperref}       
\usepackage{url}            
\usepackage{booktabs}       
\usepackage{amsfonts}       
\usepackage{nicefrac}       
\usepackage{microtype}      
\usepackage[dvipsnames]{xcolor}         

\usepackage{graphicx}
\usepackage{verbatim}         
\usepackage{colortbl}       

\usepackage{fdsymbol}

\usepackage{tablefootnote}
\usepackage{nameref}
\usepackage{enumitem}

\usepackage{multirow}
\newcommand{\parnobf}[1]{{\vspace{-0.5mm} \noindent \textbf{{#1}.}}}
\newcommand{\parnobfq}[1]{{\vspace{-0.2mm} \noindent \textbf{{#1}?}}}
\newcommand{\parnoit}[1]{{\vspace{0.0mm} \noindent \textit{{#1}.}}}

\newcommand{\DS}[1]{{\color{orange}{Dandan: #1}}}
\newcommand{\SF}[1]{{\color{magenta}{[Sanja: #1]}}}
\newcommand{\ToDo}[2]{{\noindent \textcolor{red}{\textbf{#1}}: \textcolor{red}{#2}}}
\newcommand{\note}[1]{{\color{red}{{\bf #1}}}}
\renewcommand{\note}[1]{}
\renewcommand{\ToDo}[2]{}
\renewcommand{\SF}[1]{}
\definecolor{evisorPurple}{HTML}{622D87}
\definecolor{greensuccess}{HTML}{97C40B}
\definecolor{redfailure}{HTML}{F5511D}

\title{EPIC-KITCHENS VISOR Benchmark \\VIdeo Segmentations and Object Relations}

\author{
Ahmad Darkhalil$^{\bigstar\clubsuit}$ $\quad$ Dandan Shan$^{\bigstar\spadesuit}$ $\quad$ Bin Zhu$^{\bigstar\clubsuit}$ $\quad$ Jian Ma$^{\bigstar\clubsuit}$\\ \textbf{Amlan Kar$^\vardiamondsuit$ $\quad$ Richard Higgins$^\spadesuit$ $\quad$ Sanja Fidler$^\vardiamondsuit$ $\quad$ David Fouhey$^\spadesuit$ $\quad$ Dima Damen$^\clubsuit$}\\\\
\noindent $^{\clubsuit}$Uni. of Bristol, UK $\quad^{\spadesuit}$Uni. of Michigan, US $\quad^{\vardiamondsuit}$Uni. of Toronto, CA  $\quad$ $^\bigstar$: Co-First Authors
}

\begin{document}

\maketitle

\begin{abstract}
We introduce VISOR, a new dataset of  pixel annotations and a benchmark suite for segmenting hands and active objects in egocentric video. 
VISOR annotates videos from EPIC-KITCHENS, which comes with a new set of challenges not encountered in current video segmentation datasets. Specifically, we need to ensure both short- and long-term consistency of pixel-level annotations as objects undergo transformative interactions, e.g. an onion is peeled, diced and cooked - where we aim to obtain accurate pixel-level annotations of the peel, onion pieces, chopping board, knife, pan, as well as the acting hands. 
VISOR introduces an annotation pipeline, AI-powered in parts, for scalability and quality. 
In total, we publicly release 272K manual semantic masks of 257 object classes, 9.9M interpolated dense masks, 67K hand-object relations, covering 36 hours of 179 untrimmed videos. 
Along with the annotations, we introduce three challenges in video object segmentation, interaction understanding and long-term reasoning.

For data, code and leaderboards: \url{http://epic-kitchens.github.io/VISOR}
\end{abstract}

\section{Introduction}
\label{sec:introduction}

\begin{figure*}
\centering
\includegraphics[width=\linewidth]{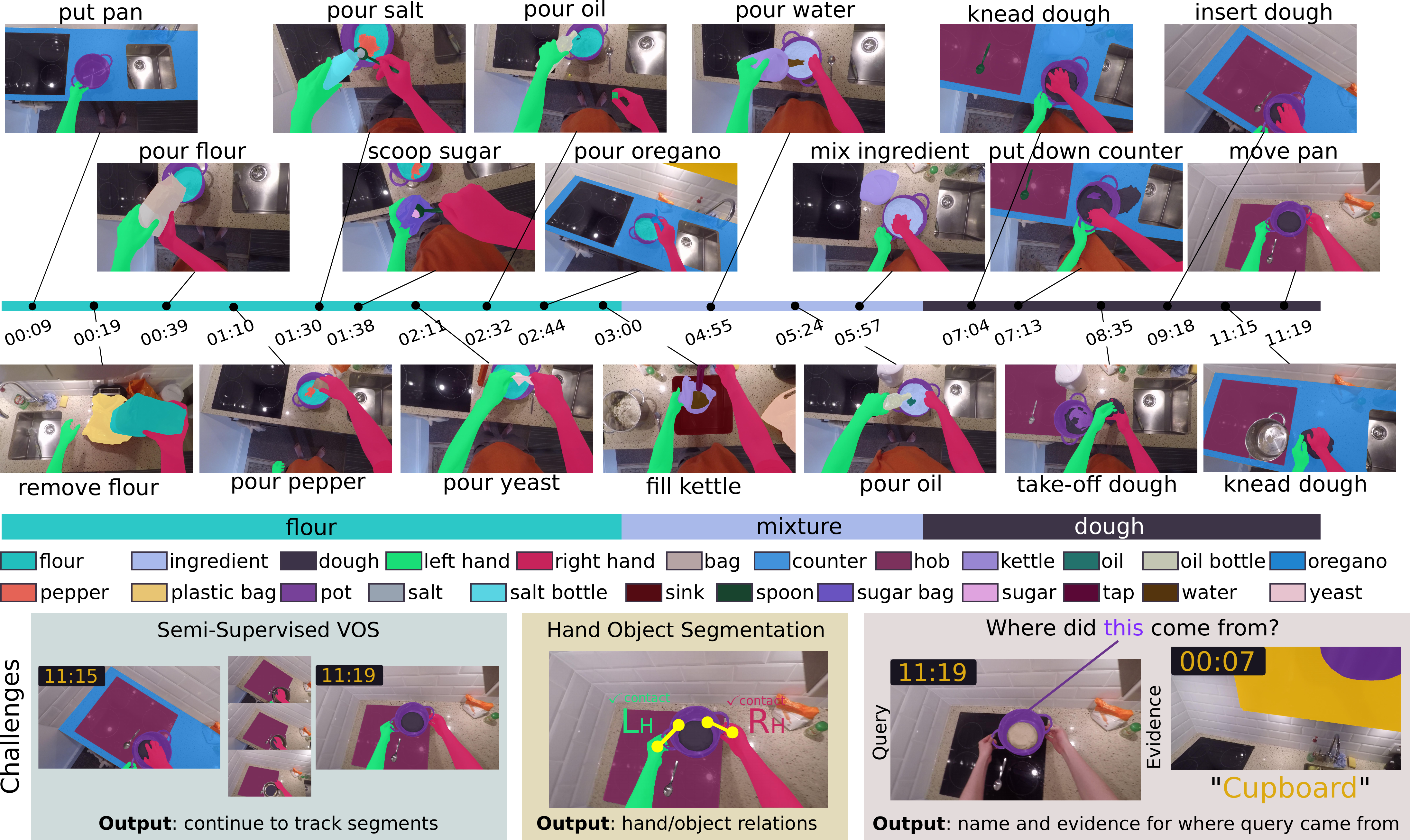} \caption{{\bf VISOR Annotations and Benchmarks.} 
Sparse segmentations from one video ({\tt P06\_03}), where flour turns into dough over the course of over 11 minutes. The colours of the timeline represent the stages: flour$\rightarrow$mixture$\rightarrow$dough. 
By building off of the EPIC-KITCHENS dataset, our segments join a rich set of action labels. 
We use the annotations to define three challenges at different timescales: (i) Semi-Supervised VOS over several actions; (ii) Hand$\leftrightarrow$Object Segmentation and contact relations in the moment; and (iii) \emph{where did \textcolor{evisorPurple}{{\bf \emph{this}}} come from?} for long-term reasoning. }%
\label{fig:long-term}
\end{figure*}

Consider a video capturing the tedious process of preparing bread, from obtaining, measuring and mixing ingredients to kneading and shaping dough. 
Despite being a routine task, the discrete nature of computer vision models, trained mostly from images, expects to recognise objects as either \emph{flour} or \emph{dough}.
Capturing the transformation through pixel-level annotations has not been attempted to date.
Building on top of EPIC-KITCHENS egocentric videos~\cite{Damen2018EPICKITCHENS},  VISOR utilises action labels and provides sparse segmentations, of hands and active objects, with a rate of annotations so as to represent both short (e.g. `add salt') and long (`knead dough') temporal actions.
In Fig.~\ref{fig:long-term}, we introduce sample annotations from one video of VISOR.

Rich pixel-level annotations have transformed image understanding~\cite{gupta2019lvis,lin2014microsoft} and autonomous driving~\cite{Cordts2016Cityscapes,geiger2012we}, amongst other tasks.
By augmenting action labels with semantic segmentations, we hope to enable complex video understanding capabilities. Previous efforts to incorporate spatio-temporal annotations and object relations~\cite{Ego4D2022CVPR,gu2018ava,ji2020action,li2020ava} have only featured bounding boxes. A few seminal video datasets~\cite{Perazzi2016,wang2021unidentified,xu2018youtube} provide pixel labels over time, but are often short-term, and are not associated with fine-grained action labels.
We compare VISOR to other datasets in \S\ref{sec:related}. 

We obtain VISOR annotations via a new pipeline consisting of three stages that we describe in \S\ref{sec:collection}: (i)~identifying {\it active objects} that are of relevance to the current action, as well as their semantic label; (ii) annotating pixel-level segments with an AI-powered interactive interface iterated with manual quality-control checks; and (iii) relating objects spatially and temporally for short-term consistency and hand-object relations. 

In total, we  obtain annotations of 271.6K masks across 50.7K images of 36 hours, interpolated to 9.9M dense masks, which we analyse in \S\ref{sec:statistics}. 
Moreover, these new annotations directly plug in with a strong existing ecosystem of annotations, from action labels to closed-vocabulary of objects.

In \S\ref{sec:challenges}, we use these new annotations to define three benchmarks, depicted in Fig.~\ref{fig:long-term} (bottom). Our first task, {\it Semi-Supervised Video Object Segmentation (VOS)}, requires tracking object segments across {\it consecutive actions}. We extend previous VOS efforts~\cite{Perazzi2016,wang2021unidentified,xu2018youtube} by tackling 2.5-4x longer sequences. Our second task, {\it Hand Object Segmentation (HOS)}, involves predicting the contact between hands and the active objects, and extends past efforts by predict both the relation and accurate segmentations of hands and objects. 
Our final task, {\it Where Did This Come From (WDTCF)?}, is a taster benchmark for long-term perception with scene segmentation, where one traces back a given segment to the container it came from (e.g., milk from the fridge, and a plate from a particular cupboard) and identifies the frame(s) containing evidence of the query object emerging from the container.

\definecolor{dred}{RGB}{242, 220, 219}
\definecolor{dblue}{RGB}{220, 230, 242}
\definecolor{dpurple}{RGB}{235, 212, 225}

\newcolumntype{m}{>{\columncolor{dred}}r}
\newcolumntype{t}{>{\columncolor{dblue}}r}
\newcolumntype{s}{>{\columncolor{dpurple}}r}

\newcolumntype{x}{>{\columncolor{dred}}c}
\newcolumntype{y}{>{\columncolor{dblue}}c}
\newcolumntype{z}{>{\columncolor{dpurple}}c}

\section{Related Efforts}
\label{sec:related}
We compare VISOR with  previous efforts that have {\it both} pixel-level and action annotations in Table~\ref{tab:comparative}.
We divide our comparison into three sections: {\it basic statistics} on the videos themselves, {\it pixel-level annotation statistics} on images annotated, masks and average masks per image and {\it action annotations} where entities or action classes have been labelled.
Our videos are significantly longer. On average, VISOR videos are 12 {\it minutes} long. 
To make the challenge feasible to current state-of-the-art models, we divide our videos into shorter sequences that are 12 {\it seconds} long on average.
VISOR is also the only dataset that combines pixel-level segmentations with both action and entity class labels.

The closest effort to VISOR is that of the seminal UVO work~\cite{wang2021unidentified} that annotates videos from the Kinetics dataset~\cite{carreira2017quo}. 
As every video in Kinetics captures a single action, UVO only offers short-term object segmentations, focusing on translated objects and occlusions.
In contrast, VISOR annotates videos from EPIC-KITCHENS~\cite{Damen2020RESCALING} that capture sequences of 100s of fine-grained actions.
VISOR segmentations thus capture long-term object segmentations of the same instance undergoing a series of transformations – the same potato is seen being picked from the fridge, washed, peeled, cut, and cooked. 
Object transformations are also significantly more varied, beyond translation and deformation, offering a challenge previously unexplored.
Ensuring long-term temporal consistency in VISOR requires a more complex framework for relating objects over sequences and through semantic classes.
In doing so, we take inspiration from UVO~\cite{wang2021unidentified} in how sparse segmentations were used with bi-directional interpolations to obtain dense annotations (See appendix~\ref{sec:dense}).

VISOR is also related to efforts that capture hand-object relations, including for identifying active objects~\cite{shan2020understanding,Baradel_2018_ECCV}, next active objects~\cite{ragusa2022meccano,Liu_2022_CVPR}, object articulation~\cite{xu2021d3dhoi} and grasping~\cite{Hasson19,Taheri2020,Taheri_2022_CVPR,Turpin2022}. 
Distinct from these works, we provide the first pipeline, annotations and benchmark suite for pixel-level segmentations.

\begin{table}[t]
    \centering
    \caption{{\bf Comparison with Current Data.} Compared to past efforts featuring both pixel- and action-level annotations, VISOR features longer sequences and provides the largest number of manually annotated masks on diverse actions and objects. *:Stats from train/val public annotations. $^\dagger$Avg VISOR video is {\bf 12 \emph{minutes}} long. We divide these into sub-sequences w/ consistent entities (See \S\ref{sec:collection}). Masks are semantically consistent, through class knowledge, across all videos. }
    \resizebox{\textwidth}{!}{%
    \begin{tabular}{lmmmtttsss} \toprule
          & \multicolumn{3}{x}{Basic Statistics} &\multicolumn{3}{y}{Pixel-Level Annotations} &\multicolumn{3}{z}{Action Annotations} \\
          &     & Total & Avg & Total & Total & Avg Masks & & \#Action & \#Entity  \\
          Dataset & Year & Mins & Seq Ln & Masks & Images & per Image  & Actions & Classes & Classes \\ \midrule
          EgoHand~\cite{Bambach15} &2015 &72 & - &15.1K &4.8K &3.2 &- &- & 2 \\
          DAVIS~\cite{Caelles_arXiv_2019} &2019 &8 & 3s &32.0K &10.7K &3.0 &- &- &-\\
          YTVOS~\cite{xu2018youtube} &2018 & 335 & 5s &197.2K &\bf ~120.4K &1.6&- &- &94\\
          UVOv0.5 (Sparse)~\cite{wang2021unidentified} &2021& 511 & 3s &*200.6K &30.5K &\bf*8.8 &10,213 &300 &-\\
          VISOR (Ours) &2022 & \bf 2,180 & \bf 12s$^\dagger$ & \bf 271.6K  &50.7K &5.3 &\bf 27,961 & \bf 2,594 & \bf 257\\ \bottomrule
    \end{tabular}}
\vspace*{-10pt}
    \label{tab:comparative}
\end{table}

\section{VISOR annotation pipeline}
\label{sec:collection}

Our annotation pipeline consists of multiple stages. In the first stage, we identify the frames and {\it active} entities to be annotated (\S\ref{sec:annotation_entities}). Having identified {\it what} should be segmented, we obtain pixel-level annotations, substantially accelerated by an AI-powered tool. To achieve consistency, we subject annotators to extensive training, and employ manual verification (\S\ref{sec:annotation_annotators}). Finally, we collect additional annotations that are needed for our challenges (\S\ref{sec:annotation_relations}). We next describe our pipeline in detail.

\parnobf{Background: EPIC-KITCHENS Dataset and Annotations} We build on the large-scale egocentric \href{https://epic-kitchens.github.io/}{EPIC-KITCHENS-100} dataset~\cite{Damen2020RESCALING}, collected with University of Bristol faculty ethics approval and signed consent forms of participants (anonymous). 
Participants wear a head-mounted camera, start recording before they enter the kitchen and only stop recording when they leave the room. These videos have fine-grained action labels comprising: (i) video file (e.g., {\tt P06\_108}); (ii) start and end times of an action (e.g., from {\tt 03:52.6} to {\tt 03:58.4}); (iii) short open-vocabulary narration describing the action (e.g., `take mozzarella out of packet'); (iv) closed-vocabulary of \emph{verb} and \emph{noun} classes (e.g.~mapping `mozzarella' to the class `cheese'). Table 2 of~\cite{Damen2020RESCALING} compares EPIC-KITCHENS to publicly available action recognition datasets in size, annotations and classes (e.g., ~\cite{carreira2017quo,gu2018ava,sigurdsson2016hollywood}). 
\begin{figure*}[t]
    \centering
    \includegraphics[width=1.0\textwidth]{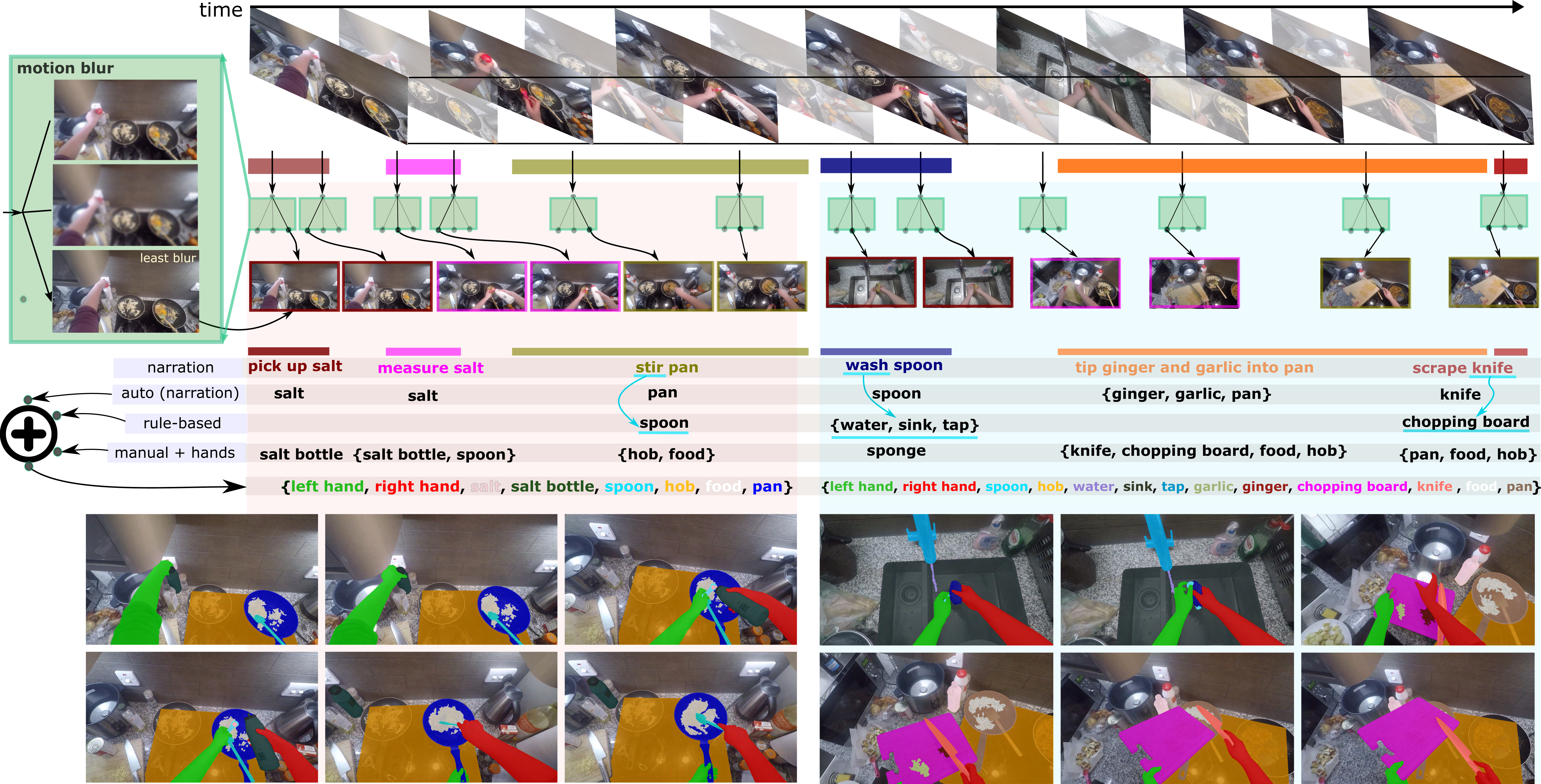}
    \caption{{\bf Part of the VISOR Annotation Pipeline.} We divide untrimmed videos to subsequences of 3 consecutive actions, sampling 6 frames/seq. We find entities to annotate via the union of automatic parsing of narrations, manual rules, and human annotations. These are then manually segmented.}
    \vspace*{-8pt}
    \label{fig:VISOR_pipeline}
\end{figure*}

\subsection{Entities, frames and sub-sequences}
\label{sec:annotation_entities}

Our first stage aims to identify what {\it data} and {\it entities} ought to be annotated. Our data selection aims to find clear frames to annotate and distribute annotation uniformly across actions rather than time, to cover actions of varying length. Entity selection aims to find all active objects, relevant to the action, including tools, crockery and appliances.
This step is needed since the narrations do not include objects that are implied by common-sense, and are different per sample: e.g., one example of the action `peel carrot' might involve a peeler and a bowl while another might use a knife and chopping board. 
Note that as videos are captured in people's homes, frames can be highly cluttered with background/irrelevant objects we do not wish to segment.

We show the pipeline for the selection of frames and entities in Fig.~\ref{fig:VISOR_pipeline}. We divide the untrimmed videos into sequences comprising 3 non-overlapping actions. This achieves a mean of 12 seconds of temporally consistent annotations. We chose 3 actions so as to introduce slight but reasonable challenge to previous video segmentation datasets, increasing sequence lengths from 3-5s to 12s (see Table~\ref{tab:comparative}). We annotate 6 frames per sequence, avoiding blurred frames when possible for better object boundaries. 
We candidate active entities to annotate from parsing nouns in narrations, correlations with actions from parsed verbs (e.g., `cut' and `knife') and manual annotations. Annotators select the entities that are present per frame from this over-complete list. A showcase of the dataset's diversity in actions and entities is visualised in Fig.~\ref{fig:images} and more details are in the appendix.

\vspace{-2mm}
\subsection{Tooling, annotation rules and annotator training}
\label{sec:annotation_annotators}

During the project's 22 months, we worked closely with 13 annotators: training, answering questions, and revising their annotations. A more complete description of the annotation pipeline and interface, including cross-annotator agreement, is included in the appendix.

\parnobf{AI-Powered Segmentation Tooling}
To reduce the cost of pixel-level annotation, we use TORonto Annotation Suite (TORAS)~\cite{torontoannotsuite}, which uses interactive tools built from variants of~\cite{CurveGCN2019,DELSE2019}. The interface combines a mix of manual and AI-enhanced tools that we found to dramatically improve the speed of obtaining accurate annotations. The AI models were trained on generic segmentation datasets like ADE-20K~\cite{zhou2017ade20k}. We found they generalised well to our domain as shown in~\cite{polyrnnpp,polyrnn,CurveGCN2019}.

\parnobf{Annotator Recruiting, Onboarding, and Rules} We hired freelance annotators via Upwork. They received an overview of the project, training sessions with the tool, and a glossary of objects. We devised rules to facilitate consistent segmentations across frames for containers, packaging, transparency and occlusion, detailed in Appendix.
To give an example, first image in Fig~\ref{fig:images} shows our `container' rule where other inactive objects in the cupboard are not segmented out, to show the full extent of the cupboard.
Each video was annotated by one annotator to maximise temporal consistency. 
In addition to actively resolving annotator confusions, each video was manually verified. We created an interface for fast correction of annotations, focusing on temporal consistency.

\parnobf{Splits} In Table~\ref{tab:dataset_splits} we showcase the Train/Val/Test splits. 
We provide a train/val publicly-accessible split along with a held-out test set of 21  videos ($\approx$ 20\% of the frames and the masks). 
Note that the Test split aligns with the EPIC-KITCHENS test split, which is used for the action recognition challenge. 
For both Val and Test, we keep some kitchens unseen to test generality to new environments.
The Train/Test annotations are available \href{https://doi.org/10.5523/bris.2v6cgv1x04ol22qp9rm9x2j6a7}{here}. Evaluating on the test set is through the challenge leaderboards - see \href{https://epic-kitchens.github.io}{the EPIC-KITCHENS challenges} for details.

\begin{figure*}[t]
    \centering
    \includegraphics[width=1.0\textwidth]{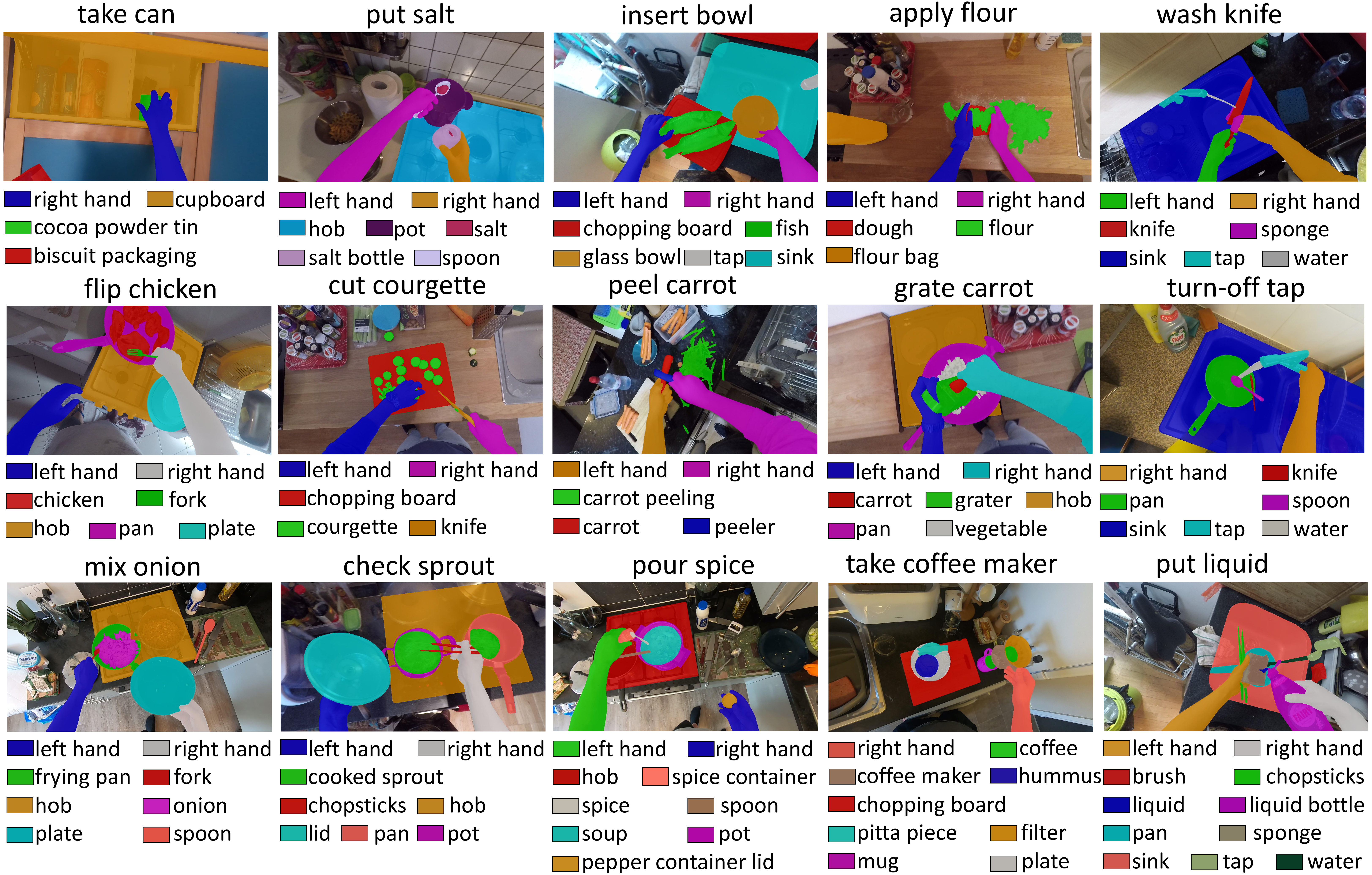}
    \vspace*{-12pt}
    \caption{{\bf Segmentations from VISOR} with entity names (legend) and ongoing action (above).}
    \vspace{-10pt}
    \label{fig:images}
\end{figure*}

\begin{table}[t]
\centering 
\caption{{\bf VISOR splits statistics}. Details of VISOR splits. 
}
\label{tab:dataset_splits}
\resizebox{\textwidth}{!}{
\begin{tabular}{llllll}
\toprule
VISOR splits & Train &Val & Train+Val& Test & Total    \\
\midrule
\# Kitchens & 33& 24 (5 unseen)  & 38& 13 (4 unseen) & 42\\
\# Untrimmed Videos & 115 &43     & 158          & 21 & 179  \\
\# Images (and \%) & 32,857 (64.8\%) &7,747 (15.3\%)     & 40,604 (80.0\%)          & 10,125 (20.0\%)  & 50,729  \\
\# Masks (and \%) & 174,426 (64.2\%) &41,559 (15.3\%)     & 215,985 (79.5\%)          & 55,599 (20.5\%)  & 271,584  \\
\# Entity Classes & 242 &182 (9 zero-shot)     & 251          & 160 (6 zero-shot)  & 257  \\
 \bottomrule
\end{tabular}}
\vspace*{-20pt}
\end{table}

\subsection{VISOR Object Relations and Entities}
\label{sec:annotation_relations}

We now describe additional meta-data we collected to utilise spatial relations between the segmentations, using crowdsourcing with gold-standard checks (details in appendix).

\parnobf{Entities}  
EPIC-KITCHENS-100 contains 300 noun {\it classes}, grouping open-vocabulary entity names. 
Of these, VISOR segmentations cover 252 classes from the annotated 36 hours.
VISOR introduces 263 new entities spanning 83 noun classes. We manually re-cluster these into the 300 EPIC-KITCHENS classes. These entities are mainly objects that are {\it handled} in everyday actions but not mentioned in narrations, such as  `coffee package', which is grouped into the `package' class. 
We also introduce five new classes: `left hand', `right hand' (previously only `hand'), `left glove', `right glove' (previously only `glove') and `foot'. As a result, VISOR contain 257 entity classes.

\parnobf{Hand-Object Relations} The activities depicted in VISOR are driven by hands interacting with the scene. We augment each hand segmentation in the dataset with contact information indicating whether it is in contact and, if so, which segment the hand is best described as contacting.  
We identify candidate segments as those sharing a border with the hand. Human annotators select from these segments, or alternatively identify the hand as ``not in physical contact'' with any object or select ``none of the above'' for occluded hands.
Where Annotators cannot come to a consensus, we mark these as inconclusive. We ignore hands annotated with ``none of the above'' or inconclusive decisions during training and evaluation.
Occasionally, in the kitchen, the hand is concealed by a glove, in cases of washing, cleaning or oven mats for hot surfaces. 
We thus also annotate all gloves to find cases where the glove is on the hand.
We further annotate these on-hand gloves and their relations to active objects, in the same manner as visible hands.
We consider both hands and on-hand gloves as acting hands for all relevant benchmarks.

\parnobf{Exhaustive Annotations} 
Recall that we annotate {\it active} objects rather than {\it all objects}. Thus, one cupboard in use can be segmented while another cupboard would not be  segmented. To facilitate use of our annotations in other tasks, we flag each segment indicating whether it has been {\it exhaustively} annotated~\cite{gupta2019lvis}, i.e., there are no more instances of the entity in the image.
Similarly, lack of consensus is marked as inconclusive.
Since the task is imbalanced towards exhaustive, we add one decoy task, that is artificially made non-exhaustive, for every four real tasks to prevent annotation fatigue. 

\subsection{Dense annotations}
\label{sec:dense}

We interpolate manual sparse annotations to produce dense interpolations where feasible.
Inspired by~\cite{wang2021unidentified}, we perform forward and backward multi-mask propagation between every two manually annotated {\it sparse} frames. The two directions' logits are averaged with a weight on each based on the distance from the manual frame. 
We exclude objects that appear in only one of the two frames,, except for hands which are always interpolated.
Additionally, we exclude interpolations longer than~5s.

This process yields 14.5M generated masks in 3.2M train/val images (compared to 5.6M masks in 683K images in~\cite{wang2021unidentified}). 
We filter interpolations by starting from the middle frame, in each interpolation, and track these automatic segmentations back to the first manually-annotated frame and forward to the last frame.
We compute the average $\mathcal{J}$\&$\mathcal{F}$~\cite{Perazzi2016} VOS metric to score each reconstructed mask. We reject all interpolated masks with $\mathcal{J}$\&$\mathcal{F}$ $\le 85$.
As a result of filtering, we keep 69.4\% of automatic interpolations with a total of 9.9M masks, which we release as dense annotations (\href{https://epic-kitchens.github.io/VISOR}{sample in video}).

\section{Dataset Statistics and Analysis} 
\label{sec:statistics}

\begin{figure*}[t]
    \centering
    \includegraphics[width=1.0\textwidth]{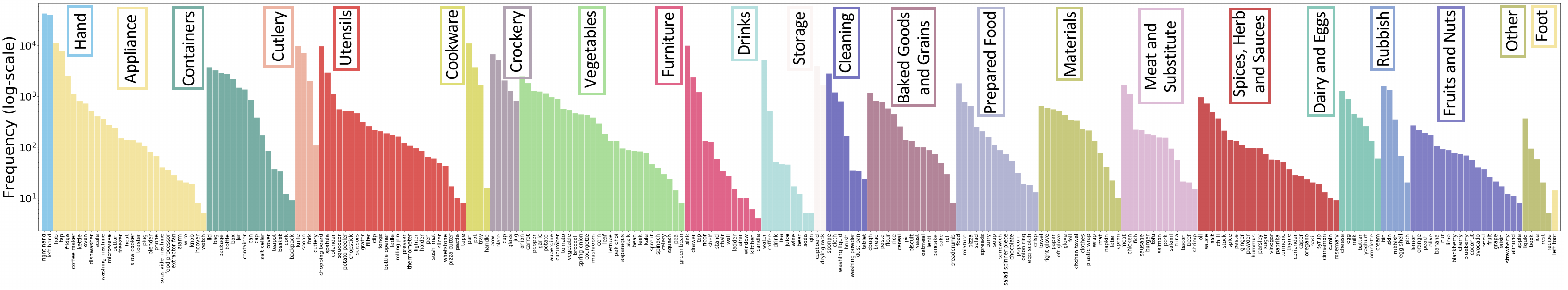} 
    \vspace*{-14pt}
    \caption{{\bf Frequency of Segmented Entity Classes.} (Log y-axis) Some classes are far more frequent. Histogram is long tailed with many rare objects (e.g., `hoover', `cork'). Best viewed with zoom.} 
    \label{fig:Entity_Distribution}
    \includegraphics[width=1.0\textwidth]{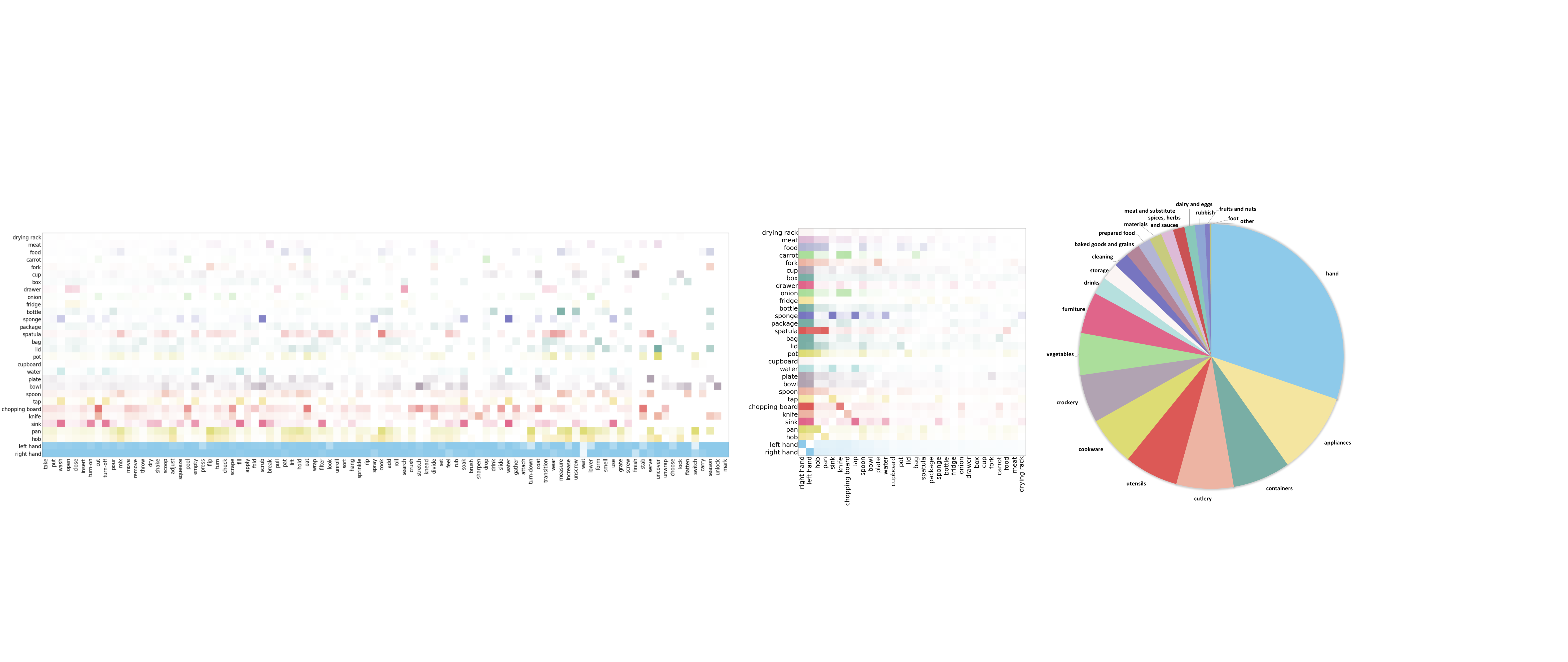} 
    \vspace*{-10pt}
     \caption{Frame-level co-occurences of  verbs-entities ({\bf Left}) and intra-entities ({\bf Middle}). Percentage of annotated pixels per macro-class ({\bf Right}) - large objects (e.g. furniture, storage) are visible.}
     \label{fig:Confusion_Matrix}
    \includegraphics[width=1.0\textwidth]{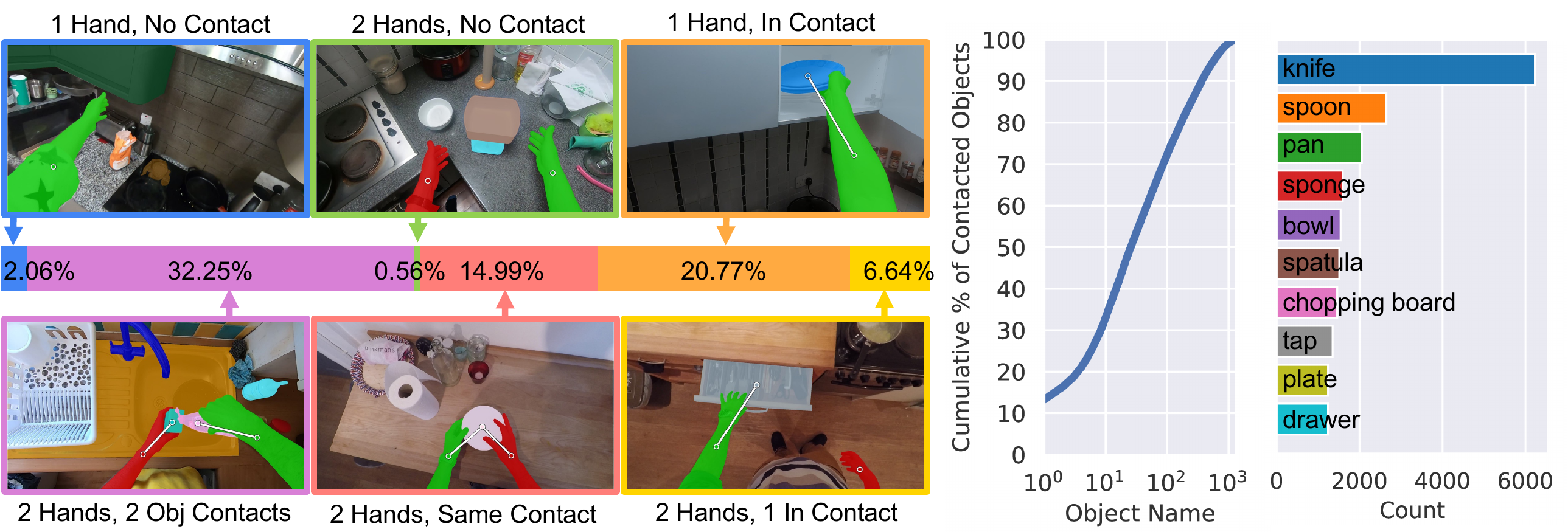}
    \caption{{\bf Hand Object Relation Statistics.} ({\bf Left}) Samples from 6 kinds of hand-object contacts with 1 or 2 hands. ({\bf Middle}) Statistics about the fraction of entities making up the contacted objects, showing there is a long tail. ({\bf Right}) Top-10 most common objects in contact with hands.}
    \vspace*{-12pt}
    \label{fig:hos_stats}
\end{figure*}

Having introduced our dataset and  pipeline, 
we present analysis by answering four questions.

\parnobfq{What objects are segmented}
Fig.~\ref{fig:Entity_Distribution} shows the segmented entity classes, clustered into 22 macro-classes, ordered by the macro-class frequency. Hands are common due to the egocentric perspective, and right hands are slightly more frequent than left ones. The next most frequent categories tend to be appliances, cutlery, and cookware like `hob', `knife', and `pan'. However, the data has a long tail with segments for many rare objects such as `watch' (5) and `candle' (4).

\parnobfq{What Objects Co-Occur} Correlations in Fig.~\ref{fig:Confusion_Matrix}
show frequent co-occurrences between actions and objects (e.g., `measure'/`spoon') and amongst objects (e.g., `hob'/`pan'). Although our data shows many co-occurrences, there is substantial entropy too, demonstrating the dataset's diversity.

\parnobfq{What are hands doing} 
The data shows diverse hand interactions. 
Hands are marked as visible in 94.8\% of annotated images. 
We annotated 72.7K hand states, of which 67.2K are in contact.
Of these in-contact relations, 0.9K are contacts with on-hand gloves (76\% of all gloves are marked as on-hand).
Fig.~\ref{fig:hos_stats} presents hand-object relation statistics on 77\% of all images, for which conclusive annotations are present for both hands.
The figure shows that 54\% of these images include both hands in some contact with the world. Some objects like knives are  over-represented in hand-object relations, but because the data is unscripted, there is a heavy tail: the least-contacted 90\% of entities make up 24\% of hand-object relations. 

\parnobfq{How prevalent are active objects}
Although we annotate active objects, 91.1\% of the objects are exhaustively annotated. 4.1\% of the objects are inexhaustive, and 4.8\% had inconclusive annotations. Consistently, exhaustively annotated objects include `chopping board' and `spatula'. In contrast, `sponge' and `plate' are likely to have more inactive instances in view. Inconclusive samples occur with motion blur or objects with semantic ambiguity (e.g., `sauce', `pan'/`pot').

\section{Challenges and Baselines}
\label{sec:challenges}

We now define three challenges that showcase our annotations and comprise the new VISOR benchmark suite. 
We select the three challenges so as to demonstrate the various aspects of VISOR: short-term understanding of object transformations, in-frame hand-object relations and ultra-long video understanding respectively.
Our first task  (\S\ref{sec:challenge_videoseg}) is {\it Semi-Supervised Video Object Segmentation}, or tracking multiple segments from the first frame over time through a short subsequence. Our second task, detecting {\it Hand-Object Segmentation Relations} (\S\ref{sec:challenge_hos}), segments in-contact objects or active objects along with the manipulating hand. Our final task, titled {\it Where Did This Come From} (\S\ref{sec:challenge_where}) tracks highlighted segmentations back in time to find where they were acquired from.
{\it We report all baseline implementation details in the appendix.}

\subsection{Semi-Supervised Video Object Segmentation (VOS)}
\label{sec:challenge_videoseg}

\parnobf{Definition} We follow the definition of this task from DAVIS~\cite{pont20172017}. 
As explained in \S\ref{sec:annotation_entities}, we divide our videos into shorter sub-sequences. 
Given a sub-sequence of frames with M object masks in the first frame, the goal is to segment these through the remaining frames.
Other objects not present in the first frame of the sub-sequence are excluded from this benchmark.
Note that any of the $M$ objects can be occluded or out of view, and can reappear during the subsequene.  
We include statistics of train/val/test split for this benchmark in Appendix~\ref{sec:app:vos:data}.

\parnobf{Evaluation Metrics} Following the standard protocol used by~\cite{pont20172017, xu2018youtube}, we use the Jaccard Index/Intersection over Union ($\mathcal{J}$) and Boundary F-Measure ($\mathcal{F}$) metrics.
Unlike other datasets, such as DAVIS~\cite{pont20172017}, we use all annotated frames, including the last frame, in our evaluation. Moreover, we report the scores for unseen kitchens to assess generalisation.

\parnobf{Baselines and Results} We use Space-Time Memory Networks (STM)~\cite{Oh_2019_ICCV} as a baseline on this task and follow their 2-stage training strategy. First, we train using COCO~\cite{lin2014microsoft} images by synthesising a video of 3 images from random affine transforms.
We fine-tune this model using VISOR training data, again sampling 3 images from a sub-sequence. To learn long-term appearance change, we increase the gap between sampled frames during training in a curriculum learning fashion.

\begin{figure*}[t]
    \centering
    \includegraphics[width=1.0\textwidth]{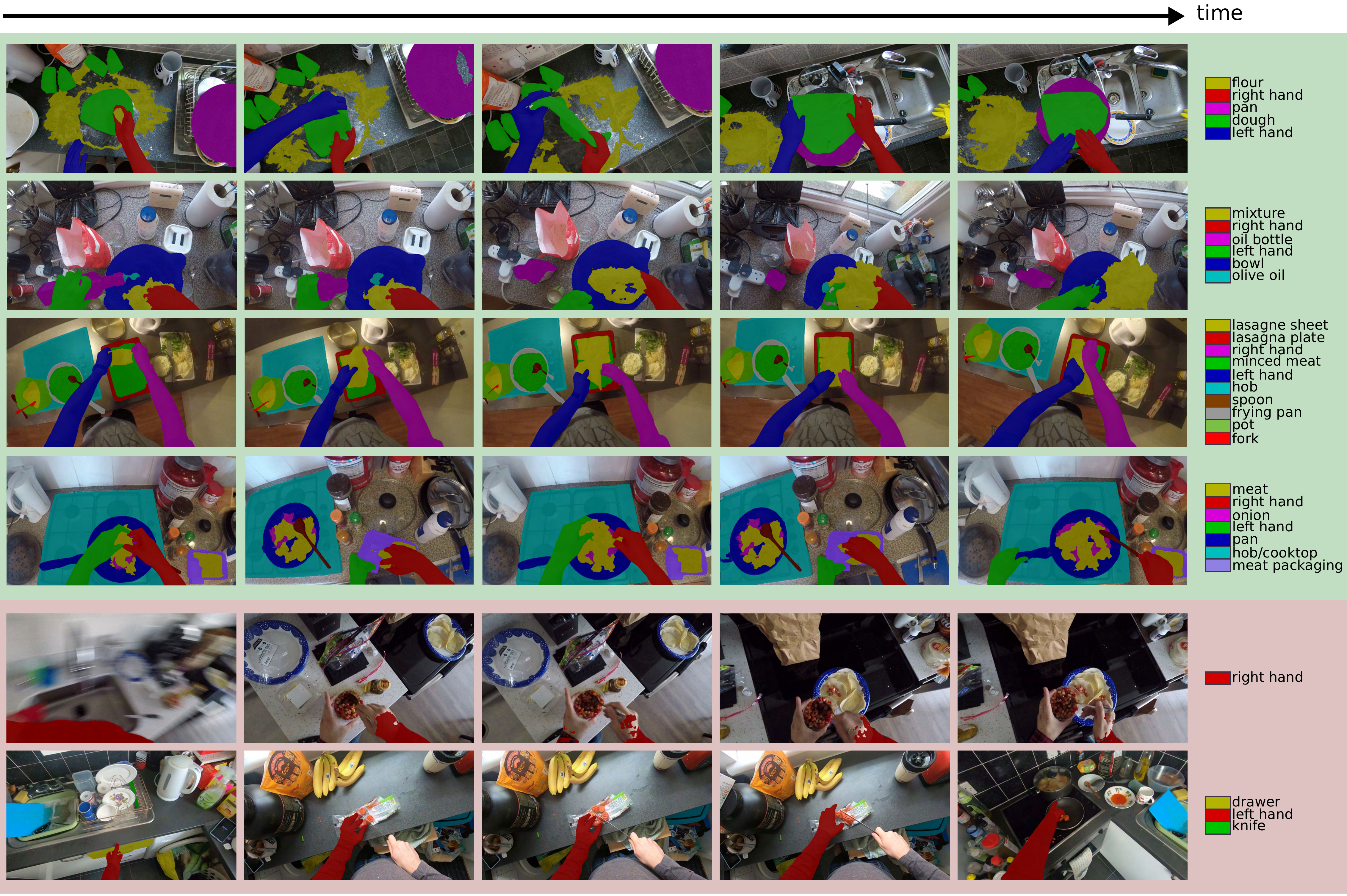}
    \caption{{\bf Qualitative Results of STM for VOS on  Validation Set.} \textcolor{greensuccess}{4 successes}  and \textcolor{redfailure}{2 failures}. Frames are ordered chronologically left-to-right with the first being the reference frame.}
    \vspace*{-12pt}
    \label{fig:vos_qualitative}
\end{figure*}

We report the results in Table~\ref{tab:challenge_videoseg}. {\it Pre-training on COCO} suffers as many categories are specific to VISOR (e.g. left/right hands) or not present in COCO (e.g. potato), and due to the limitations of synthetic sequences.
{Training on \it VISOR only} without pre-training gives slightly better results, but fine-tuning increases the overall score by 13\% on Val.
We show the gap between VISOR and~\cite{pont20172017, xu2018youtube} by reporting the performance of a model trained on these benchmarks.
Fig.~\ref{fig:vos_qualitative} shows 5 qualitative examples.

\begin{table}[t]
    \centering 
    \caption{{\bf VOS Performance}. We use STM~\cite{Oh_2019_ICCV} as a baseline and report varying training schemes. We include the scores for unseen kitchens in both Val and Test. These are considerably lower.
    }
    \label{tab:challenge_videoseg}
    \resizebox{\textwidth}{!}{
    \begin{tabular}{ccccccc|cccc} \toprule
    Pre-train &Fine-Tune &Fine-Tune &\multicolumn{4}{c|}{Val Set Performance} & \multicolumn{4}{c}{Test Set Performance} \\
    COCO~\cite{lin2014microsoft} &Davis\cite{pont20172017}+YT\cite{xu2018youtube}&VISOR &$\mathcal{J}$\&$\mathcal{F} $& $\mathcal{J}$ & $\mathcal{F}$ &{$\mathcal{J}$\&$\mathcal{F} $}\textsubscript{unseen}  &$\mathcal{J}$\&$\mathcal{F}$ & $\mathcal{J}$ & $\mathcal{F}$ &{$\mathcal{J}$\&$\mathcal{F} $}\textsubscript{unseen}  \\ \midrule
$\checkmark$ & & & 56.9 & 55.5 & 58.2 & 48.1 & 58.7 & 57.2 & 60.3 & 57.7
\\
 &&$\checkmark$ & 62.8 & 60.6 & 64.9 &53.9 & 63.7 & 61.9 & 65.5 & 62.2
\\
$\checkmark$ &$\checkmark$ &  &  60.9  &  59.4  &  62.4 &  55.8  &  64.5 &  62.2  & 66.7  &  63.7\\
$\checkmark$ &&$\checkmark$  & \bf 75.8 & \bf 73.6 & \bf 78.0 & \bf 71.2 &  \bf78.0 & \bf 75.8 &\bf 81.0 & \bf 76.6
\\ \bottomrule
    \end{tabular}}
    \vspace*{-12pt}
\end{table}

\parnobf{Code and Models}
Available from \url{https://github.com/epic-kitchens/VISOR-VOS}

\parnobf{Limitations and Challenges}
Segmenting objects through transformations offers opportunities for new research. Common failures arise from: occlusions, particularly in the reference frame (Fig.~\ref{fig:vos_qualitative} row~4 where the knife is initially occluded by the drying rack), tiny objects, motion blur (Fig.~\ref{fig:vos_qualitative} row~5), and drastic transformations, e.g. objects being cut, broken or mixed into ingredients.

\subsection{Hand-Object Segmentation (HOS) Relations}
\label{sec:challenge_hos}

\parnobf{Definition} The goal of this benchmark is to estimate the relation between the hands and the objects given a single frame as input.
Our first task is {\it Hand-Contact-Relation} as in  ~\cite{shan2020understanding}. We characterize each hand in terms of side (left vs right), contact (contact vs no-contact), and segment each hand and contacted object.
Each contacted segment must be associated with a hand, and in the case of segments with multiple connected components (e.g., multiple onions), we only use the component in contact with the hand. Our second task is {\it Hand-And-Active-Object}, or segmenting hands and all active objects (in-contact or otherwise). 
In both tasks, we also consider on-hand gloves as hands.

\parnobf{Related Datasets} Multiple past datasets have provided similar data for hands~\cite{VIVA,Bambach15,Mittal11,Narasimhaswamy2019} as well as hands in contact with objects~\cite{narasimhaswamy2020detecting,ragusa2021meccano,shan2020understanding}. Our work is unique in having large numbers of detailed segments for {\it both} hands and objects, distinguishing it from work that has boxes~\cite{shan2020understanding}. Most work with segments has focused only on hands~\cite{Bambach15} or has been gathered at the few thousand scale~\cite{goyal2022human,Shan21}.

\parnobf{Evaluation Metrics} We evaluate via  instance segmentation tasks using the COCO Mask AP~\cite{lin2014microsoft}. Due to large differences in AP for hands and objects, we evaluate per-category. We evaluate {\it Hand-Contact} by instance segmentation with in-contact objects as a class and three schemes for hand classes: all hands as one class; hands split by side; and hands split by contact state. Hand-object association is  evaluated implicitly by requiring each in-contact object to associate with a hand. 

\parnobf{Baselines and Results} For our baseline, we build on PointRend~\cite{kirillov2020pointrend} instance segmentation. We add three auxiliary detection heads to detect hand side (left/right), contact state (contact/no-contact), and an offset to the contacted object (using the scheme and parameterisation from~\cite{shan2020understanding}).
We select the in-contact object as the closest centre to the hand plus offset.

\begin{figure*}[t]
    \centering
    \includegraphics[width=1.0\textwidth]{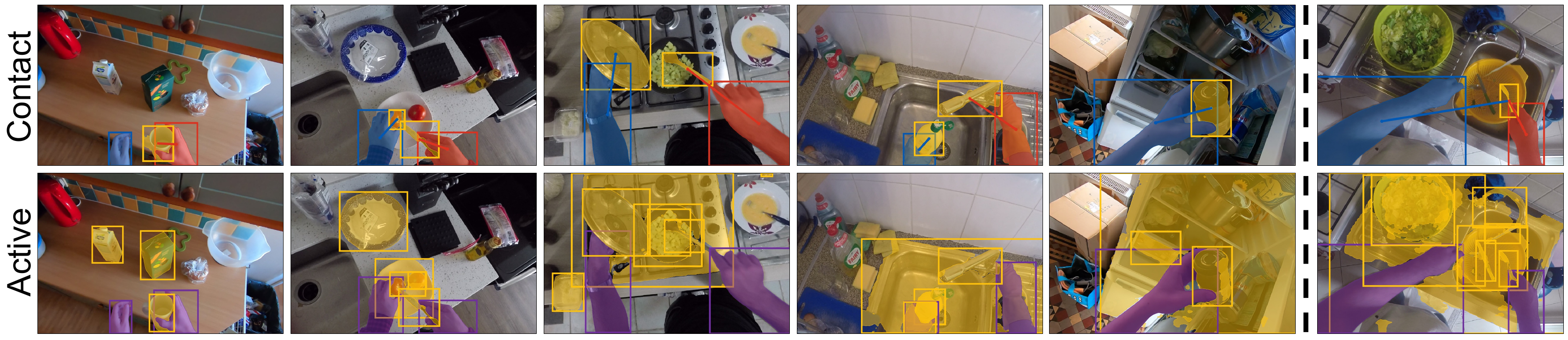} \vspace*{-10pt}
    \caption{{\bf HOS Relations Validation Results.} Contact ({\bf Top}):\textcolor{blue}{{ Left Hands}}, \textcolor{red}{{ Right Hands}}, \textcolor{Dandelion}{{ Objects}};
    Active Object ({\bf Bottom}):\textcolor{Thistle}{{ Hands}}, \textcolor{Dandelion}{{Objects}}. Hands are well-segmented, as are many of the in-contact objects. In `Active', the method finds all active objects regardless of whether they are in contact. The rightmost shows a failure when the object is not detected leading to wrong hand-object relation.}
    \vspace*{-10pt}
   
    \label{fig:hos_active_results}
\end{figure*}

We use the Training set to evaluate on Val, then use both Train and Val to report results on the Test set.
Table~\ref{tab:challenge_hos} shows quantitative results, and Fig.~\ref{fig:hos_active_results} shows sample qualitative examples on Val. Hands are segmented accurately, with side easier to predict than contact state. 
Objects are harder, which we attribute to the long tail of objects and occlusion. Recognising active objects is harder still, particularly from single images.

\parnobf{Code and Models}
Available from \url{https://github.com/epic-kitchens/VISOR-HOS}

\begin{table}[t]
    \centering
    \caption{{\bf HOS Performance.} Hands are segmented accurately, but  identifying contact is challenging. }
    \label{tab:challenge_hos}
    \begin{tabular}{ccccccccc}
    \toprule
     & \multicolumn{4}{c}{Hand-Contact} & & 
     \multicolumn{2}{c}{Hand-Active Object}
     \\ 
     & Hand & Hand, Side & Hand, Contact & Object & & Hand & Active Object \\ \midrule


Val Mask AP  & 90.9 & 87.1 & 73.5 & 30.5 & & 91.1 & 24.1 \\ 
Test Mask AP & 95.4 & 92.4 & 78.7 & 33.7 & & 95.6 & 25.7  \\ 
\bottomrule
    \end{tabular}
\end{table}

\parnobf{Limitations and Challenges} Predicting hand contact state and segmenting contacted objects is hard due to the difficulty of distinguishing hand overlap and contact, the diversity of held objects, and hand-object and hand-hand occlusion.
Using multiple frames can improve performance.

\subsection{Where Did This Come From (WDTCF)? A Taster Challenge}
\label{sec:challenge_where} 

\parnobf{Definition}
Given a frame from an untrimmed video with a mask indicating a query object, we aim to trace the mask back through time to identify  `where did \emph{this} come from?', where the pronoun \emph{this} refers to the indicated mask. 
For tractability, we use 15 sources from which objects emerge in the dataset: \{fridge, freezer, cupboard, drawer, oven, dishwasher, bag, bottle, box, package,  jar, can, pan, tub, cup\}.
To avoid trivial solutions, the answer must be spatiotemporally grounded in an \emph{evidence} frame(s) when the query object emerges from the source. 

\begin{figure*}[t]
    \centering
    \includegraphics[width=1.0\textwidth]{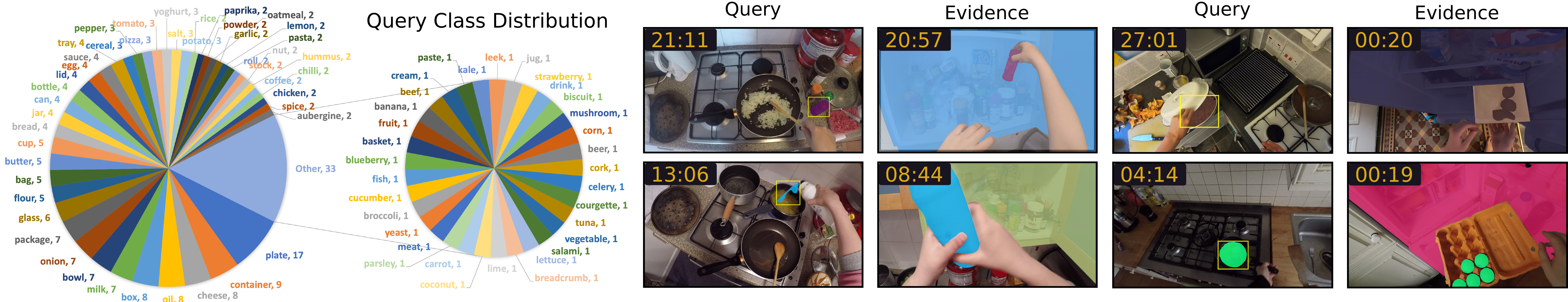}
    \caption{{\bf WDTCF Annotations.} Distributions and examples of query entities and 4 samples.}
    \label{fig:where_confusion_matrix}
\end{figure*}

\parnobf{Annotation Statistics} 
Annotating long-term relations in videos is challenging.
We obtain 222 $\mathit{WDTCF}$ examples from 92 untrimmed videos.
These form this challenge's \emph{test} set.
We encourage self-supervised learning for tackling this challenge.
The distribution of the sources is long-tailed, where `fridge' and `cupboard' occupy the largest portions (37\% and 31\% respectively). 
Fig.~\ref{fig:where_confusion_matrix} shows the 78 query objects and four samples with timestamp differences between the query and evidence frames. 
The gap between the query and evidence frames adds challenge. The gap is 5.4 mins (19K frames) on average, but it varies widely with a standard deviation of 8 mins (min=1s, max=52 mins).

When annotating, we aim to find the furthest container as a source.
For example, if `jam' emerges from a `jar', which is a potential source container, but the `jar' itself is seen retrieved from a `cupboard', the correct answer to the \emph{WDTCF} question is the `cupboard'.
In another case, where the `jar' is only seen on the counter throughout the video, then the `jar' would be considered as the correct source.

\parnobf{Related Datasets}
This challenge is related to VQA tasks in previous datasets~\cite{vizwizGrounding2022,Ego4D2022CVPR,Mercier_2021_WACV}. In~\cite{Ego4D2022CVPR}, the aim is to find the previous encounter of the object itself and~\cite{vizwizGrounding2022} aims to ground the object in a single image. In both, the query object is visible in the evidence image, unlike in our challenge where the query object is occluded and being manipulated. We are also looking for its source, where it was before it is first encountered, rather than when/where it was spotted last.

\parnobf{Evaluation Metrics}
We evaluate by classifying the correct source (accuracy), locating the evidence frame and segmenting the source and query objects in the evidence frame (IoU). If the evidence frame is incorrect, the IoU is 0.

\parnobf{Annotations and code}
Available from \url{https://github.com/epic-kitchens/VISOR-WDTCF}

\parnobf{Baselines and Results}
No prior method has tackled this challenge, so we test baselines using different levels of oracle knowledge.
As shown in Table~\ref{tab:challenge_WDTCF}, guessing from the source distribution gets 24.8\% accuracy; while guessing the most frequent source (`fridge') achieves 37\% accuracy but cannot localise the evidence frame and segment.
We thus trained a PointRend instance segmentation model on the all VISOR classes using our Train and Val data to detect objects. 
Given the query mask, we predict its class using the mask with the highest IoU, then propose the first three frames containing both the object and a source as candidates.
The relative performance suggests that the hardest task is temporally localising the evidence frame; when this is given, performance is significantly boosted.

\begin{table}[t]
    \centering 
    \caption{{\bf WDTCF Performance.} Performance using \emph{oracle} knowledge of distribution of source classes (prior), location of (evidence) frame, (query class) and a model trained with (GT Masks).}
    \label{tab:challenge_WDTCF}
\resizebox{0.9\linewidth}{!}{
    \begin{tabular}{l|cccc|ccc} \toprule
    Method & Prior & Evidence & Query Cls &GT Masks & Source & Query IoU & Source IoU  \\ \midrule
    Random & $\checkmark$ &  &  & & 24.8 & - & -
    \\ \midrule
    \multirow{4}{*}{PointRend~\cite{kirillov2020pointrend}}   & & & & $\checkmark$ & 28.8 & 30.3 & 25.4
    \\
    & $\checkmark$ & & $\checkmark$ & $\checkmark$ & 33.8 & 31.6 & 29.7
    \\
    & & $\checkmark$ & & $\checkmark$ & 88.3 & 79.4 & 78.8
    \\
    & $\checkmark$ & $\checkmark$ & & $\checkmark$ & 94.1 & 79.4 & 84.1
\\ \bottomrule
    \end{tabular}}
    \vspace*{-12pt}
\end{table}

\parnobf{Limitations and Challenges}
Since objects often originate from the same source container, e.g., milk comes from the fridge, bias is expected.
While we have introduced the evidence frame grounding task, shortcuts via the presence of source classes can still be used. 
As noted above, the baselines use oracle knowledge; an AI system to answer WDTCF is left for future work.

\section{Conclusion and Next Steps}

We introduced VISOR, a new dataset built on EPIC-KITCHENS-100 videos, with rich pixel-level annotations across space and time. Through the release of these annotations and the proposed benchmarks, we hope VISOR enables the community to investigate new problems in long-term understanding of the interplay between actions and object transformations.

\acksection{
We gratefully acknowledge valuable support from: Michael Wray for revising the EPIC-KITCHENS-100 classes; Seung Wook Kim and Marko Boben for technical support to TORAS; Srdjan Delic for quality checks particularly on the Test set; several members of the MaVi group at Bristol for quality checking: Toby Perrett, Michael Wray, Dena Bazazian, Adriano Fragomeni, Kevin Flanagan, Daniel Whettam, Alexandros Stergiou, Jacob Chalk, Chiara Plizzari and Zhifan Zhu.

Annotations were funded by a charitable unrestricted donations to the University of Bristol from Procter and Gamble and DeepMind.

Research at the University of Bristol is supported by UKRI Engineering and Physical Sciences Research Council (EPSRC) Doctoral Training Program (DTP), EPSRC Fellowship UMPIRE~(EP/T004991/1) and EPSRC Program Grant Visual AI (EP/T028572/1).
We acknowledge the use of the ESPRC funded Tier 2 facility, JADE, and the University of Bristol's Blue Crystal 4 facility.

Research at the University of Michigan is based upon work
supported by the National Science Foundation under Grant No. 2006619.

Research at the University of Toronto is in part sponsored by NSERC. S.F. also acknowledges support through the Canada CIFAR AI Chair program.  }

{
\small
\bibliographystyle{ieee_fullname}
\bibliography{local}

\begin{thebibliography}{10}\itemsep=-1pt

\bibitem{VIVA}
The vision for intelligent vehicles and applications {(VIVA)} challenge,
  laboratory for intelligent and safe automobiles, {UCSD}.
\newblock http://cvrr.ucsd.edu/vivachallenge/.

\bibitem{polyrnnpp}
David Acuna, Huan Ling, Amlan Kar, and Sanja Fidler.
\newblock Efficient interactive annotation of segmentation datasets with
  polygon-{RNN}++.
\newblock In {\em Proceedings of the IEEE Conference on Computer Vision and
  Pattern Recognition}, 2018.

\bibitem{Bambach15}
Sven Bambach, Stefan Lee, David Crandall, and Chen Yu.
\newblock Lending a hand: Detecting hands and recognizing activities in complex
  egocentric interactions.
\newblock In {\em Proceedings of the IEEE International Conference on Computer
  Vision}, 2015.

\bibitem{Baradel_2018_ECCV}
Fabien Baradel, Natalia Neverova, Christian Wolf, Julien Mille, and Greg Mori.
\newblock Object level visual reasoning in videos.
\newblock In {\em European Conference on Computer Vision (ECCV)}, 2018.

\bibitem{benenson2019large}
Rodrigo Benenson, Stefan Popov, and Vittorio Ferrari.
\newblock Large-scale interactive object segmentation with human annotators.
\newblock In {\em Proceedings of the IEEE Conference on Computer Vision and
  Pattern Recognition}, 2019.

\bibitem{Caelles_arXiv_2019}
Sergi Caelles, Jordi Pont-Tuset, Federico Perazzi, Alberto Montes,
  Kevis-Kokitsi Maninis, and Luc {Van Gool}.
\newblock The 2019 {DAVIS} challenge on {VOS}: Unsupervised multi-object
  segmentation.
\newblock {\em arXiv:1905.00737}, 2019.

\bibitem{carreira2017quo}
Joao Carreira and Andrew Zisserman.
\newblock Quo vadis, action recognition? a new model and the kinetics dataset.
\newblock In {\em Proceedings of the IEEE Conference on Computer Vision and
  Pattern Recognition}, 2017.

\bibitem{polyrnn}
Lluis Castrejon, Kaustav Kundu, Raquel Urtasun, and Sanja Fidler.
\newblock Annotating object instances with a polygon-{RNN}.
\newblock In {\em Proceedings of the IEEE Conference on Computer Vision and
  Pattern Recognition}, 2017.

\bibitem{vizwizGrounding2022}
Chongyan Chen, Samreen Anjum, and Danna Gurari.
\newblock Grounding answers for visual questions asked by visually impaired
  people.
\newblock In {\em Proceedings of the IEEE Conference on Computer Vision and
  Pattern Recognition}, 2022.

\bibitem{Cordts2016Cityscapes}
Marius Cordts, Mohamed Omran, Sebastian Ramos, Timo Rehfeld, Markus Enzweiler,
  Rodrigo Benenson, Uwe Franke, Stefan Roth, and Bernt Schiele.
\newblock The {C}ityscapes {D}ataset for {S}emantic {U}rban {S}cene
  {U}nderstanding.
\newblock In {\em Proceedings of the IEEE Conference on Computer Vision and
  Pattern Recognition}, 2016.

\bibitem{Damen2020RESCALING}
Dima Damen, Hazel Doughty, Giovanni~Maria Farinella, , Antonino Furnari, Jian
  Ma, Evangelos Kazakos, Davide Moltisanti, Jonathan Munro, Toby Perrett, Will
  Price, and Michael Wray.
\newblock Rescaling egocentric vision.
\newblock {\em International Journal of Computer Vision (IJCV)}, 2022.

\bibitem{Damen2018EPICKITCHENS}
Dima Damen, Hazel Doughty, Giovanni~Maria Farinella, Sanja Fidler, Antonino
  Furnari, Evangelos Kazakos, Davide Moltisanti, Jonathan Munro, Toby Perrett,
  Will Price, and Michael Wray.
\newblock Scaling egocentric vision: The {EPIC-KITCHENS D}ataset.
\newblock In {\em Proceedings of the European Conference on Computer Vision},
  2018.

\bibitem{geiger2012we}
Andreas Geiger, Philip Lenz, and Raquel Urtasun.
\newblock Are we ready for autonomous driving? the {K}itti vision benchmark
  suite.
\newblock In {\em Proceedings of the IEEE Conference on Computer Vision and
  Pattern Recognition}. IEEE, 2012.

\bibitem{goyal2022human}
Mohit Goyal, Sahil Modi, Rishabh Goyal, and Saurabh Gupta.
\newblock Human hands as probes for interactive object understanding.
\newblock In {\em Proceedings of the IEEE/CVF Conference on Computer Vision and
  Pattern Recognition}, pages 3293--3303, 2022.

\bibitem{Ego4D2022CVPR}
Kristen Grauman, Andrew Westbury, Eugene Byrne, Zachary Chavis, Antonino
  Furnari, Rohit Girdhar, Jackson Hamburger, Hao Jiang, Miao Liu, Xingyu Liu,
  Miguel Martin, Tushar Nagarajan, Ilija Radosavovic, Santhosh~Kumar
  Ramakrishnan, Fiona Ryan, Jayant Sharma, Michael Wray, Mengmeng Xu,
  Eric~Zhongcong Xu, Chen Zhao, Siddhant Bansal, Dhruv Batra, Vincent
  Cartillier, Sean Crane, Tien Do, Morrie Doulaty, Akshay Erapalli, Christoph
  Feichtenhofer, Adriano Fragomeni, Qichen Fu, Christian Fuegen, Abrham
  Gebreselasie, Cristina Gonzalez, James Hillis, Xuhua Huang, Yifei Huang,
  Wenqi Jia, Weslie Khoo, Jachym Kolar, Satwik Kottur, Anurag Kumar, Federico
  Landini, Chao Li, Yanghao Li, Zhenqiang Li, Karttikeya Mangalam, Raghava
  Modhugu, Jonathan Munro, Tullie Murrell, Takumi Nishiyasu, Will Price,
  Paola~Ruiz Puentes, Merey Ramazanova, Leda Sari, Kiran Somasundaram, Audrey
  Southerland, Yusuke Sugano, Ruijie Tao, Minh Vo, Yuchen Wang, Xindi Wu,
  Takuma Yagi, Yunyi Zhu, Pablo Arbelaez, David Crandall, Dima Damen,
  Giovanni~Maria Farinella, Bernard Ghanem, Vamsi~Krishna Ithapu, C.~V.
  Jawahar, Hanbyul Joo, Kris Kitani, Haizhou Li, Richard Newcombe, Aude Oliva,
  Hyun~Soo Park, James~M. Rehg, Yoichi Sato, Jianbo Shi, Mike~Zheng Shou,
  Antonio Torralba, Lorenzo Torresani, Mingfei Yan, and Jitendra Malik.
\newblock Ego4{D}: Around the {W}orld in 3,000 {H}ours of {E}gocentric {V}ideo.
\newblock In {\em Proceedings of the IEEE Conference on Computer Vision and
  Pattern Recognition}, 2022.

\bibitem{gu2018ava}
Chunhui Gu, Chen Sun, David~A Ross, Carl Vondrick, Caroline Pantofaru, Yeqing
  Li, Sudheendra Vijayanarasimhan, George Toderici, Susanna Ricco, Rahul
  Sukthankar, et~al.
\newblock A{VA}: A video dataset of spatio-temporally localized atomic visual
  actions.
\newblock In {\em Proceedings of the IEEE Conference on Computer Vision and
  Pattern Recognition}, 2018.

\bibitem{gupta2019lvis}
Agrim Gupta, Piotr Dollar, and Ross Girshick.
\newblock L{VIS}: A dataset for large vocabulary instance segmentation.
\newblock In {\em Proceedings of the IEEE Conference on Computer Vision and
  Pattern Recognition}, 2019.

\bibitem{Hasson19}
Yana Hasson, G{\"u}l Varol, Dimitrios Tzionas, Igor Kalevatykh, Michael~J.
  Black, Ivan Laptev, and Cordelia Schmid.
\newblock Learning joint reconstruction of hands and manipulated objects.
\newblock In {\em Proceedings of the IEEE Conference on Computer Vision and
  Pattern Recognition (CVPR)}, 2019.

\bibitem{ji2020action}
Jingwei Ji, Ranjay Krishna, Li Fei-Fei, and Juan~Carlos Niebles.
\newblock Action {G}enome: Actions as compositions of spatio-temporal scene
  graphs.
\newblock In {\em Proceedings of the IEEE Conference on Computer Vision and
  Pattern Recognition}, 2020.

\bibitem{torontoannotsuite}
Amlan Kar, Seung~Wook Kim, Marko Boben, Jun Gao, Tianxing Li, Huan Ling, Zian
  Wang, and Sanja Fidler.
\newblock Toronto annotation suite.
\newblock \url{https://aidemos.cs.toronto.edu/toras}, 2021.

\bibitem{kirillov2020pointrend}
Alexander Kirillov, Yuxin Wu, Kaiming He, and Ross Girshick.
\newblock Pointrend: Image segmentation as rendering.
\newblock In {\em Proceedings of the IEEE Conference on Computer Vision and
  Pattern Recognition}, 2020.

\bibitem{li2020ava}
Ang Li, Meghana Thotakuri, David~A Ross, Joao Carreira, Alexander Vostrikov,
  and Andrew Zisserman.
\newblock The {AVA}-{K}inetics localized human actions video dataset.
\newblock {\em arXiv preprint arXiv:2005.00214}, 2020.

\bibitem{lin2014microsoft}
Tsung-Yi Lin, Michael Maire, Serge Belongie, James Hays, Pietro Perona, Deva
  Ramanan, Piotr Doll{\'a}r, and C~Lawrence Zitnick.
\newblock Microsoft {COCO}: Common objects in context.
\newblock In {\em Proceedings of the European Conference on Computer Vision}.
  Springer, 2014.

\bibitem{CurveGCN2019}
Huan Ling, Jun Gao, Amlan Kar, Wenzheng Chen, and Sanja Fidler.
\newblock Fast interactive object annotation with curve-gcn.
\newblock In {\em Proceedings of the IEEE Conference on Computer Vision and
  Pattern Recognition}, 2019.

\bibitem{Liu_2022_CVPR}
Shaowei Liu, Subarna Tripathi, Somdeb Majumdar, and Xiaolong Wang.
\newblock Joint hand motion and interaction hotspots prediction from egocentric
  videos.
\newblock In {\em Proceedings of the IEEE/CVF Conference on Computer Vision and
  Pattern Recognition (CVPR)}, 2022.

\bibitem{Mercier_2021_WACV}
Jean-Philippe Mercier, Mathieu Garon, Philippe Giguere, and Jean-Francois
  Lalonde.
\newblock Deep template-based object instance detection.
\newblock In {\em Proceedings of the IEEE Winter Conference on Applications of
  Computer Vision}, January 2021.

\bibitem{Mittal11}
A. Mittal, A. Zisserman, and P.~H.~S. Torr.
\newblock Hand detection using multiple proposals.
\newblock In {\em British Machine Vision Conference}, 2011.

\bibitem{narasimhaswamy2020detecting}
Supreeth Narasimhaswamy, Trung Nguyen, and Minh Hoai.
\newblock Detecting hands and recognizing physical contact in the wild.
\newblock In {\em Neural Information Processing Systems}, 2020.

\bibitem{Narasimhaswamy2019}
Supreeth Narasimhaswamy, Zhengwei Wei, Yang Wang, Justin Zhang, and Minh Hoai.
\newblock Contextual attention for hand detection in the wild.
\newblock In {\em Proceedings of the IEEE International Conference on Computer
  Vision}, 2019.

\bibitem{Oh_2019_ICCV}
Seoung~Wug Oh, Joon-Young Lee, Ning Xu, and Seon~Joo Kim.
\newblock Video object segmentation using space-time memory networks.
\newblock In {\em Proceedings of the IEEE International Conference on Computer
  Vision}, October 2019.

\bibitem{Perazzi2016}
F. Perazzi, J. Pont-Tuset, B. McWilliams, L. {Van Gool}, M. Gross, and A.
  Sorkine-Hornung.
\newblock A benchmark dataset and evaluation methodology for video object
  segmentation.
\newblock In {\em Proceedings of the IEEE Conference on Computer Vision and
  Pattern Recognition}, 2016.

\bibitem{pont20172017}
Jordi Pont-Tuset, Federico Perazzi, Sergi Caelles, Pablo Arbel{\'a}ez, Alex
  Sorkine-Hornung, and Luc Van~Gool.
\newblock The 2017 {DAVIS} challenge on video object segmentation.
\newblock {\em arXiv preprint arXiv:1704.00675}, 2017.

\bibitem{ragusa2022meccano}
Francesco Ragusa, Antonino Furnari, and Giovanni~Maria Farinella.
\newblock Meccano: A multimodal egocentric dataset for humans behavior
  understanding in the industrial-like domain.
\newblock {\em arXiv preprint arXiv:2209.08691}, 2022.

\bibitem{ragusa2021meccano}
Francesco Ragusa, Antonino Furnari, Salvatore Livatino, and Giovanni~Maria
  Farinella.
\newblock The {M}eccano dataset: Understanding human-object interactions from
  egocentric videos in an industrial-like domain.
\newblock In {\em Proceedings of the IEEE Winter Conference on Applications of
  Computer Vision}, 2021.

\bibitem{shan2020understanding}
Dandan Shan, Jiaqi Geng, Michelle Shu, and David~F Fouhey.
\newblock Understanding human hands in contact at internet scale.
\newblock In {\em Proceedings of the IEEE Conference on Computer Vision and
  Pattern Recognition}, 2020.

\bibitem{Shan21}
Dandan Shan, Richard~E.L. Higgins, and David~F. Fouhey.
\newblock {COHESIV}: Contrastive object and hand embedding segmentation in
  video.
\newblock In {\em Neural Information Processing Systems}, 2021.

\bibitem{sigurdsson2016hollywood}
Gunnar~A Sigurdsson, G{\"u}l Varol, Xiaolong Wang, Ali Farhadi, Ivan Laptev,
  and Abhinav Gupta.
\newblock Hollywood in homes: Crowdsourcing data collection for activity
  understanding.
\newblock In {\em Proceedings of the European Conference on Computer Vision}.
  Springer, 2016.

\bibitem{Taheri_2022_CVPR}
Omid Taheri, Vasileios Choutas, Michael~J. Black, and Dimitrios Tzionas.
\newblock Goal: Generating 4d whole-body motion for hand-object grasping.
\newblock In {\em Proceedings of the IEEE/CVF Conference on Computer Vision and
  Pattern Recognition (CVPR)}, 2022.

\bibitem{Taheri2020}
Omid Taheri, Nima Ghorbani, Michael~J. Black, and Dimitrios Tzionas.
\newblock Grab: A dataset of whole-body human grasping of objects.
\newblock In Andrea Vedaldi, Horst Bischof, Thomas Brox, and Jan-Michael Frahm,
  editors, {\em European Conference on Computer Vision (ECCV)}, 2020.

\bibitem{Turpin2022}
Dylan Turpin, Liquan Wang, Eric Heiden, Yun-Chun Chen, Miles Macklin, Stavros
  Tsogkas, Sven Dickinson, and Animesh Garg.
\newblock Grasp’d: Differentiable contact-rich grasp synthesis for
  multi-fingered hands.
\newblock In {\em European Conference on Computer Vision (ECCV)}, 2022.

\bibitem{wang2021unidentified}
Weiyao Wang, Matt Feiszli, Heng Wang, and Du Tran.
\newblock Unidentified video objects: A benchmark for dense, open-world
  segmentation.
\newblock In {\em Proceedings of the IEEE International Conference on Computer
  Vision}, 2021.

\bibitem{DELSE2019}
Zian Wang, David Acuna, Huan Ling, Amlan Kar, and Sanja Fidler.
\newblock Object instance annotation with deep extreme level set evolution.
\newblock In {\em Proceedings of the IEEE Conference on Computer Vision and
  Pattern Recognition}, 2019.

\bibitem{xu2018youtube}
Ning Xu, Linjie Yang, Yuchen Fan, Dingcheng Yue, Yuchen Liang, Jianchao Yang,
  and Thomas Huang.
\newblock Youtube-{VOS}: A large-scale video object segmentation benchmark.
\newblock {\em arXiv preprint arXiv:1809.03327}, 2018.

\bibitem{xu2021d3dhoi}
Xiang Xu, Hanbyul Joo, Greg Mori, and Manolis Savva.
\newblock D3d-hoi: Dynamic 3d human-object interactions from videos.
\newblock {\em arXiv preprint arXiv:2108.08420}, 2021.

\bibitem{zhou2017ade20k}
Bolei Zhou, Hang Zhao, Xavier Puig, Sanja Fidler, Adela Barriuso, and Antonio
  Torralba.
\newblock Scene parsing through {ADE}20k dataset.
\newblock In {\em Proceedings of the IEEE Conference on Computer Vision and
  Pattern Recognition}, 2017.

\end{thebibliography}
}

\section*{Checklist}

\begin{enumerate}

\item For all authors...
\begin{enumerate}
  \item Do the main claims made in the abstract and introduction accurately reflect the paper's contributions and scope?
    \answerYes{This paper introduces a dataset, with annotation pipeline and challenges. Throughout the paper and appendix, we thoroughly describe our annotation pipeline, summarise the statistics, compare to other datasets. For each benchmark, we clearly define the benchmark, metrics and report baseline and qualitative results.}
  \item Did you describe the limitations of your work?
    \answerYes{We discuss limitations of our challenges in \S\ref{sec:challenge_videoseg}, \S\ref{sec:challenge_hos}, \S\ref{sec:challenge_where}}
  \item Did you discuss any potential negative societal impacts of your work?
    \answerYes{We mention this in \S\ref{sec:collection}. We also devote a section in appendix to potential bias in the data and resulting models}
  \item Have you read the ethics review guidelines and ensured that your paper conforms to them?
    \answerYes{}
\end{enumerate}

\item If you are including theoretical results...
\begin{enumerate}
  \item Did you state the full set of assumptions of all theoretical results?
    \answerNA{There are no theoretical results.}
	\item Did you include complete proofs of all theoretical results?
    \answerNA{There are no theoretical results.}
\end{enumerate}

\item If you ran experiments...
\begin{enumerate}
  \item Did you include the code, data, and instructions needed to reproduce the main experimental results (either in the appendix material or as a URL)?
    \answerYes{Train/Val annotations are publicly available from: \url{https://epic-kitchens.github.io/VISOR}. Code for reproducing the first two challenges is published on GitHub. Test set annotations, including those for the third challenge will remain hidden. A leaderboard available from: \url{https://epic-kitchens.github.io}.}
  \item Did you specify all the training details (e.g., data splits, hyperparameters, how they were chosen)?
    \answerYes{Dataset split (Train/Val/Test) is summarised in Sec 2.2, and specifically for semi-supervised VOS in Appendix. Training details per challenge are detailed in the appendix.}
	\item Did you report error bars (e.g., with respect to the random seed after running experiments multiple times)?
    \answerNo{The experiments run are too computationally costly to do multiple runs of the experiments. Our results form baselines that we expect more intelligent methods to beat. Code allows replication of baselines where needed.}
	\item Did you include the total amount of compute and the type of resources used (e.g., type of GPUs, internal cluster, or cloud provider)?
    \answerYes{We include an analysis of resources used in the appendix.}
\end{enumerate}

\item If you are using existing assets (e.g., code, data, models) or curating/releasing new assets...
\begin{enumerate}
  \item If your work uses existing assets, did you cite the creators?
    \answerYes{We build on the EPIC-KITCHENS-100 dataset, which we extensively cite throughout the paper. We also cite models we use to bulid baselines.}
  \item Did you mention the license of the assets?
    \answerYes{VISOR dataset is available under the same licenses as EPIC-KITCHENS}
  \item Did you include any new assets either in the appendix material or as a URL?
    \answerYes{We introduce new assets in the form of annotations. We share these privately with reviewers and will share publicly in August.}
  \item Did you discuss whether and how consent was obtained from people whose data you're using/curating?
    \answerYes{We mention the dataset, EPIC-KITCHENS-100, that was used. This dataset was collected with ethics approval and with  consent from the people who are in the data.}
  \item Did you discuss whether the data you are using/curating contains personally identifiable information or offensive content?
    \answerYes{The EPIC-KITCHENS videos do not contain any personally identifiable information. It was collected from participants who reviewed the footage prior to publication.}
\end{enumerate}

\item If you used crowdsourcing or conducted research with human subjects...
\begin{enumerate}
  \item Did you include the full text of instructions given to participants and screenshots, if applicable?
    \answerYes{Please see the appendix material for detailed descriptions of all instructions for annotators.}
  \item Did you describe any potential participant risks, with links to Institutional Review Board (IRB) approvals, if applicable?
    \answerNA{There are no human studies done in this paper. However, the dataset that we build upon, EPIC-KITCHENS-100 was collected with IRB approval, as described in \S\ref{sec:collection}. }
  \item Did you include the estimated hourly wage paid to participants and the total amount spent on participant compensation?
    \answerYes{Freelance annotators were paid hourly, and self-reported their hours via the Upwork platform. Hourly rate varied by experience and submitted proposal from \$6-\$9/hour, with the higher rate reserved for most experienced annotators who were also in charge of QA checks. We report compensation in the appendix.}
\end{enumerate}

\end{enumerate}


\newpage
\appendix

\clearpage

\renewcommand{\parnobf}[1]{{\vspace{1mm} \noindent \textbf{{#1}.}}}
\renewcommand{\parnobfq}[1]{{\vspace{1mm} \noindent \textbf{{#1}?}}}
\renewcommand{\parnoit}[1]{{\vspace{1mm} \noindent \textit{{#1}.}}}
\renewcommand{\DS}[1]{{\color{orange}{[Dandan: #1]}}}
\renewcommand{\SF}[1]{{\color{magenta}{[Sanja: #1]}}}
\renewcommand{\ToDo}[2]{{\noindent \textcolor{red}{\textbf{#1}}: \textcolor{red}{#2}}}
\renewcommand{\note}[1]{{\color{red}{{\bf #1}}}}

\begin{center}
{\Large EPIC-KITCHENS VISOR Benchmark \\ VIdeo Segmentations and Object Relations -- Appendix}
\end{center}

This appendix contains additional details about the VISOR dataset. 
\begin{itemize}[leftmargin=24pt,label={}]
    \item \S\ref{sec:app:impact} \nameref{sec:app:impact} describes the societal impact of this dataset, the resources used, and the availability of the dataset. This is also expanded upon in the datasheet in \S\ref{sec:app:datasheet}.
    
    \item \S\ref{sec:app:stage1} \nameref{sec:app:stage1} describes the selection of entities to be annotated, frames on which annotations are made, as well as subsequences. This is the first part of our annotation process and sets up the pixel-wise annotations.
    
    \item \S\ref{sec:app:stage23} \nameref{sec:app:stage23} describes the training of the annotators the TORAS annotation suite. 
    
    \item \S\ref{sec:app:toras} \nameref{sec:app:toras} describes the TORonto Annotation Suite (TORAS) that was used for annotating the pixel labels.
    
    \item \S\ref{sec:app:correction} \nameref{sec:app:correction} describes the corrections done to these annotations.
    
    \item \S\ref{sec:app:stage4} \nameref{sec:app:stage4} describes the annotation of the relationships between the segments. This is the last part of our annotation.
    
    \item \S\ref{sec:app:dense} \nameref{sec:app:dense} describes the dense annotations that we provide and how they were obtained.

    \item \S\ref{sec:app:VOS} \nameref{sec:app:VOS} describes the Video Object Segmentation (VOS) benchmark, including data preparation (\S\ref{sec:app:vos:data}), metrics (\S\ref{sec:app:vos:metrics}), baselines (\S\ref{sec:app:vos:baselines}), and additional results (\S\ref{sec:app:vos:visualizations}).
    
    \item \S\ref{sec:app:hos} \nameref{sec:app:hos} describes the Hand Object Segmentation (HOS) benchmark, including data preparation (\S\ref{sec:app:hos:data}), metrics (\S\ref{sec:app:hos:metrics}), baselines (\S\ref{sec:app:hos:baselines}), and additional results (\S\ref{sec:app:hos:visualizations}).
    
    \item \S\ref{sec:app:wdtcf}
    \nameref{sec:app:wdtcf} describes the {\it Where Did This Come From?} benchmark, including data preparation (\S\ref{sec:app:wdtcf:data}), metrics (\S\ref{sec:app:wdtcf:metrics}), baselines (\S\ref{sec:app:wdtcf:baselines}), and additional results (\S\ref{sec:app:wdtcf:visualizations}).
    
    \item \S\ref{sec:app:datasheet}
    \nameref{sec:app:datasheet} describes the datasheet for the VISOR dataset.
\end{itemize}

\subsection*{Supplemental References}

Implementation details in the appendix refers to two other items that were not reference in the main paper. We have put these references here:

\begin{enumerate}[label={[\Alph*]}]
\item 
\label{refA}
Georgia Gkioxari, Ross Girshick, Piotr Dollar, and Kaiming He. Detecting and recognizing human-object
interactions. {\it In Proceedings of the IEEE Conference on Computer Vision and Pattern Recognition}, 2018.

\item 
\label{refB}
Yuxin Wu, Alexander Kirillov, Francisco Massa, Wan-Yen Lo, and Ross Girshick. Detectron2. \href{https://github.com/facebookresearch/detectron2}{https://github.com/facebookresearch/detectron2}, 2019
\end{enumerate}

\clearpage

\section{Appendix - Societal Impact and Resources Used}
\label{sec:app:impact}

\parnobf{Dataset Bias and Societal Impact} While the EPIC-KITCHENS videos were collected in 4 countries by participants from 10 nationalities, it is in no way representative of all kitchen-based activities globally, or even within the recorded countries.
Models trained on this dataset are thus expected to be exploratory, for research and investigation purposes.
 
We hope fine-grained understanding of hand-object interactions can contribute positively to assistive technologies, for all individuals alike including in industrial settings. Approaches for imitation learning are expected to benefit from VISOR. We hope future models will replace mundane and dangerous tasks.

\parnobf{Computational Resources Used}

Estimating the precise computational resources used over the course of a 22 month project is challenging. However, we give a sense of the computational requirements and briefly report information about the computational resources used by each component.

\parnoit{TORAS} TORAS uses a server with 2 GPUs to run the interactive segmentation interface, one per model.

\parnoit{VOS Baseline} For both traning and inference, we used a single Tesla V100 GPU. Training took 4 days on VISOR.

\parnoit{Dense Annotations} We use the model from VISOR baseline to extract dense annotations, and calculate their scores. This took 7 days using 9 Tesla V100 GPUs. Our code is designed to use 3 GPUs but we ran multiple parallel instances.

\parnoit{HOS Baseline} Training the PointRend model as described took 2 days on 2 A40 GPUs.

\parnoit{WDTCF Baseline} Most of the computational cost of WDTCF is in our PointRend model. Training the PointRend model as described took 2 days on 2 A40 GPUs.

\parnobf{Annotation Done}
We estimate that the total number of hours to annotate the data is around 25,000 hours - including human annotations, manual checks and corrections.
It is challenging to estimate correctly the resources used as the server is used for multiple projects.
The bulk of the annotation work was done in the pixel labeling. Freelance annotators who did this were paid hourly, and self-reported their hours via the Upwork platform. Hourly rate varied by experience and submitted proposal from \$6-\$9/hour, with the higher rate reserved for most experienced annotators who were also in charge of QA checks.
We believe annotators were fairly compensated, working directly with the freelance annotators. Small amounts of the data were annotated via crowdsourcing services. We report the cost spent in those sections.

\parnobf{Storage and Availability}
As with the EPIC-KITCHENS-100 videos, the VISOR annotations is permanently available with a unique DOI from the University of Bristol Data-Bris storage facilities at: \url{https://doi.org/10.5523/bris.2v6cgv1x04ol22qp9rm9x2j6a7}.
The data management policy for Data-Bris is available from \url{http://www.bristol.ac.uk/staff/researchers/data/writing-a-data-management-plan/}.
The University of Bristol is committed to storing, backing-up and maintaining the dataset for 20 years, in-line with UKRI funding requirements.

%

\clearpage
\section{Appendix - Entities, Frames, and Subsequences (Main \S2.1)}
\label{sec:app:stage1}

This stage identifies {\it what} will be annotated in terms of entities (i.e., names of the objects) and frames (i.e., which frames to annotate). The goal of this stage is to generate an {\bf overcomplete} set of entities which annotators will later select from as well as identify a list of frames that can be accurately annotated. We describe the preparation of entities (\S\ref{sec:entity_preparation}), selection of frames (\S\ref{sec:app:frames}), and examples of our subsequences (\S\ref{sec:app:subsequence}).

\subsection{Entity Preparation}
\label{sec:entity_preparation}
Entity preparation is the cornerstone of the entire project. Annotators refer to the entities list to provide manual pixel-level segmentation. Therefore, the quality of the entity list directly determines the quality of the proposed dataset. In order to ensure quality, we first automatically extract potential active entities from narration, then task workers on Amazon Mechanical Turk with creating a list of active entities.
Pixel-label annotators can choose to ignore an entity if absent, occluded, highly-blurred or incorrect altogether.

\parnobf{Entity Extraction} Entity candidates consist of a main-object list and three additional-object lists. A main-object list contains nouns appearing in the narration and additional-object lists include the commonly used tools. We then detail both lists.

\parnoit{Main list}
We extract entity candidates per short `action' clip in EPIC-KITCHENS-100 \cite{Damen2020RESCALING} to form the main list. This dataset provides not only the ground-truth of verb and noun for action recognition, but also the narration for each clip. Naturally, the objects in the narration are active. Hence, entities for one clip are extracted from the narration. 

\parnoit{Adjacent Clips}
However, these are not enough as we found that most active objects are not mentioned in the narration, (for instance `peel potato'). In this case, `peeler' is the active object but it does not appear in the narration. Thanks to the density of the annotation of EPIC-KITCHENS-100 \cite{Damen2020RESCALING}, some missing objects are likely to be found in previous or latter actions, for example `take peeler' before `peel potato'. Therefore, the candidate entities for one clip are formed by merging the entities from the current clip, the previous four clips, and the ensuing two clips. Moreover, dense annotation of EPIC-KITCHENS-100 \cite{Damen2020RESCALING} might cause overlapping clips, meaning some frames belong to both clips. This overlap might lead to replicated work for annotators. To solve this issue, we generate the candidate entities for non-overlapping clips. 

\parnoit{Common Tools}
Although looking at adjacent portions of the video and narration can help us find missing active objects, it still has the limitation of missing objects mentioned in far away clips or even not mentioned at all. For example, in `cut celery', the `chopping board' never appears in narrations but it indeed is an active object. To tackle this problem, we propose three additional-object lists which include the high-frequency objects in terms of cutlery, utensils, as well as furniture. They contain [`fork', `knife', `spatula', `spoon', `other cutlery'], [`chopping board', `bowl', `plate', `cup', `glass', `pan', `pot'], and [`cupboard', `drawer', `tap', `drainer', `hob', `bin/garbage can/recycling bin', `fridge', `oven', `sink'], respectively.

\parnoit{Americanisms}
Finally, 34 British entity names are translated to American alternatives. Thus, we provide a main-object list and three additional-object lists to workers in Amazon Mechanical Turk to select the proper active entities.

\begin{figure}[t]
    \centering
    \includegraphics[width=1.0\textwidth]{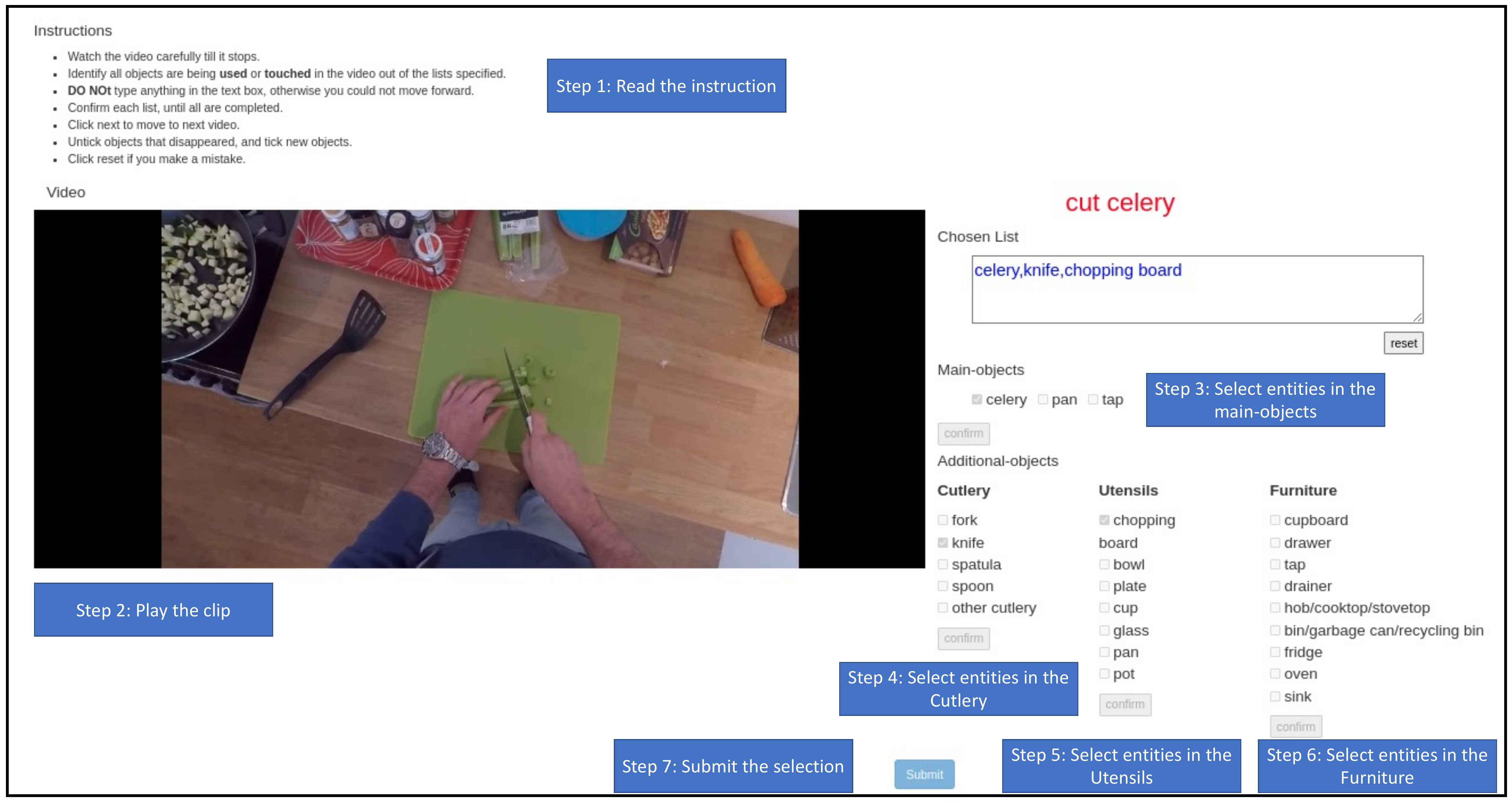}
    \caption{{\bf AMT interface for entities selection.} This is used to identify {\it which} objects will be given pixel annotations.}
    \label{fig:object_selection}
\end{figure}

\parnobf{Entity Selection} The proposed method for entity extraction brings some irrelevant entities. We design an interface for workers to select the proper active entities. The interface, shown in Fig.~\ref{fig:object_selection}, is composed of six parts. First of all, it illustrates instructions to guide workers to complete the work step by step. Below the instructions, there is a clip introducing the action which is highlighted in red on the right. The ``Chosen List'' shows the ticked selection and the ``reset'' button clears all selections. Next, the interface shows our prepared entities. Workers then select the entities from main-object and additional-object lists step by step while referring to the clip. All the selections are collected with the ``Submit'' button.

Additionally, in order to improve the quality, we design some functions for the interface. Specifically, (1) some entities selected in the previous clip will be automatically ticked in the current clip. We notice that a clip is very likely to inherit objects from the previous clip due to the continuity of the video and dense annotations of EPIC-KITCHENS-100 \cite{Damen2020RESCALING}, e.g., ``take bowl'' and then ``wash bowl''. If the ticked entities are irrelevant to the action, the ``reset'' button can help to clear all selections and workers are able to restart selecting entities. (2) The interface requires workers to watch the clip first before choosing, otherwise they cannot proceed. Also, 3) it asks the worker to select candidates one by one in order by clicking on different ``confirm'' options. (4) If an entity appears in the main-object list, it will be removed in additional lists.

We use Amazon Mechanical Turk to complete entities selection. At the very beginning, we set five different workers to complete one sequence composed of 16 consecutive clips. The reward per sequence is \$0.18. We merge all selections using majority decision. Later, we decreased the number of workers to two as we learnt from current examples and introduced our correction stage described next.
When a word is chosen by 4/5 or 2/2 Amazon Mechanical Turk workers, we consider it as an active entity. Additionally, we propose two rules for merging selections. One is verb-entity, which introduces new entities based on the verb used, e.g., `wash' in `sink', `stir' on `hob'. Another one is entity-entity, e.g., `tap' with `water' in `sink' for instance. 
By combining the selections from different workers, we acquire our over-complete list.

\begin{figure}[h]
    \centering
    \includegraphics[width=1.0\textwidth]{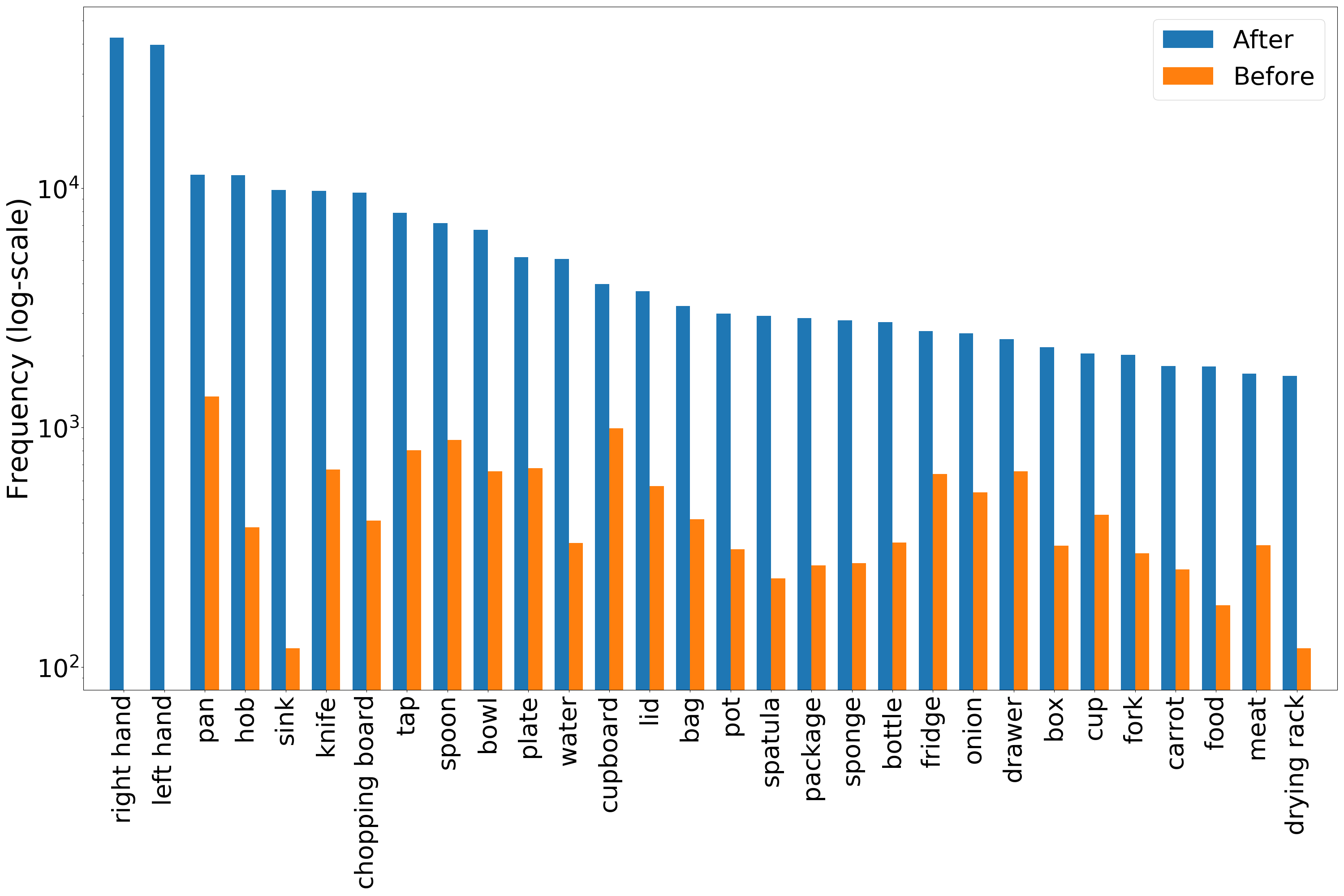}
    \caption{{\bf Changes in the number of entity classes before and after entity preparation.} Entity classes are ordered by frequency of occurrence. Our entity preparation is critical.}
    \label{fig:changes_stage1}
\end{figure}

Entity preparation matters. To compare the difference between before and after entity preparation, we visualise the changes in the number of entity classes in Fig.~\ref{fig:changes_stage1}. 
The difference between orange and blue entities shows the additional entities found 
through entity preparation. Additionally, it substantially increases the number of entity classes that appears before while bringing right and left hands. Note that the occurrence of hands is reasonable given our egocentric videos. The number of ``sink'' entities is increased the most due to the prevalence of the washing actions.

\subsection{Frame Extraction}
\label{sec:app:frames}
For the majority of videos in VISOR, we use an approach of variable frame rate for frame selection. We sample 6 frames per subsequence, where a subsequence is composed of 3 nonoverlapping actions.
At the end of the video, when only 1 or 2 actions remain, we sample less frames accordingly.

For only 7 videos in VISOR, we attempt a fixed frame rate in line with other datasets. 
Table ~\ref{tab:frame_rate_stats} shows the statistics of each variation.

\begin{table}[t]
\centering 
\caption{{\bf Frame rate variations statistics}. Variable frame rate used in the majority of the videos. We compare these to fixed size sampling at 2fps, collected for 4 videos. 
}
\label{tab:frame_rate_stats}
\resizebox{\textwidth}{!}{
\begin{tabular}{llllllll}
\toprule
                   & avg. rate &videos & seq& images & masks   & \%hands &max frames/seq  \\
                   \midrule
                   Variable frame rate & 0.9fps& 172  & 7552         & 44,777 & 239,357 & 30.0\%  & 6\\
Fixed frame rate   & 2fps &7     & 308          & 5,952  & 32,227  & 32.4\% & 186  \\
 \bottomrule
\end{tabular}}
\end{table}

\begin{figure*}[t]
    \centering
    \includegraphics[width=1.0\textwidth]{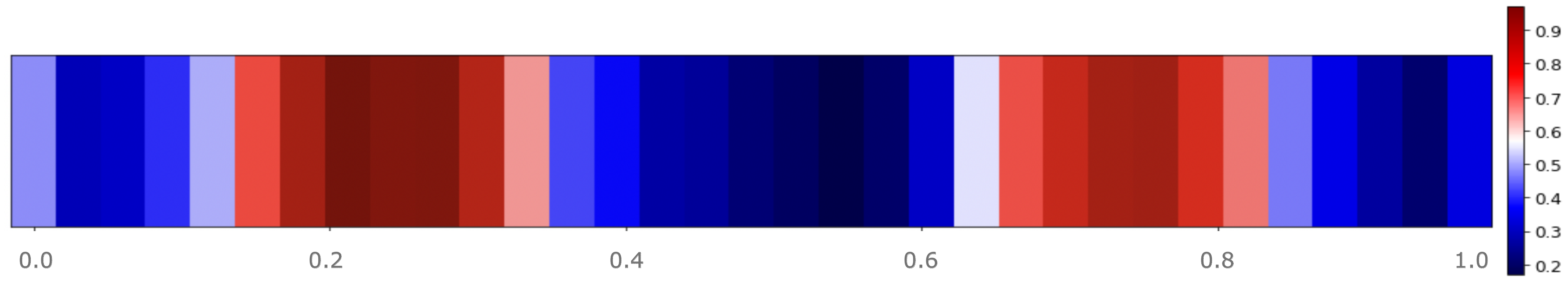}
    \caption{{\bf Distribution of the frames per action}. we considered frames within actions only.}
    \label{fig:frames_per_action}
\end{figure*}

\begin{figure*}[t]
    \centering
    \includegraphics[width=1.0\textwidth]{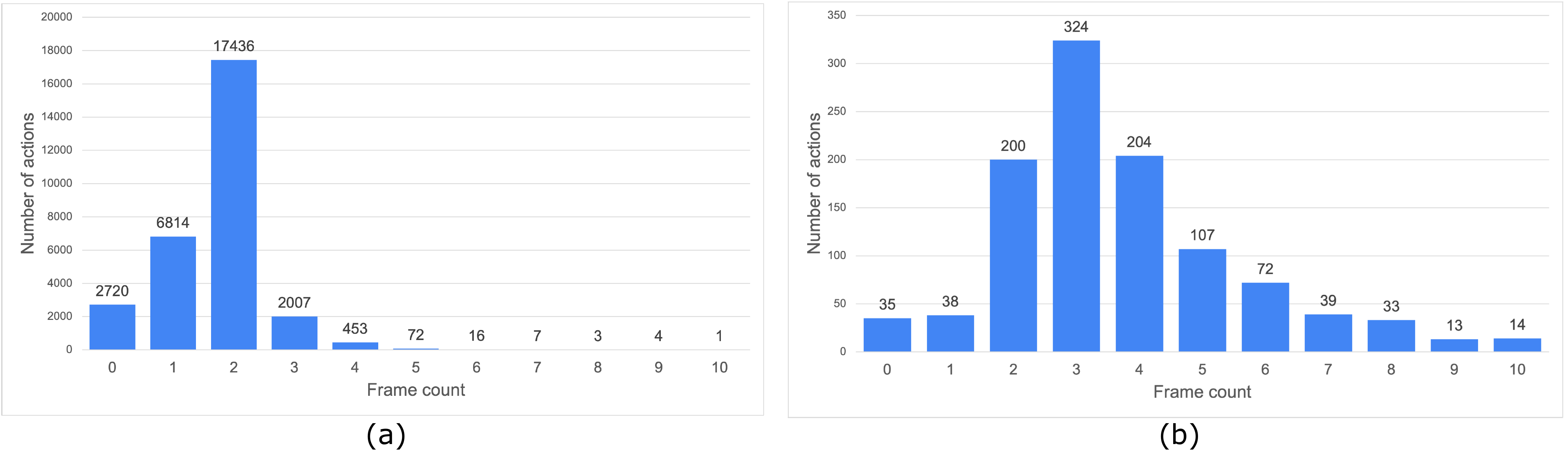}
    \caption{{\bf Distribution of the actions by each frame count.} (a) variable frame rate; (b) fixed frame rate. Both distributions show the first 10 frame counts.}
    \label{fig:actions_per_frame_count}
\end{figure*}

\parnobfq{Why use a variable frame rate} A variable frame rate is useful to: (1) concentrate the sampled frames to be within actions (rather than between actions); this concentration leads to more non-hand objects as shown in Table~\ref{tab:frame_rate_stats}. Concentrating in this way is helpful in all of our benchmarks, as 3.7\% of the frames are between actions in the variable frame rate scheme whereas 10.2\% are between actions when using a fixed frame rate. (2) Save annotation time to include more videos. (3) potentially add to the sequence difficulty of video-based benchmarks such as semi-supervised VOS as there is no count limit between the annotated frames.

\parnobfq{Why also include a fixed frame rate} A fixed frame rate is the standard in other benchmarks. We decided to collect videos this way so as to enable researchers to appreciate the difference between the two regimes. This also allows denser evaluations, and so 4 of our 7 fixed frame rate videos are left for the Test set.
We release 3 videos annotated at 2fps in Train/Val.

\parnobfq{How are the 6 frames per subsequence sampled in the variable frame rate} We sample 2 frames per action at 25\% and 75\% completion of the sequence, then we apply random frame shift of ${\pm 10 \%}$. 
Our choice of 2 frames per action allows annotating the transformation of objects within the action when present.
Next, we apply a Laplacian filter with a window of 20 frames (${\pm 10}$) to select the frame with the lowest motion blur (lowest variance).
Finally, to avoid frames with similar appearance we check if any 2 frames are closer than 25 frames (	${\sim}0.5$ seconds). If so we re-sample one of them to be in the middle of the farthest 2 frames in the subsequence. This enriches the subsequence's temporal information.

\parnobfq{How are the frames distributed in the actions} Fig. ~\ref{fig:frames_per_action} shows the actions' frame distribution of the whole dataset, the selected frames are concentrated around 25\% and 75\% of the way through the action's length.

\parnobfq{How many frames do most actions have}
Fig. ~\ref{fig:actions_per_frame_count} shows the distribution of the number of frame counts per action, comparing the variable frame rate and fixed frame rate strategies. Most actions in the variable frame regime have 1-2 sampled frames. There are some actions (2720) without any annotated frames.
These are typically short actions that overlap with other actions.
As our subsequence focuses on non-overlapping actions, these overlapping actions might not have frames sampled within their boundaries.
For the fixed frame rate, most actions have 2-4 sampled frames which is significantly higher than the variable frame rate. There is also a long tail of many frames that have many, many frames (e..g, 171 actions have six or more frames). This means that much of the annotation budget is spent on annotating the same action repeatedly.

\subsection{Subsequence Examples}
\label{sec:app:subsequence}

\begin{figure*}[t]
    \centering
    \includegraphics[width=1.0\textwidth]{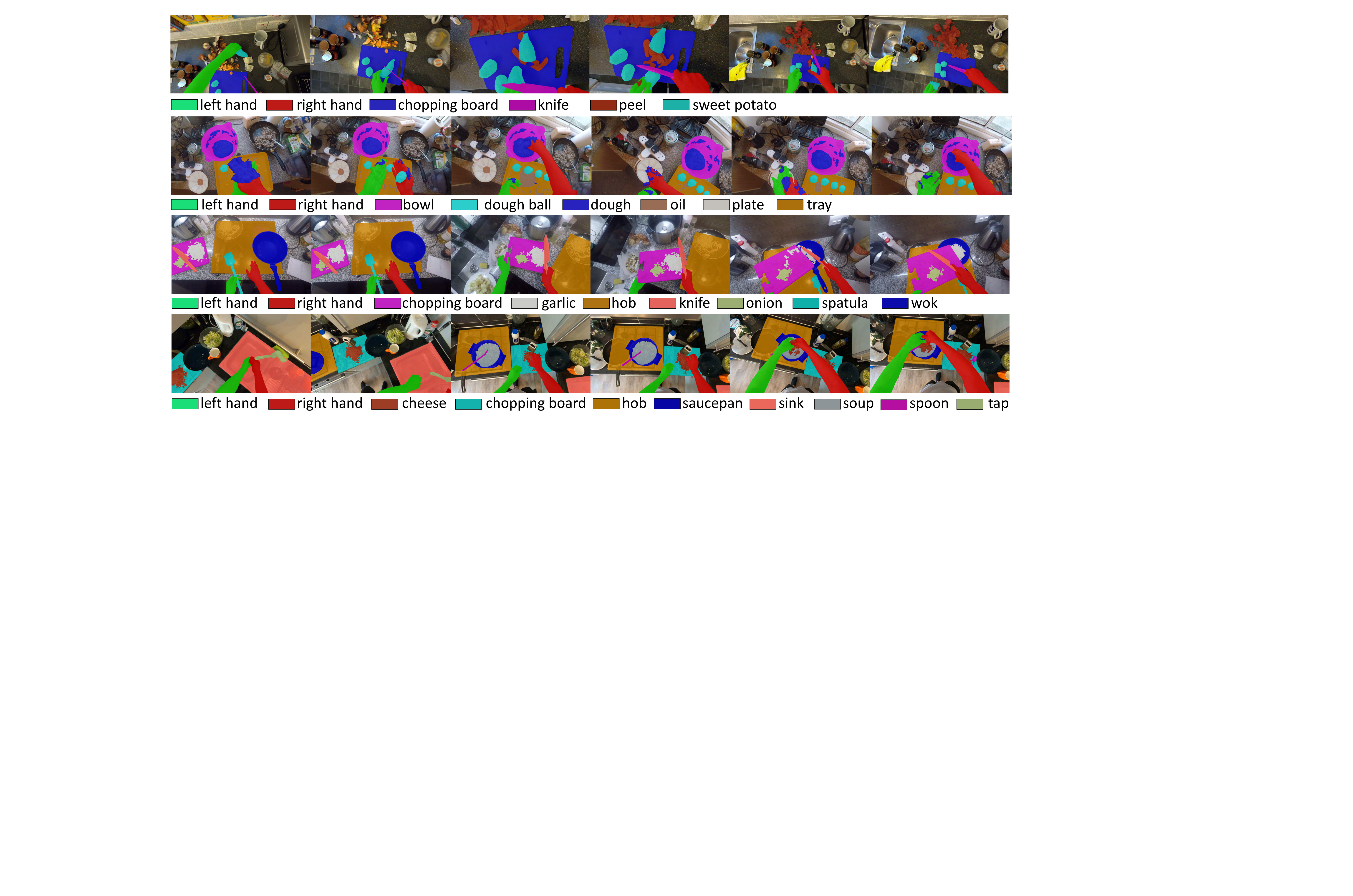}
    \caption{\textbf{Four subsequences from different videos.} In each row, we plot overlays from six consecutively annotated images, with temporally consistent segmentations (legend). Actions are `peel sweet potato', `make dough ball', `pour garlic' and `put cheese', respectively.}
    \label{fig:sequences}
\end{figure*}

In Fig~\ref{fig:sequences}, we show 4 subsequences from various videos. As noted in the main paper, we use the term `subsequence' to refer to the 6 frames from 3 consecutive non-overlapping action labels in EPIC-KITCHENS. 
These have a consistent set of entities throughout.
As shown in the figure, entities are temporally consistent over these short-term subsequences, demonstrated by consistent legend under each subsequence.

\clearpage
\section{Appendix - Annotation rules and annotator training (Main \S2.2) / Annotator Training and Rules}
\label{sec:app:stage23}

\begin{figure}[t]
    \includegraphics[width=1.0\textwidth]{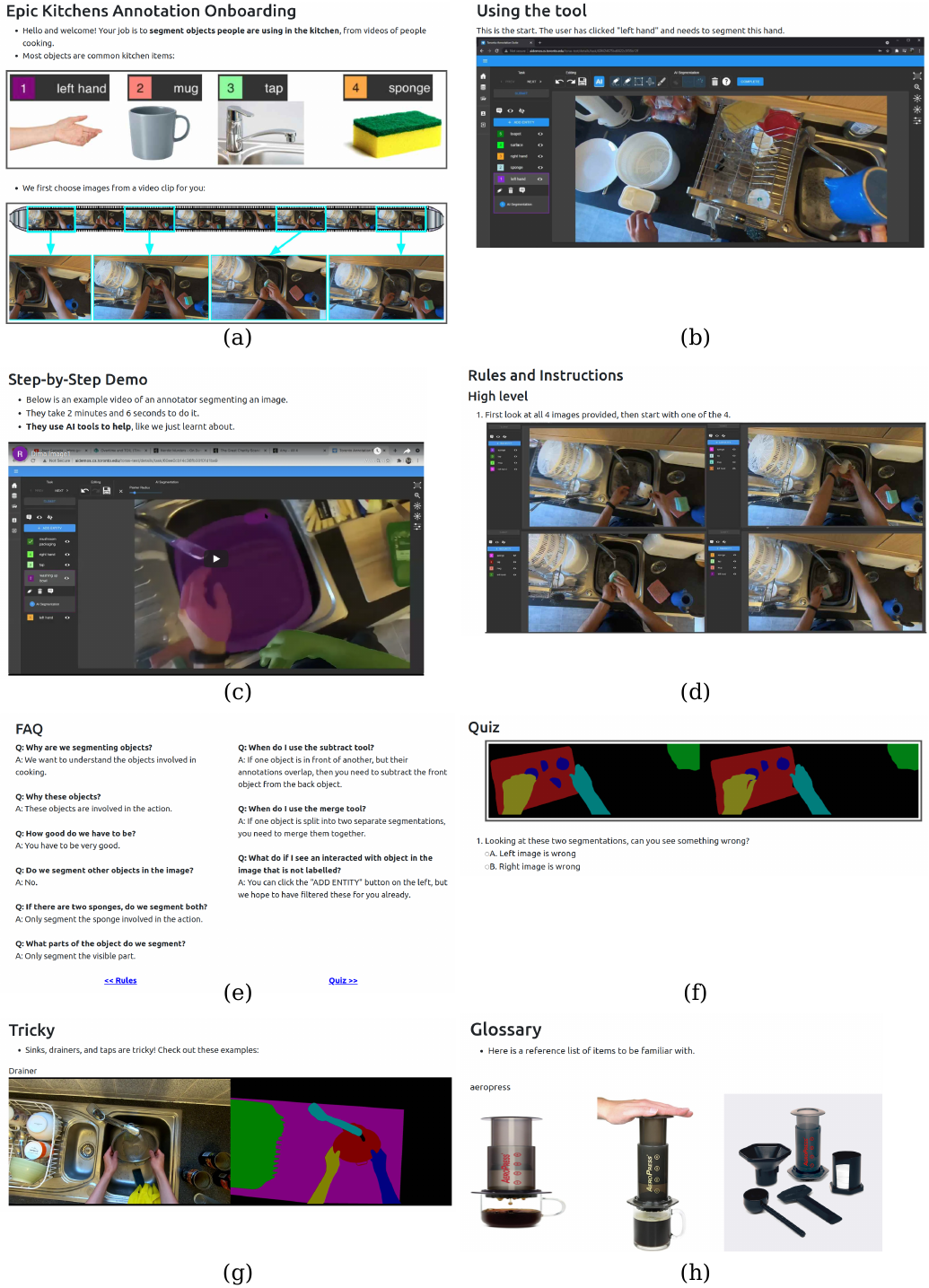}
    \caption{{\bf A page-by-page breakdown of the onboarding process.} (a) A front-page that introduces the project. (b) An overview on using the tool. (c) A step-by-step fine-grained walkthrough explaining everything that happens in the example annotation video. (d) The complete rule handbook with examples and instructions per rule. (e) An FAQ for tricky situations and general information on the project. (f) A quiz to test for learning. (g) Selected challenging situations that arise when annotating. (h) A glossary of items in the dataset, including example images for each class.}\label{fig:onboarding_frontpage}
\end{figure}

Once entities and frames have been identified,  we annotate the frames with the labels. A key step is the training of annotators. We now describe their recruitment (\S\ref{sec:app:recruiting}), training (\S\ref{sec:app:training_material}), and the rules that they followed (\S\ref{sec:app:rules}).

\subsection{Annotator Recruiting}
\label{sec:app:recruiting}

We hired annotators via Upwork due to its convenient system for communicating with annotators. We started with a large pool of freelancers, with whom we shared a document explaining the AI tool. The freelancers then annotated a common set of images and we encouraged them to use the AI tool. We 
then selected a smaller subset based on annotation speed and quality, and quick adaptation to use of the tool.

\subsection{Training Material}
\label{sec:app:training_material}

To train annotators, we designed an onboarding website (with salient pages shown in Fig.~\ref{fig:onboarding_frontpage}).
The onboarding website functions like a tutorial.
The main pages of the tutorial are listed next:
\begin{enumerate}[leftmargin=16pt]
\item an overview of the project and the role of annotators, so trainees can get familiar with the task.
\item how to use the AI annotation tool, specifically the functionalities of the various buttons in the interface.
\item step-by-step annotation of a single sequence, where every action taken across the course of the video is explained in detail.
\item a breakdown of the annotation rules, highlighting example images where each of the rules applies, and how one would segment objects in relevant situations.
\item an FAQ, with general questions as well as an explanation for some confusing corner cases encountered with the tool.
\item a quiz to test annotators for understanding.
\item some tricky/difficult annotation examples, and how these should be segmented.
\item a glossary of all the object classes annotators need to be familiar with, as well as examples for each class.
We opted to include the glossary to avoid any guessing from non-native speakers on what a `spatula' or `sieve' are.
The glossary is populated via a Google Image search, where for every object we request to be labelled in the dataset, we display the top-3 images that a Google search for that object returns. 
\end{enumerate}

\subsection{Rules}
\label{sec:app:rules}

The rules are designed to maximise consistency and minimise errors when using the AI automatic annotation tool. 
Workers follow rules about {\it how} to segment to ensure temporal consistency within a video and {\it what} to segment to ensure consistency across the dataset.

\parnobf{How to Segment} As segmentations are only provided for certain in-contact objects and objects that occlude them, the first instruction is to consider the six frames that make up a subsequence, and the objects requested for segmentation, in order to identify which objects should be segmented. Next, the user is encouraged to proceed frame by frame. Within a frame, they segment objects visible in the frame if present. They also mark objects that are listed but are invisible in that frame. 
After segmentations for a frame are complete, they should press the submit button. 

\begin{figure}[h!]
    \centering
    \includegraphics[width=1.0\textwidth]{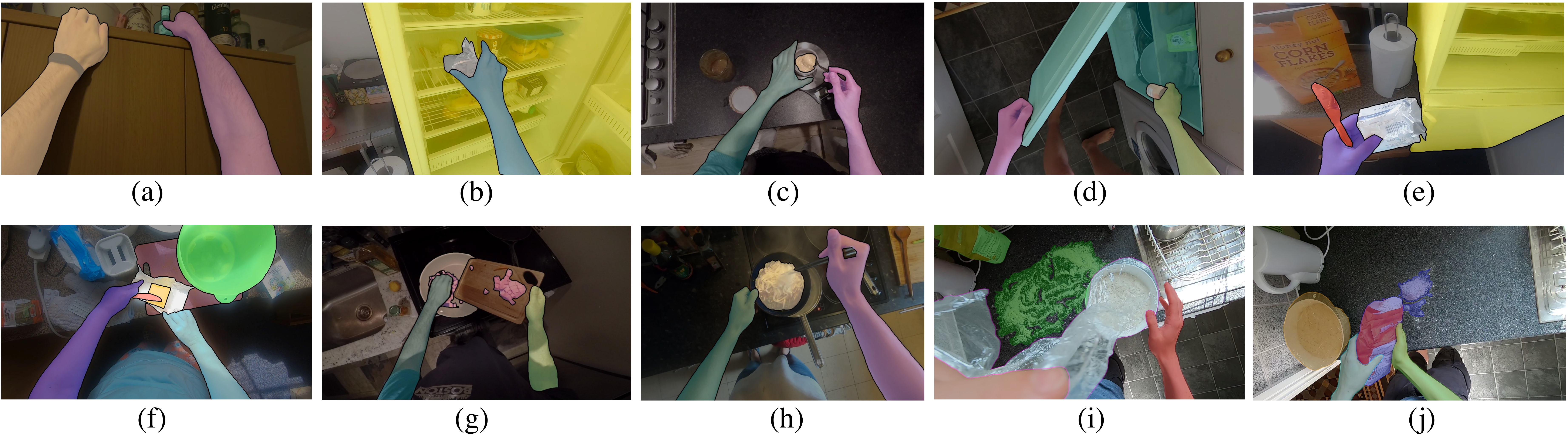}
    \caption{{\bf Demonstrations of rules for hands and active objects segmentation.} \\ 
    - {\bf (a)} For left and right hands, segmentations include all visible parts of the hand and arms. For consistency, we requested that annotators include watches or wristbands on both hands.\\
    - {\bf (b)} For situations where a container should be segmented but objects internal to the container are not requested for segmentation, we request they segment the entire container, e.g., all of a fridge. If an internal object is requested and visible, the user should separately segment the visible parts of both, e.g., the butter package and the fridge in (b). \\
    - The most tricky cases are to segment the contained objects. Specifically, {\bf (c)} if the object is contained within another (but still partially or fully visible), we ask the user to only segment the visible parts of the contained object and ignore the container. This allows us to segment both the `paste' and the `jar' separately in this example. \\ 
    - Instead, {\bf (d)} if an object is fully contained within another (and thus invisible) e.g., ‘salt’ inside a bottle, we ask the user to segment the whole container, i.e. the `bottle' as the `salt' in this case. In this case, when the user is retrieving the `salt' from the cupboard, they are in fact segmenting the salt bottle which has salt inside.\\
    - {\bf (e)} If objects are fully contained but invisible, and packaging is requested, we ask the user to segment only the packaging, e.g., the butter package. \\
    - Next, {\bf (f)} if partial occlusions occur, we instruct the user to only separate the visible parts of the occluded objects. For example, the knife is above the butter which is in the package. \\
    - {\bf (g)} For objects with several small pieces, e.g., the pieces of carrots, we instruct users to segment each piece individually and merge the segmentations into one. \\
    - However, {\bf (h)} we mention that if objects are truly tiny and overlapping, e.g., noodles, they can have their internal borders ignored. 
    - {\bf (i)} If a transparent object, e.g., packaging, can be segmented, we instruct users to segment it. \\
    - Finally, {\bf (j)} we tell users that they can ignore this rule if the packaging is extremely transparent, thus implausible to segment, and the contained object is much more visible to segment accurately. }
    \label{fig:rules_visualisation}
\end{figure}

\parnobf{What to Segment} After this overview of ``how to segment'', we next explain ``what to segment'' with Fig.~\ref{fig:rules_visualisation}. We explain each of the rules in the caption as these correspond to a single example in the figure.

\clearpage
\section{Appendix -- Tooling: The TORonto Annotation Suite (Main \S2.2) } 
\label{sec:app:toras}

Annotating our dataset is made substantially easier by the use of an AI-assisted annotation tool, the TORonto Annotation Suite (TORAS)~\cite{torontoannotsuite}. 
This section summarises 
the features and corresponding interfaces that were used in annotating the segmentation masks for EPIC-KITCHENS VISOR. 
The TORAS interface has been assessed and approved by the University of Toronto ethics review committee.

The TORAS interface was used to annotate segmentation masks for a fixed list of objects per image derived from entity identification. Later, the interface was used by annotators to correct segmentations and add new objects by incorporating comments collected after a round of corrections. Specifically, we discuss the segmentation tools, including AI enabled tools made available to annotators (\S\ref{sec:app:tools}), project management features (\S\ref{sec:app:toras-proj-mgmt}), report annotator agreement statistics (\S\ref{sec:app:toras-agreement}) and discuss annotation efficiency(\S\ref{sec:app:toras-efficiency}).

\subsection{Segmentation Tools}
\label{sec:app:tools}

In this section, we describe the TORAS tools used to create and edit segmentation masks. Note that we always use polygons with floating point coordinates as the base representation to ensure high precision segmentations can be created. We also note that annotators have the ability to not use the AI through a toggle button. On beginning annotation of a task, annotators see on the left a list of objects to annotate, pre-populated using entity identification annotation outputs. Annotators can select any object at any time and continue annotating it.
The annotators can choose to lock a segmentation upon completion, as well as toggle its visibility.

\label{sec:toras-ai}
\parnobf{AI Box Segmentation} The box segmentation tool expects a user to draw a tight bounding box around an object of interest, where the algorithm (based on ~\cite{CurveGCN2019, DELSE2019}) outputs a single closed polygon that encloses the salient object within the bounding box. While it is possible to generate multiple polygons that better represent the object, deal with holes etc., a more predictable tool is easier to interact with in our experience.

\parnobf{AI Trace Segmentation}
This tools expects a user to scribble a quick and coarse boundary around the object, where the algorithm outputs a single closed polygon that captures fine details and "snaps" to the boundaries of the object of interest.

\parnobf{AI Correction}
When a user corrects the position of a single vertex in a polygon, we algorithmically predict offsets for vertices in a local neighbourhood of the corrected vertex. Once a vertex is corrected by a human, it is marked such that subsequent AI interactions do not displace it.

\parnobf{AI Refinement}
The refinement tool allows "snapping" any polygon to better fit it to object boundaries.

\parnobf{Path Correction}
For correcting large errors, users can draw a new polyline or scribble connecting any two vertices of an existing polygon. The newly drawn polyline or scribble is "snapped" to boundaries using the same algorithm as in AI Refinement. 

\parnobf{Paint/Erase}
This tool allows users to draw freeform segmentations using a circular brush with adjustable radius. This freeform segmentation can be used to draw a new polygon, add to an existing polygon or subtract from an existing polygon.

\parnobf{Segmentation Booleans}
Segmentation booleans have multiple uses. Any polygon can be subtracted from or merged with any other polygon. This allows annotators to draw holes, reuse precise boundaries drawn for a different object adjacent to the current object of interest, and deal with occlusions faster.

\subsection{Project Management}
\label{sec:app:toras-proj-mgmt}
Managing annotations for a large dataset requires scalable methods to assign, monitor and review annotation tasks. For this purpose, we extensively use the python API accompanying TORAS. All task management was done using python scripts and progress was monitored through a slack bot in a shared workspace. Comments obtained from stage3 were sent back to annotators using the API, which are shown clearly in the task UI. Annotators were able to see whether a task has a comment using a special indicator in their list of annotation tasks.
\newline

\subsection{Annotator Agreement}
\label{sec:app:toras-agreement}
To verify annotation quality, five of our annotators annotated the same set of 186 images independently. These 186 images had 1422 total objects listed for annotation, obtained from our over-complete entity identification stage. Out of the 1422 objects, 940 objects were annotated by all five annotators. We compute segmentation agreement on these 940 objects in Table~\ref{tab:toras-annotator-agreement} and report 90.3 average pairwise mean IoU between all annotators (i.e. averaged from mean IoU of 10 combinations of two annotators from our group of five). This is in line with the 90 IoU human agreement per instance reported for OpenImages~\cite{benenson2019large} and much higher than the ${\approx}$80 IoU agreement reported for MS-COCO polygons in the OpenImages paper~\cite{benenson2019large}. 

Out of the 1422 objects, 940 were annotated by all five annotators. Out of the 482 not annotated by all, 295 were left empty by all annotators i.e. they consistently agreed that the object does not appear in the image. 
Thus, 13\% (187 out of 1422) of the total objects in this subset were inconsistently annotated. 
Of the 187 remaining objects, 97 were annotated by four (out of five) annotators, 22 by three, 30 by two and 38 by one annotator.  We note that our subsequent correction annotation aims to fix these errors of missed objects (among other errors). 

\subsection{Annotation Efficiency}
\label{sec:app:toras-efficiency}
We measured annotation speed-up using our AI tools over manual annotation, by annotating a subset of 50 images from different videos, containing 253 objects. One of our trained annotators annotated these 50 images with all of our tools (including AI tools) on one day, and manually on another day. Only using manual tools resulted in an average time of 50.5 seconds / object, while obtaining 89.92 IoU agreement with our released ground truth. With access to AI tools, annotation time went down to an average of 37.5 seconds / object, i.e. at approximately three quarters of the time, while obtaining 90.11 IoU agreement. The IoUs obtained are within one standard deviation of each other. Thus, on this subset, we verify that there is no accuracy degradation with using AI tools for segmentation, while saving a quarter of the time over manual segmentation. We also find that our annotators got faster through the project, showing up to 2x speed improvement in number of images annotated per hour.

\begin{table}[]
    \centering
    \caption{{\bf Inter-annotator agreement.} We report mean intersection-over-union (IoU) of segmentations from five different annotators on a subset of 186 images with 940 objects. Our results show a high average inter-annotator agreement of 90.3 IoU, consistent with OpenImages~\cite{benenson2019large}}
    \label{tab:toras-annotator-agreement}
    \begin{tabular}{l|ccccc} \toprule
         & Ann. 1 & Ann. 2 & Ann. 3 & Ann. 4 & Ann. 5  \\ \midrule
         Ann. 1 & ~ & 88.81 & 89.51 & 90.39 & 90.76 \\
         Ann. 2 & 88.81 & ~ & 89.56 & 89.28 & 90.45 \\
         Ann. 3 & 89.51 & 89.56 & ~ & 90.98 & 92.30 \\
         Ann. 4 & 90.39 & 89.28 & 90.98 & ~ & 91.49 \\
         Ann. 5 & 90.76 & 90.45 & 92.30 & 91.49 & ~\\ \bottomrule
    \end{tabular}
\end{table}

\clearpage

\section{Appendix -- Correction (Main \S2.2) / Correction}
\label{sec:app:correction}

The goal of our correction stage covering all previous annotations is to make sure the masks are accurate within the current frame and sequence. We plot out all annotations on images and then manually scroll through the frames, in order to confirm/correct consistency. We describe our interface~(\S\ref{sec:app:correctioninterface}), how the corrections are performed on TORAS 
(\S\ref{sec:app:correctionontoras}), and statistics about the corrections (\S\ref{sec:app:correctionstatistics}).

\subsection{Correction Interface for Collecting Comments}
\label{sec:app:correctioninterface}

\begin{figure}[t]
    \centering
    \includegraphics[width=1.0\textwidth]{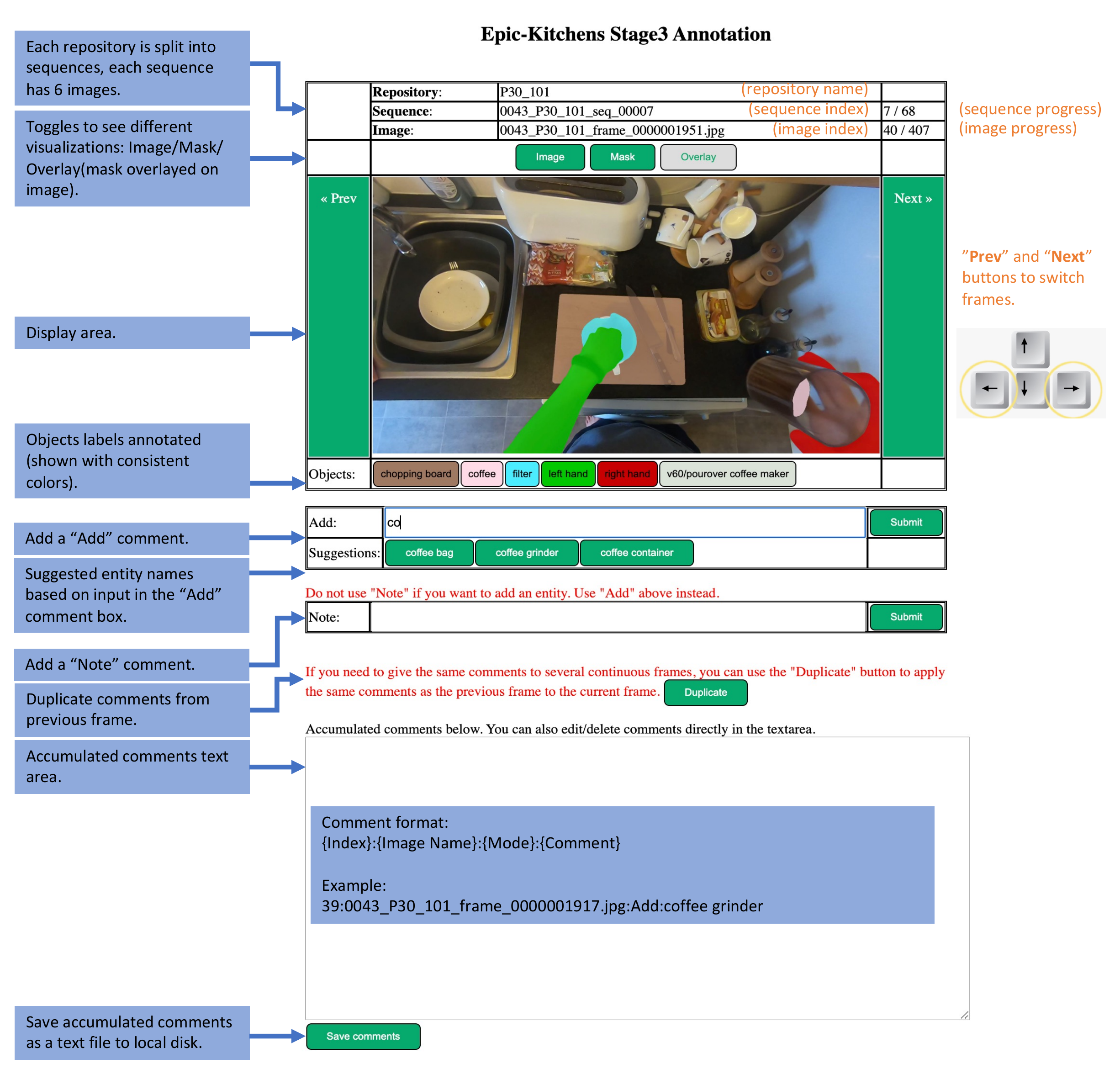}
    \caption{{\bf Correction stage user interface for writing comments for each image}. For each annotation repository (i.e., sequence of frames given to the annotator), we generate a HTML page to show its annotation from the pixel labels.}
    \label{fig:stage3_instructions}
\end{figure}

\begin{figure}[t]
    \centering
    \includegraphics[width=1.0\textwidth]{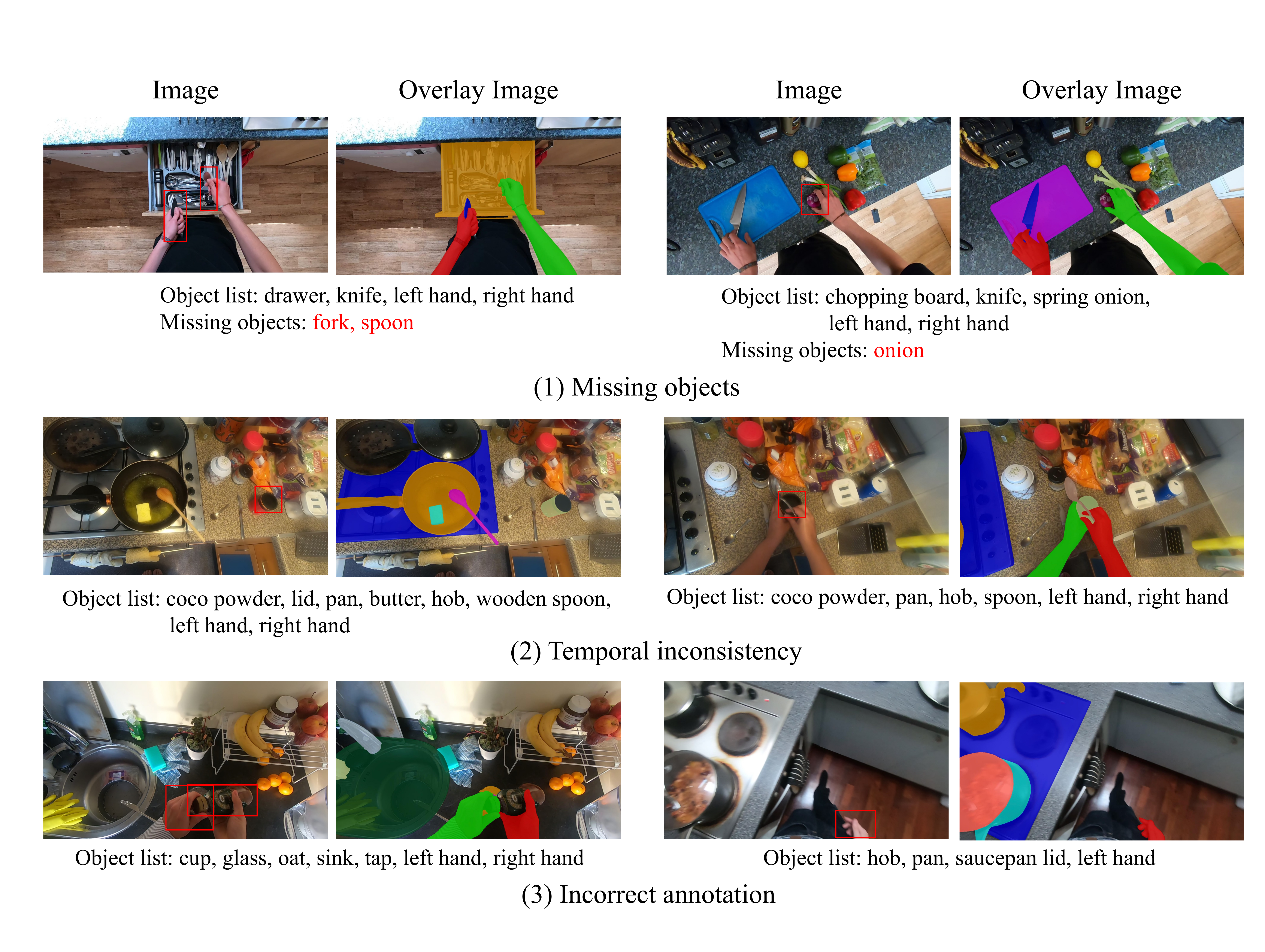}
    \caption{{\bf Typical errors that are fixed in the correction stage}. We show three kinds: missing objects, temporal inconsistency, and incorrect annotation.}
    \label{fig:stage3_type_error}
\end{figure}

The goal of our correction interface is to spot annotation errors in pixel-level segmentations. First, we ask commenters to give comments for each frame using the interface in Fig.~\ref{fig:stage3_instructions}. The interface contains three blocks, display block, commenting block and comments block.

\parnobf{Display} On the top, the display block shows information about the current repository, sequence and image as well as indicating the sequence progress and image progress in text to let the commenters know their progress. In the middle, the current image/mask/overlay(masks plotted on the image) as displayed in the center, with toggle button to choose which mode (Image/Mask/Overlay) to show. With Overlay set as the default toggle, the Image/Mask display is very helpful when the overlaid annotation makes it hard to check some details.
The prev/next button is to switch between images which can also done by using the left/right key on the keyboard, enabling swift switching. On the bottom, the object list shows the entity names highlighted with corresponding colors as in the overlaid display.

\parnobf{Add/Note/Duplicate} We invite the corrector to provide any of three types of corrections. First, the ``Add'' mode allows choosing from the suggested entity names by clicking. There is a matching algorithm behind this suggestion technique. Candidate entities come from other entities in the same video. 
New entity names can also entered. The second ``Note'' mode is used mainly for submitting comments about renaming (e.g., ``rename paper to carton'', ``rename mixture to pancake'') and correcting existing annotations (e.g., ``missing part of the sink segmentation'', ``table is incorrectly segmented (missing corner)''). The last one is ``Duplicate'' which copies comments from the previous frame to the current frame, as some errors persist in more than a single image, thus accelerating the correction stage.

The comments are aggregated in the text box listed in reverse order. Once the repository is done, the comments can be saved as a text file locally.

\subsection{Correction on TORAS}
\label{sec:app:correctionontoras}

We upload comments to TORAS for annotators to do correction for all the repositories. The comments are shown for each frame and the annotators are asked to correct and resubmit images.

For the Train and Val sets, we typically used a single round of corrections, except where a video was marked for a double-round of corrections.
For the Test set, all videos passed through two rounds of corrections.

\subsection{Statistics of Collected Comments}
\label{sec:app:correctionstatistics}

Fig.~\ref{fig:stage3_type_error} shows three typical kinds of errors detected in this correction stage, namely missing objects, temporal inconsistencies and incorrect annotations. First, missing objects are the most common error, which occur when some active objects do not exist in the object list for annotation. Two examples of missing objects are presented in Fig.~\ref{fig:stage3_type_error} (1). In the left example, ``fork'' and ``spoon'' are missing while ``onion'' is missing in the right example. Second, temporal inconsistencies refer to annotations in consecutive frames or sequences that are not consistent across sequences. In Fig.~\ref{fig:stage3_type_error} (2), the ``lid'' of ``coco powder'' is annotated in the left frame, but missing from the right. If the two frames are examined independently and temporal consistency is not taken into account, both annotations can be regarded as correct. 
Third, incorrect annotations, such as Fig.~\ref{fig:stage3_type_error} (3) left, show the ``lid'' of the ``syrup bottle'' segmented as part of the ``left hand'', with right showing the ``right hand'' annotated as ``left hand''.

Regarding the statistics on the number of images that were passed back for corrections, we report that 12,442 images out of our 50.7K images were passed back with at least one comment. The proportion of images that needed correction is thus 24.5\%. The proportion reported above is of all the images. On average, images that required correction had 1.8 comments from the manual verification stage. Of these, 80\% of the comments requested an addition (or renaming) of an active object, while 20\% of the comments were related to refining the segmentation boundary.

\clearpage
\section{Appendix - VISOR Object Relations and Entities (Main \S2.3)}
\label{sec:app:stage4}

Our object relations and exhaustive annotations are obtained from a crowdsourcing company (Hive aka \href{https://thehive.ai/}{thehive.ai}). This section describes the quality controls done by this platform (\S\ref{sec:app:hiveqc}) as well as the annotation instructions for Hand Object Segmentation (\S\ref{sec:app:hosannotation}) and Exhaustive Annotation (\S\ref{sec:app:exhaustiveannot}). We hired Hive to provide the annotation, and they in turn hired annotators, so translating directly into an hourly rate is difficult. However, we paid \$20 per thousand instances labelled for both tasks.

\subsection{Quality Control}
\label{sec:app:hiveqc}

Hive implements standard quality control during the annotation process consisting of qualifiers (tests before annotation), gold standard checks (tests during annotation), and consensus labelling (aggregation of multiple judgements).

\parnobf{Qualifiers} Before workers start a task, they are asked to complete a qualifier. This qualifier explains the annotation task and administers a test that workers must pass before they start the process. For all annotations, we required workers to achieve at least an 80\% on the qualifier test in order to start annotating.

\parnobf{Gold Standard Checks} Throughout the annotation process, workers are shown questions with known answers (usually clearcut cases). Workers that do not perform accurately on these known cases are dismissed. This helps catch guess-work efforts. For all annotations, we required workers to maintain a 90\% accuracy rate on known samples to continue annotating.

\parnobf{Annotator Consensus} Each instance was annotated by multiple annotators. A data instance was only considered labelled if six out of (up to) nine workers agreed on its label. Data in which this consensus could not be obtained were marked as inconclusive.

\subsection{Annotation Instructions for Hand-Object Segment Relations}
\label{sec:app:hosannotation}

\begin{figure}[t]
\begin{tabular}{cc}
\includegraphics[width=0.5\linewidth]{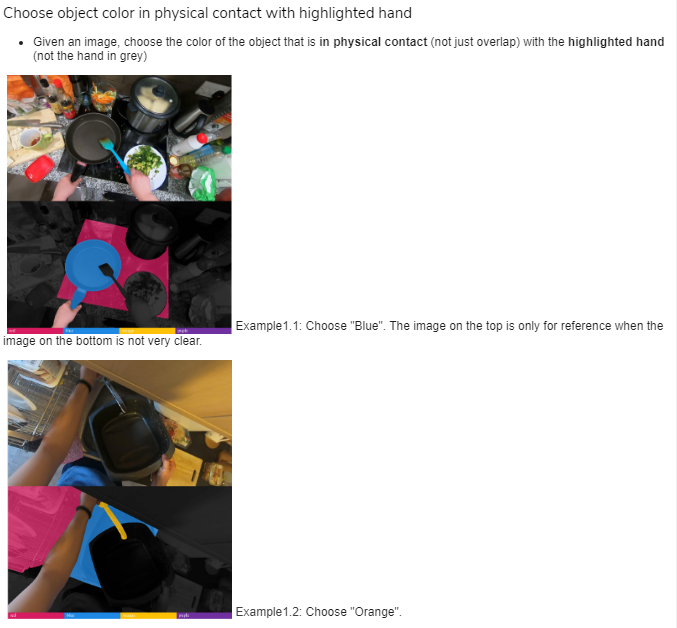} & 
\includegraphics[width=0.5\linewidth]{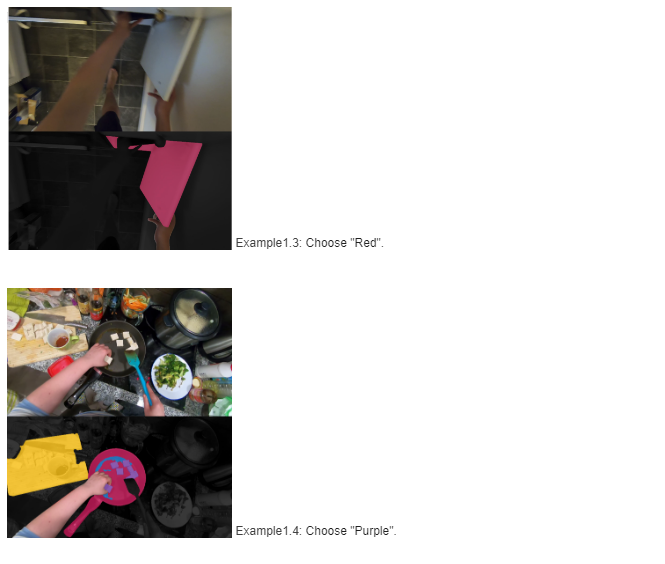} \\
Page 1  -- General Case & Page 2 -- General Case \\
\includegraphics[width=0.5\linewidth]{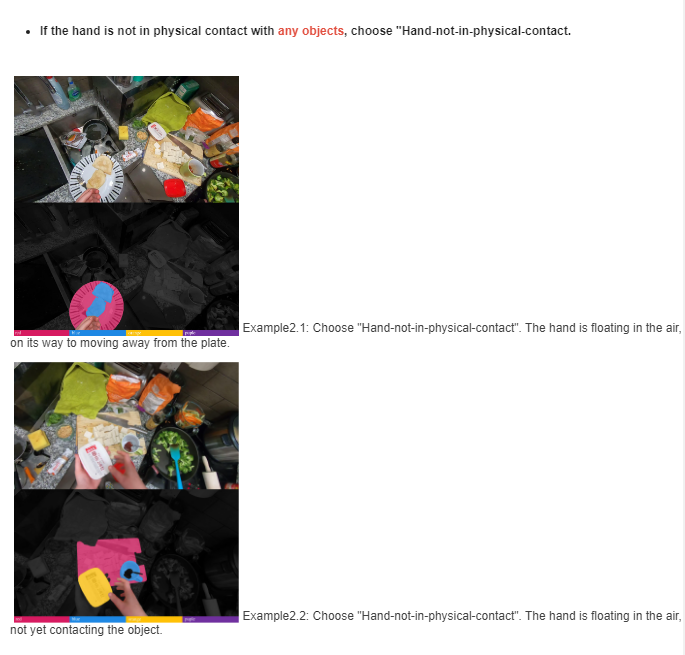} & 
\includegraphics[width=0.5\linewidth]{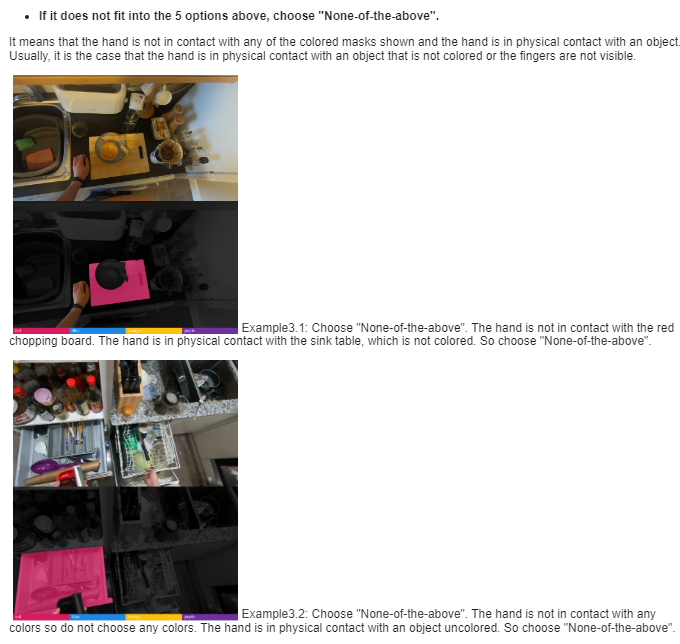} \\
Page 3 -- Not in Contact & Page 4 -- None of the Above \\
\includegraphics[width=0.5\linewidth]{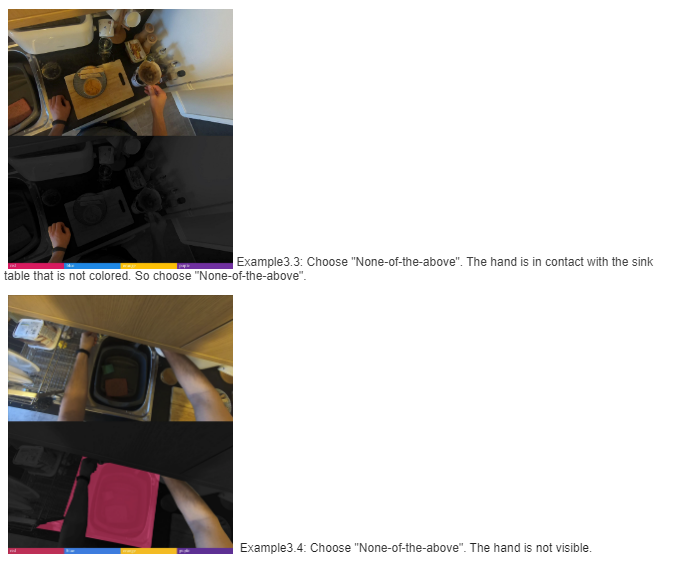} & 
\includegraphics[width=0.5\linewidth]{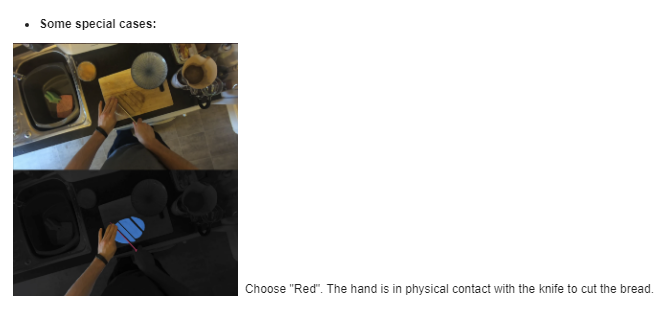} \\
Page 5 -- None of the above & Page 6 -- One Edge Case \\
\end{tabular}
\caption{{\bf Annotation Instructions for the HOS Task in Hive}. These cover several cases: a general case, not in contact, non-of-the-above, and a common edge case.}
\label{fig:hive_hos}
\end{figure}

We show the annotation instructions for the Hand-Object Relations task in Fig.~\ref{fig:hive_hos}. To give a consistent interface regardless of which object the hand is contacting, we colour potentially in-contact objects. We define potentially in-contact objects as any segmented object that shares a border with the hand. A small number (295) of frames have more than four potentially-in-contact objects. We annotate these manually ourselves.

\subsection{Annotation Instructions for Exhaustive Annotation}
\label{sec:app:exhaustiveannot}

\begin{figure}[t]
    \centering
    \includegraphics[width=1.0\textwidth]{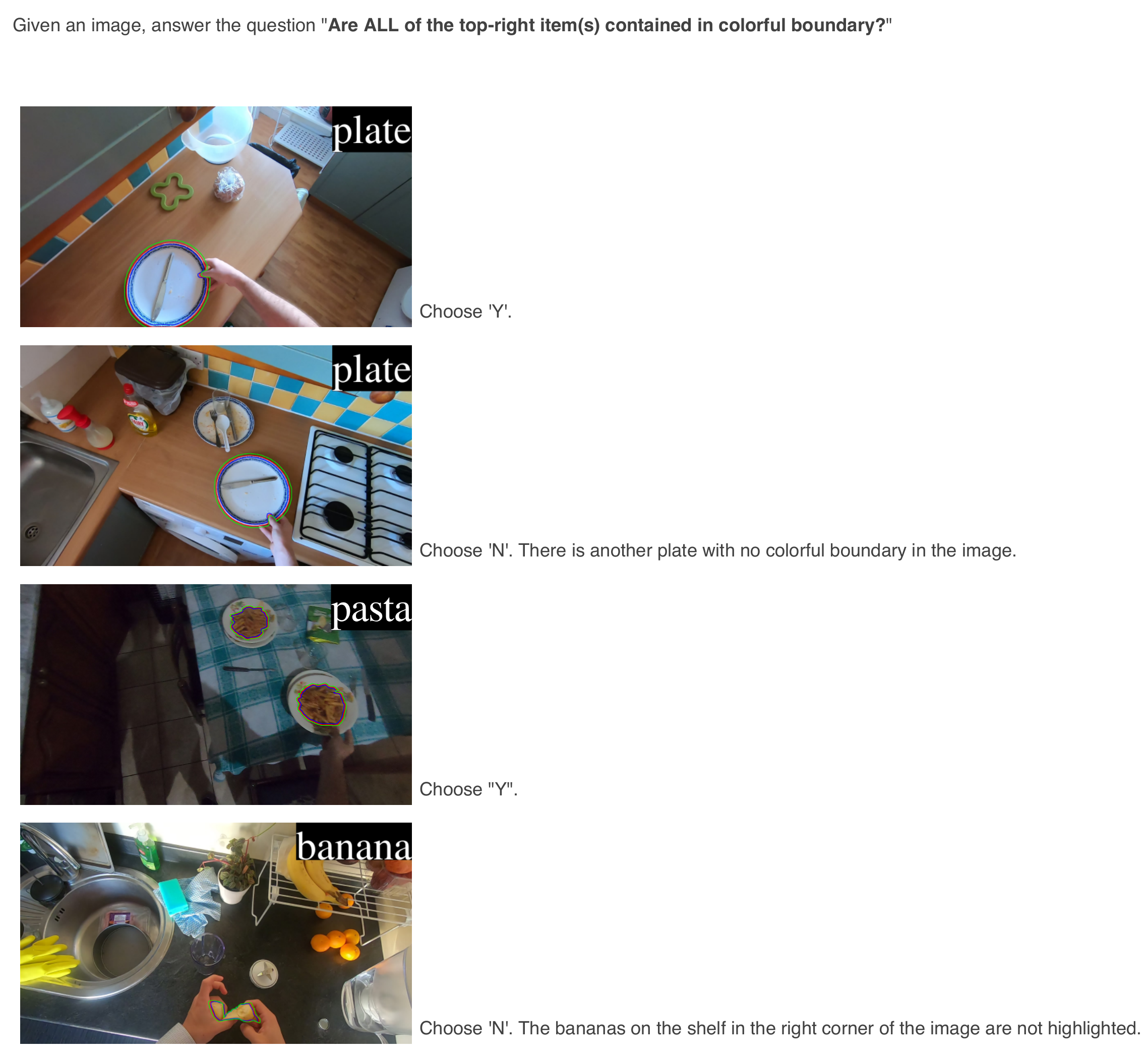}
    \caption{{\bf Annotation Instructions for exhaustive annotations in Hive}. We outline the object in question with colorful boundaries.}
    \label{fig:hive_exhaustive}
\end{figure}

We show the annotation instructions for the exhaustive annotations in Fig.~\ref{fig:hive_exhaustive}. Entity name is provided in the top right corner of each image, and triplet boundaries with three different colours are used to highlight each entity with a visual cue to avoid confusion.
As shown in the figure, the annotators should give a binary decision on whether all pixels related to the class listed on the top-right have been segmented.
Consistent annotations are used in the exhaustive flag.

\clearpage
\section{Appendix - Dense Annotations (Main \S2.4)}
\label{sec:app:dense}
{\bf Interpolation details.} We use STM model, pretrained on MS-COCO, and fine-tune it on VISOR annotations. We use 5 memory frames during inference. For evaluation, we use one memory frame to reconstruct the annotated frames for faster inference. Our code uses three Tesla V100 GPUs to generate the interpolation, two of them are used for the forward and backward STM passes, and one for combining the logits and evaluating the interpolations.

\begin{table}[t]
    \centering
    \caption{{\bf Detailed statistics of the automatic interpolations.} The numbers include the 2 annotated frames of each interpolation}
    \label{tab:interpolation_stats}
    \begin{tabular}{lccc}
    \toprule
    & Before filtering               &  After filtering                    \\ \midrule

\# interpolations & 35.2K&32.2K \\
\# interpolated objects  &172.0K & 119.4K& \\
\# images & 3.2M &2.8M \\
\# masks & 14.7M & 9.9M \\

    \bottomrule
    \end{tabular}
\end{table}

\begin{figure}[h!]
    \centering
    \includegraphics[width=1.0\textwidth]{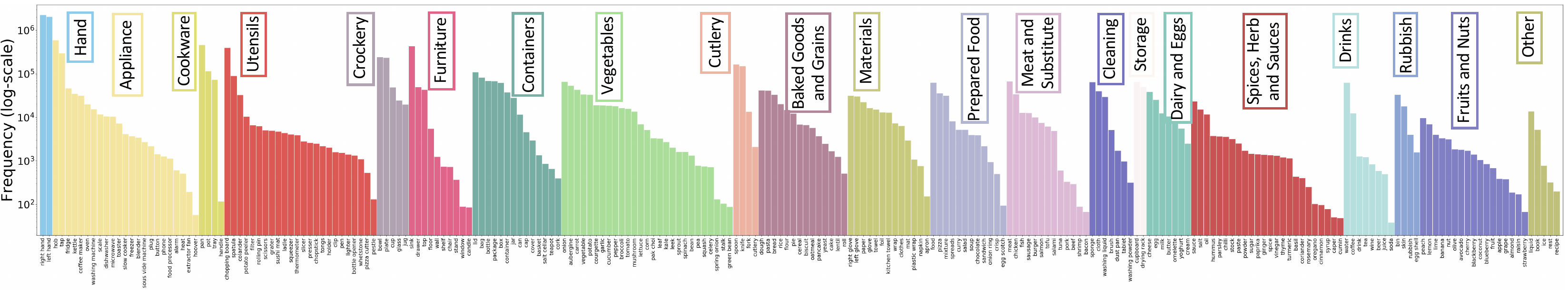}
    \caption{{\bf Frequency of interpolated entity classes} (Log y-axis) The histogram is long tailed. Best view with zoom}
    \label{fig:dense_classes}
\end{figure}

\begin{figure}[h!]
    \centering
    \includegraphics[width=0.65\textwidth]{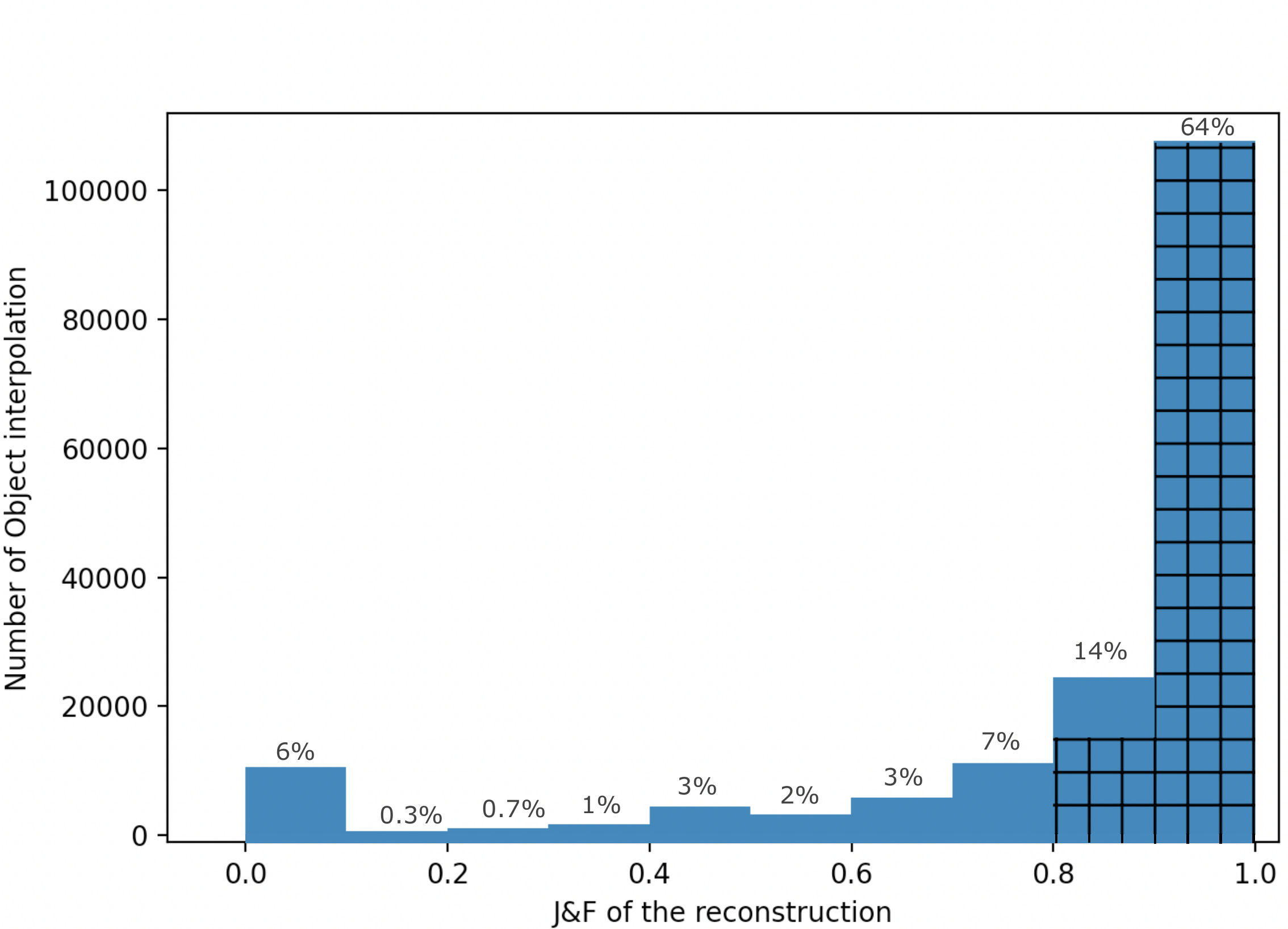}
    \caption{{\bf Histogram of interpolations' scores.} The hatched area is the selected area after filtering any $\mathcal{J}$\&$\mathcal{F}$ scores that are less 85\%.}

    \label{fig:interpolations_hist}
\end{figure}

{\bf Interpolation statistics.} Table~\ref{tab:interpolation_stats} shows the full statistics of the interpolation before and after filtering, we filter out interpolated objects with $\mathcal{J}$\&$\mathcal{F}$ < 85\% as shown in Fig.~\ref{fig:interpolations_hist}, so we keep 69.4\% of the object interpolations, having 9.9M masks. Fig. ~\ref{fig:dense_classes} shows the distribution of the of the interpolated segmentations per class, the distribution is similar to the sparse one in Fig.~\ref{fig:Entity_Distribution}. Hands masks form 40\% of all filtered interpolations. 

\begin{figure}[t]
    \centering
    \includegraphics[width=1.0\textwidth]{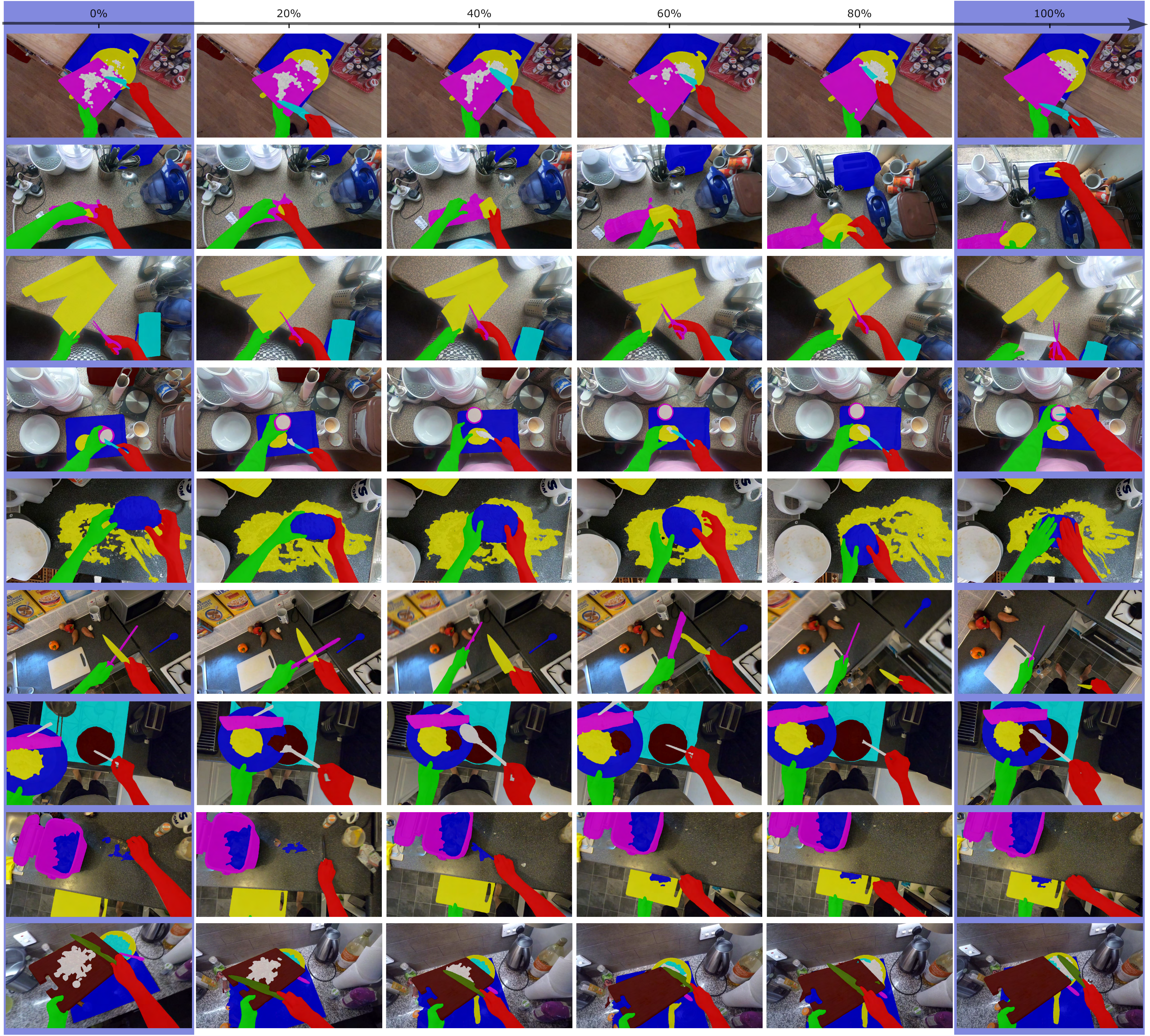}
    \caption{{\bf Sample segmentations from the dense automatic annotations.} First and last columns show the ground truth manually annotated frames and the rest are samples from the automatic interpolations at different ratio of the interpolation length.}
    \label{fig:dense1}
\end{figure}

\begin{figure}[t]
    \centering
    \includegraphics[width=1.0\textwidth]{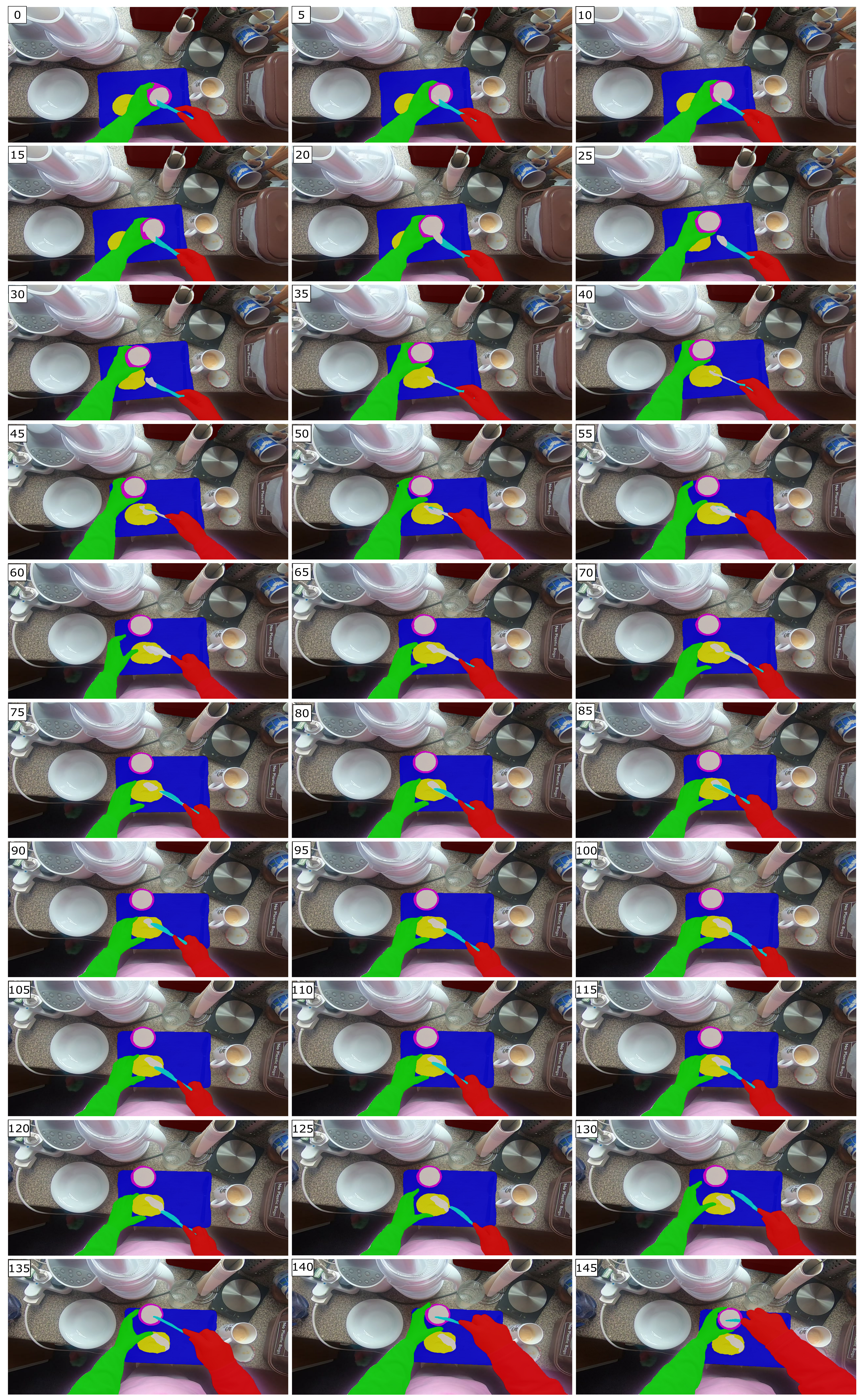}
    \caption{{\bf Example dense segmentations from the dense automatic annotations sampled every 5 frames.} The frame number included in the top right corner of each frame}
    \label{fig:dense2}
\end{figure}

\begin{figure}[t]
    \centering
    \includegraphics[width=1.0\textwidth]{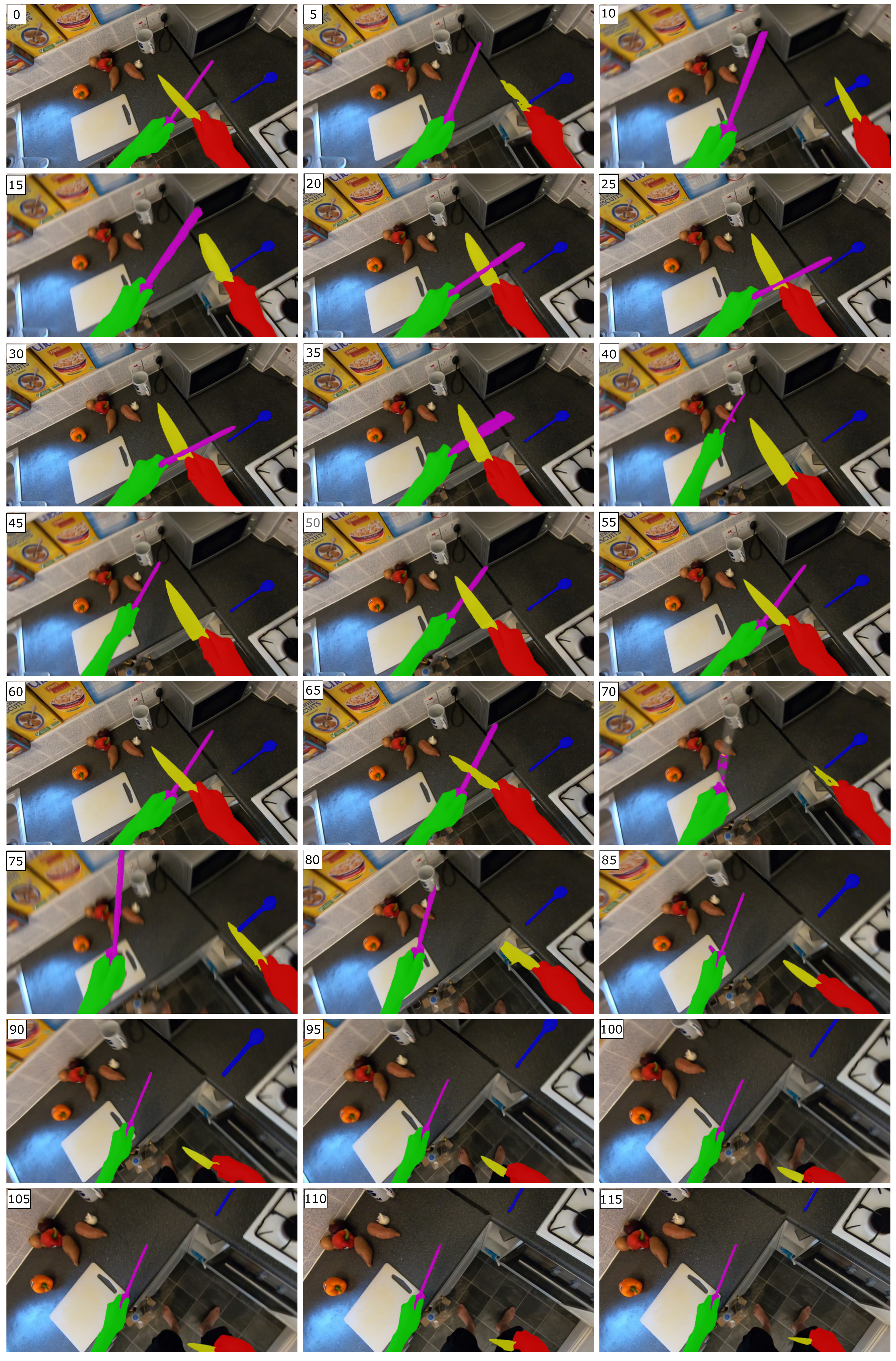}
    \caption{{\bf Second example dense  from the dense automatic annotations sampled every 5 frames.} The frame number included in the top right corner of each frame}
    \label{fig:dense3}
\end{figure}

{\bf Visualisations.} We show sample automatic dense annotations at 20\%, 40\%, 60\% and 80\% of the interpolated sequence length in Fig.~\ref{fig:dense1} which includes objects with different sizes and segmentation challenges. The figure shows how accurate the interpolations could be as we use information from the two annotated frames.
Fig.~\ref{fig:dense2} and Fig.~\ref{fig:dense3} show more detailed visualizations of the interpolations to showcase the temporal information they provide. 
We plot a frame every 5 frames for two long interpolation sequences.
In Fig.~\ref{fig:dense2}, we show butter being spread on toast where the knife, jar, butter, chopping board, toast and both hands are accuratetly segmented.
In Fig.~\ref{fig:dense3}, we show very fast motion as a knife is sharpened.
Some masks are inaccurate like frame because of the severe motion blur.
\\
Additional examples in the form of videos can be found on \href{https://epic-kitchens.github.io/VISOR/}{the project webpage}.


\clearpage
\section{Appendix - VOS Benchmark Details (Main \S4.1)}
\label{sec:app:VOS}

This section describes the semi-supervised VOS Benchmark, including data preparation (\S\ref{sec:app:vos:data}), metrics and evaluation (\S\ref{sec:app:vos:metrics}), baseline training details (\S\ref{sec:app:vos:baselines}), and additional results (\S\ref{sec:app:vos:visualizations}).

\subsection{Data Preparation}
\label{sec:app:vos:data}

\begin{table}[t]
\centering 
\caption{{\bf Statistics of VISOR for semi-supervised VOS.}  }
\label{tab:visor_semi_supervised_vos_stats}
\resizebox{\textwidth}{!}{
\begin{tabular}{llrrlll}\toprule
           & Kitchens/unseen & Seq & Masks &
           Obj/Seq $\mu/\sigma$, min, max & Img/Seq $\mu/\sigma$, min, max & Cls/unseen \\
           \midrule
Train      & 33                & 5,309           & 174,108 & 5.2 / 2.1, 1, 13                                                               & 6.2 / 3.4, 3, 171                                                               & 242            \\
Val & 24 / 5              & 1,244           & 34,160  & 5.3 / 2.3, 1, 13                                                               & 6.0 / 3.5, 2, 93                                                             & 165 / 9          \\
Test       & 13 / 4              & 1,283           & 46,998  & 5.4 / 1.8, 1, 10                                                               & 7.7 / 9.3, 2, 186                                                              & 151 / 6 \\ \bottomrule
\end{tabular}}

\end{table}

We adapt splits to be suitable for this benckmark. For each subsequence in Val/Test: we keep objects that appear in the first frame and ignore others from evaluation. This is because the benchmark only tracks segmentations that are present in the first frame.
 For the train set, we have not ignored any masks, we have just ignored any subsequence with less than 3 frames since at least 3 frames are required to train the network. Table ~\ref{tab:visor_semi_supervised_vos_stats} shows the statistics of the adopted version for semi-supervised VOS.
 We highlight unseen kitchens in Val/Train as well as unseen (or zero-shot) classes.

\subsection{Metrics and Evaluation}
\label{sec:app:vos:metrics}

We report our results in this benchmark using two measures as proposed in~\cite{pont20172017}:  \textbf{(1) Jaccard Index ($\mathcal{J}$)} Given the ground-truth $G$ and predicted mask $P$, the Jaccard Index is defined as ${\mathcal{J} = (|P \cap G|) / |P \cup G|)}$. This metric is not particularly sensitive to boundary accuracy, but is a common metric for evaluating segmentation. \textbf{(2) Boundary F-Measure ($\mathcal{F}$)} Given the ground-truth and predicted contours, the boundary F-measure $\mathcal{F}$ is defined as the F-score of the precision ($P$) and recall ($R$), or $(2PR)/(PR)$. This metric is more sensitive to boundary accuracy. 

The final score is the mean of the two metrics for all subsequences, and the score for each subsequence is the mean of its constituent objects. For full details refer to \emph{\S3} in~\cite{pont20172017}.

\begin{figure*}[h]
    \centering
    \includegraphics[width=0.65\textwidth]{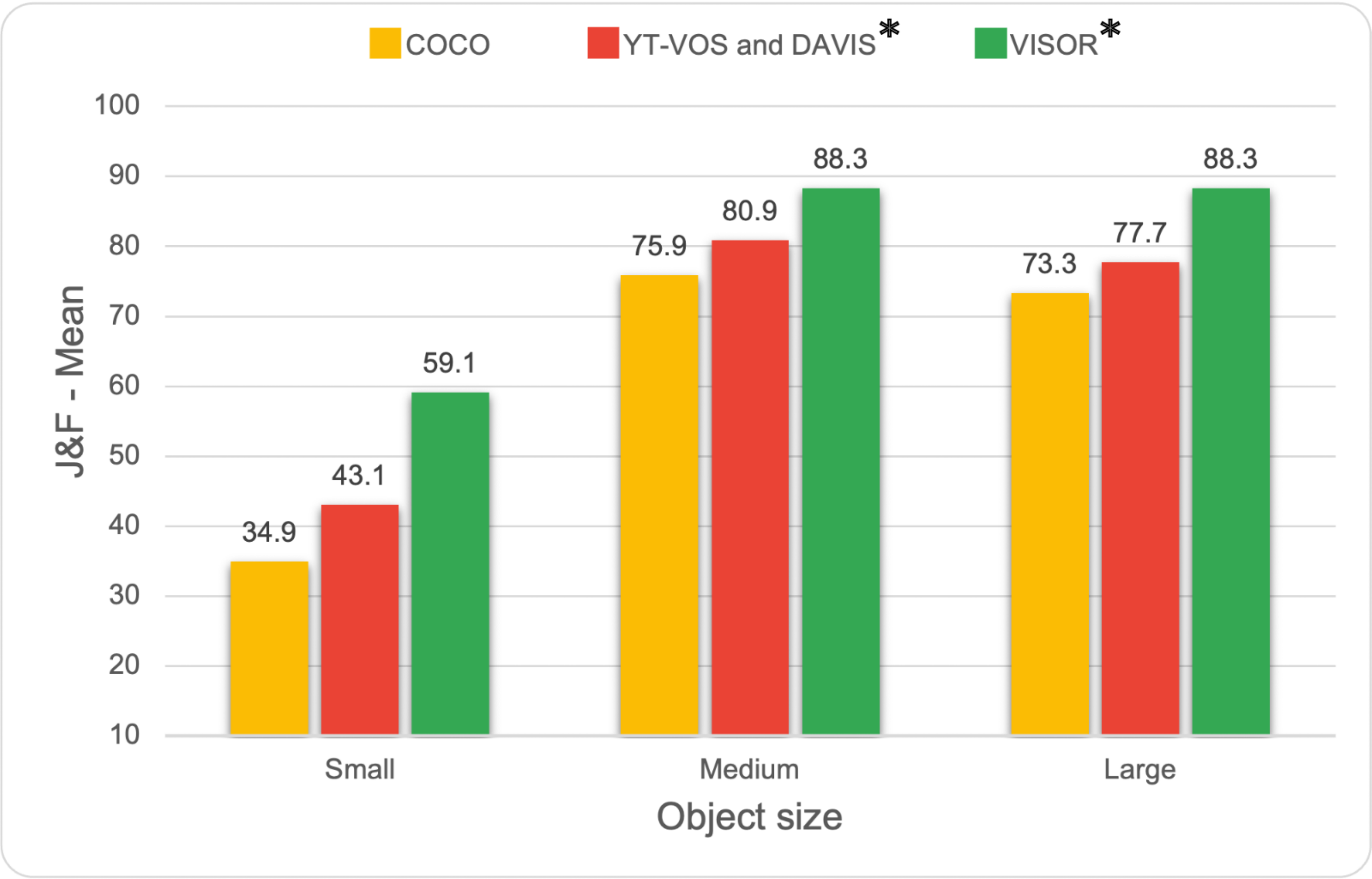}
    \caption{{\bf Size-based performance of STM trained on different datasets}. * means the model is pretrained on COCO. $\mathcal{J}$\&$\mathcal{F}$ is calculated on VISOR Val set and averaged on all object masks of each object size category (small, medium and large).Small objects are harder to be segmented}
    \label{fig:vos_size_score}
\end{figure*}

\begin{figure*}[t]
    \centering
    \includegraphics[width=1.0\textwidth]{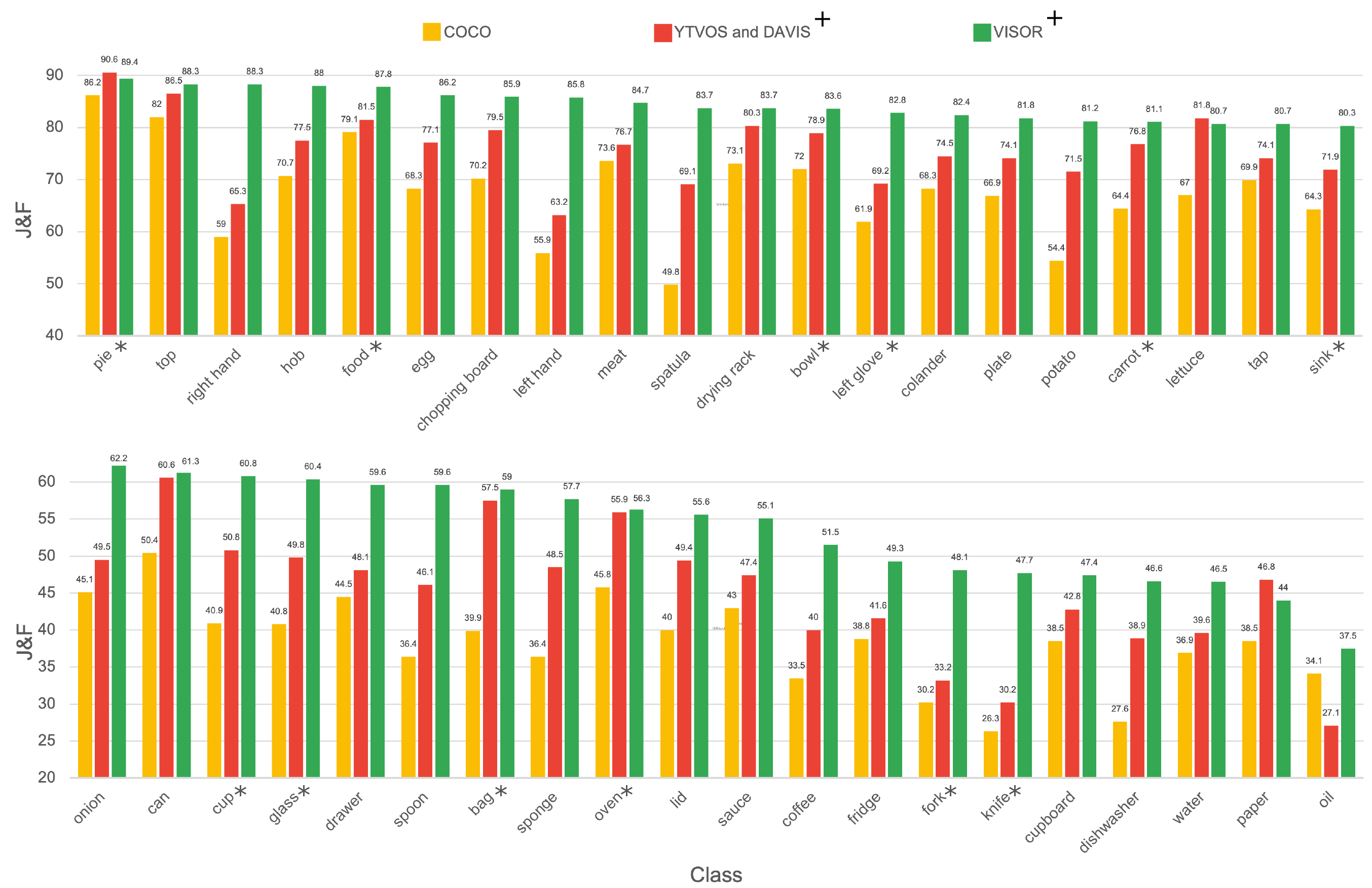}
    \caption{{\bf Class-based performance of STM trained on different datasets}. + means the model is pretrined on COCO, * means that the class is part of COCO categories.}
    \label{fig:scores_by_ocategory}
\end{figure*}

\subsection{Baseline Training Details}
\label{sec:app:vos:baselines}

\parnobf{Training Details} To train the baseline, we sampled 3 random images resized to 480p (854x480) from a subsequence with dynamic skipping between them. As we have sparse annotations with a dynamic number of intermediate frames, we initially sample without skipping and then use a maximum dynamic skipping of 1 half-way through the training process in a curriculum learning fashion. 

We use Resenet-50 as backbone. We train using a fixed learning rate of $10^{-5}$, batch of 32 and 400,000 iterations to fine-tune the COCO pretrained model. We also use cross-entropy loss and the Adam optimizer.  We use single Tesla V100 GPU to train and it took 4 days to train on our dataset.

During inference, since we have multiple objects per sequence as mentioned in Table ~\ref{tab:visor_semi_supervised_vos_stats} we segment each of them then we use their logits to classify the class for each pixel. Also, since the average number of the sparse frames per sequence is low (6.0 and 7.7) frames for validation and test set respectively as mentioned in Table ~\ref{tab:visor_semi_supervised_vos_stats}, there is no need to keep frames every N frames. We report results using 2 memory frames sampled uniformly throughout the sequence. We have just evaluated the model using the sparse annotations, however, denser frames could be extracted from EPIC-KITCHENS-100 videos (i.e.~50fps) to use during inference, this may help to get better results, but it is time and memory costly. 

\subsection{Additional Results}
\label{sec:app:vos:visualizations}

In addition of the results reported in the main paper, we provide more detailed results, breaking down results by size and class.

\parnobf{Size-based performance} In Fig.~\ref{fig:vos_size_score} we calculated the object size of the ground-truth masks of val, then we splitted them equally into small, medium and large objects. The figure shows that all models suffer with small objects, but the model fine-tuned on VISOR is 16\% better than the one pretrained on COCO and fine-tuned on both YTVOS and DAVIS. This gap is reduced to 7.4\% and 10.6\% for medium and large respectively. 

\parnobf{Class-based performance} Fig.~\ref{fig:scores_by_ocategory} shows the best and worst performing EPIC-KITCHENS-100 classes for 3 different models. The gap between the model fine-tuned on VISOR and others is not too large for the classes that are part of COCO dataset such as pie, food whereas the left/right hands and spatula have largest margins. In the worst classes, knife and fork have poor scores since they usually change their appearance during the subsequnece (based on the view point and object orientation) and as tools they usually are occluded most of the time as part of their functionality.

\clearpage
\section{Appendix - HOS Benchmark Details (Main \S4.2)}
\label{sec:app:hos}

We now describe the HOS Benchmark, including data preparation (\S\ref{sec:app:hos:data}), metrics and evaluation (\S\ref{sec:app:hos:metrics}), baseline training details (\S\ref{sec:app:hos:baselines}), and additional results (\S\ref{sec:app:hos:visualizations}).

\subsection{Data Preparation}
\label{sec:app:hos:data}

\parnobf{Hand Object Relation Annotations from Hive} The annotated labels for hand object relations from Hive are shown in Table ~\ref{tab:hos_hive}. ``Failed upload'' means samples that failed uploading to Hive, which we manually annotated.  ``More than 4 candidate'' cases are samples that have more than 4 candidate masks potentially in contact with the hand. Since we only showed the annotators 4 colored options, we manually checked and annotated these ourselves.

\begin{table}[t!]
    \centering
    \caption{{\bf Hand Object Relation Annotation Stats.} Most of the annotations are marked as being in contact with one object.}
    \label{tab:hos_hive}
    \begin{tabular}{ccc}
    \toprule

                        & Train+Val               &  Test                    \\ \midrule
In Contact              & 52,685 (81.2\%)          &  14,233 (82.4\%) \\
Not-in-contact          & 4,144 (6.4\%)            &  1,341 (7.8\%)               \\
None-of-the-above       & 3,079 (4.7\%)            & 592 (3.4\%)                \\
Inconclusive            & 4,943 (7.6\%)            & 1,104 (6.4\%)               \\
Failed upload                     & 1              & 1                     \\
$>$ 4 candidate objects           & 272                      & 23                       \\

\bottomrule
    \end{tabular}
\end{table}

\parnobf{Hand in Gloves} As noted before, hands are frequently in gloves during some kitchen activities like cleaning or using the oven. When the glove is worn on a hand, we consider the glove as part of the hand, which means the current hand mask is now any visible hand parts plus the glove mask. The object that the glove is in contact with is thus considered as an object in contact with the hand.
Gloves that are not worn on hands are considered as normal objects/masks. 

\begin{figure}[t]
    \centering
    \includegraphics[width=1.0\textwidth]{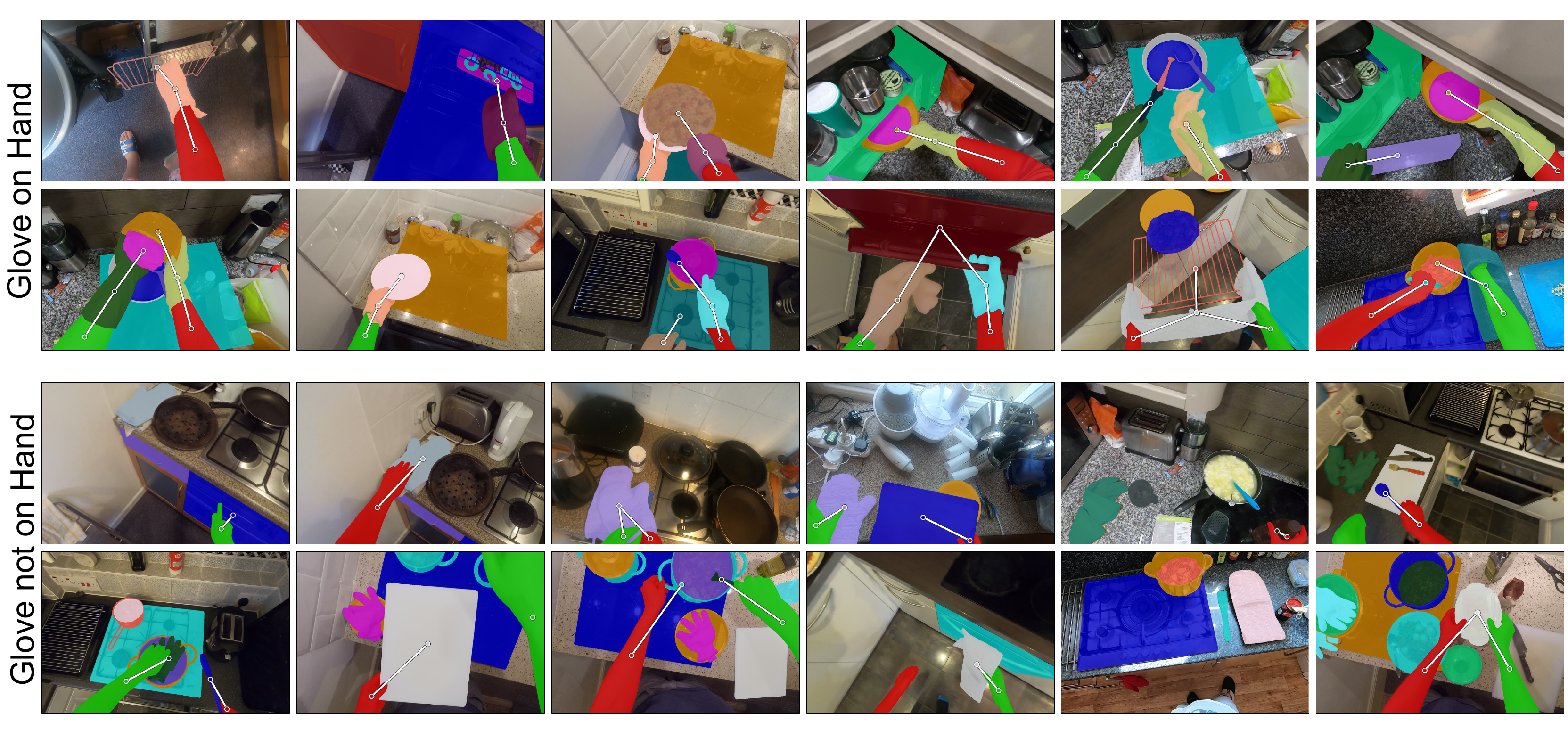}
    \caption{{\bf Glove object relation annotation}. Glove-on-hand examples are shown on the top two rows and Glove-not-on-hand examples are shown on the bottom two rows.}
    \label{fig:glove_obj}
\end{figure}

In the data set, there are 941 (674/247/20 in train/val/test set) images with 1,409 (957/432/20) glove entities, out of which 1,105 (698/396/11) gloves are on hands. For hands in gloves, we follow the same way to annotate the in-contact objects as we did with hands. Table~\ref{tab:glove_on_which_hand} shows distribution of glove annotations and Table~\ref{tab:glove_obj} shows the distribution of the annotated glove-object relations. Some examples of glove object relations are shown in Fig~\ref{fig:glove_obj}.

\begin{table}[t!]
\parbox{.45\linewidth}{

    \centering
    \caption{{\bf Gloves-on-which-hands annotation distribution.}}
    \label{tab:glove_on_which_hand}
    \begin{tabular}{ccc}
    \toprule
                      & Glove     \\ \midrule
on left hand          & 497       \\
on right hand         & 561       \\
on both hands         & 13        \\
not on hand           & 304        \\
inconclusive          & 34        \\ \midrule
total                 & 1409      \\
\bottomrule
    \end{tabular}
    
}
\hfill
\parbox{.45\linewidth}{

\centering
    \caption{{\bf Glove object relation annotation distribution.}}
    \label{tab:glove_obj}
    \begin{tabular}{ccc}
    \toprule
                      & Glove-on-Hand     \\ \midrule
glove-in-contact         & 930       \\
glove-not-in-contact  & 31        \\
none-of-the-above     & 9       \\
inconclusive          & 101       \\ \midrule
total                 & 1071      \\
                
\bottomrule
    \end{tabular}
}
\end{table}

\parnobf{Training and Evaluation Data} We prepare our data annotations in COCO format for each task separately following the same train/val/test split of VISOR.

For {\it Hand-Contact-Relations} task, we prepare two-class annotations, one class for ``hand'' (both left and right hand), the other one for ``object'' which are the contacted objects annotated in the VISOR Object Relation annotations. Additionally we add:  hand side, in-contact and offset, in the original COCO annotation. Hand side is binary where 0 indicates left hand and 1 indicates right hand. Contact is also binary, where 0 indicates the hand is not in contact and 1 indicates the hand is in contact. The offset from the hand bounding box centre to the in-contact object bounding box centre is factored into a unit vector $v \in \mathbb{R}^2$ and a magnitude $m \in \mathbb{R}$, as in~\ref{refA} 

For {\it Hand-And-Active-Objects} task, we prepare two-class annotations, one class for ``hand'' (both left and right hand), the other one for ``object'' which are all other objects. 

\subsection{Metrics and Evaluation}
\label{sec:app:hos:metrics}
We evaluate via instance segmentation tasks using the COCO Mask AP~\cite{lin2014microsoft}. We evaluate per-class to better show the performance on each class. We only keep images  with conclusive annotations in Val/Tes. In other words, if there is a hand in the image that has inconclusive or none-of-the-above annotation for contact state we do not use it in evaluation. 

For {\it Hand-Contact-Relations} task, we prepare three schemes for hand classes: all hands as one class; hands split by side (left/right); and hands split by contact state. These 3 individual evaluations better show the performance on purely hand mask prediction, hand mask + hand side prediction and hand mask + contact prediction. 
The offset between hand and object is evaluated implicitly by associating each hand that is predicted as in-contact with an object if exists. Here the object evaluation is on the object prediction after this post-processing.

For {\it Hand-And-Active-Objects} task, we do normal per-class instance segmentation evaluation.

\subsection{Baselines and Training Details}
\label{sec:app:hos:baselines}

For both tasks, we use PointRend ~\cite{kirillov2020pointrend} instance segmentation model implemented in Detectron2~\ref{refB} with R50-FPN backbone and standard 1× learning rate scheduled configuration by default. It is trained for 90,000 iterations with batch size of 24 and base learning rate of 0.02.

Specifically for {\it Hand-Contact-Relations} task, we add 3 additional linear layers after the ROI-Pooled feature to predict hand side (feature size, 2), contact state (feature size, 2) and offset (feature size, 3). During training, we use Cross-Entropy loss for hand side and contact state and MSE loss for offset. We skip and do not supervise hand contact and offset on inconclusive and none-of-the-above hand contact annotations.

\subsection{Additional Results}
\label{sec:app:hos:visualizations}
\begin{figure}[t]
    \centering
    \includegraphics[width=1.0\textwidth]{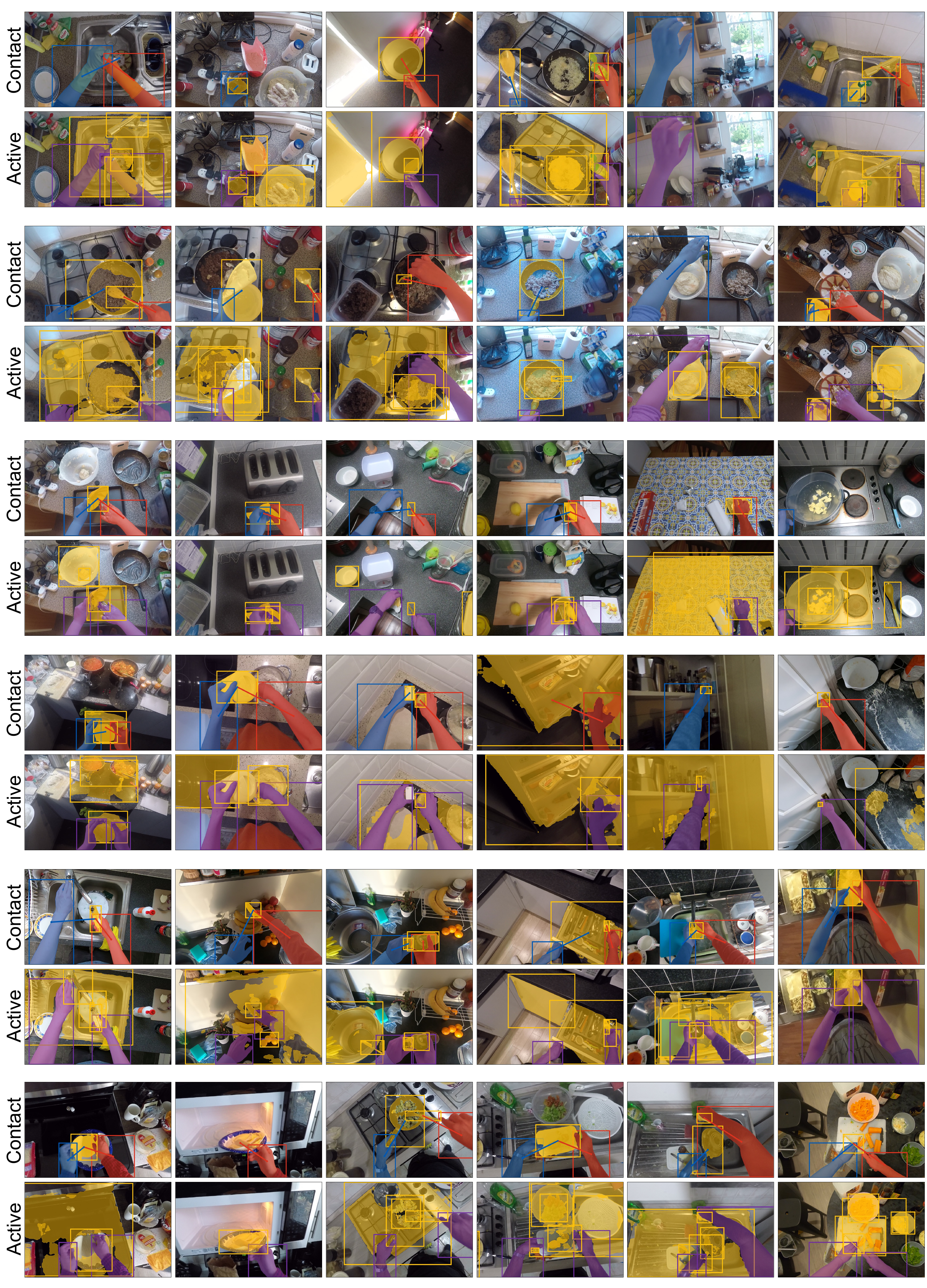}
    \caption{{\bf Qualitative results of HOS on Val set}. Hand segments are often very accurate and many objects are segmented correctly, although this is still a challenging task.}
    \label{fig:hos-quali}
\end{figure}

We showcase additional HOS qualitative results on validation set in Fig~\ref{fig:hos-quali}.

\clearpage
\section{Appendix - WDTCF Benchmark Details (Main \S4.3)}
\label{sec:app:wdtcf}

In this section, we introduce the data annotation (\S\ref{sec:app:wdtcf:data}), evaluation (\S\ref{sec:app:wdtcf:metrics}) and baseline details (\S\ref{sec:app:wdtcf:baselines}) and additional results (\S\ref{sec:app:wdtcf:visualizations}) for the WDTCF benchmark. 

\subsection{Data Preparation and Annotation} 
\label{sec:app:wdtcf:data}

First, we select query object candidates that are meaningful to ask the question ``Where did this come from?", for instance, static objects (e.g., `fridge', `oven' and `sink'), hands, mixture and food are excluded. 
Then, we extract the query and evidence frame candidates based on rigorous rules. Concretely, given an untrimmed video, the frames for each query object are linked throughout, (e.g., `bowl' and `milk'). 
We consider the last three frames that feature the query object as potential query frame candidates, and the first three frames with co-occurrence with the object and any of our 15 containers as evidence frame candidates. 
The number of query and evidence frames are empirically set to be three.

\begin{figure}[h]
    \centering
    \includegraphics[width=1.0\textwidth]{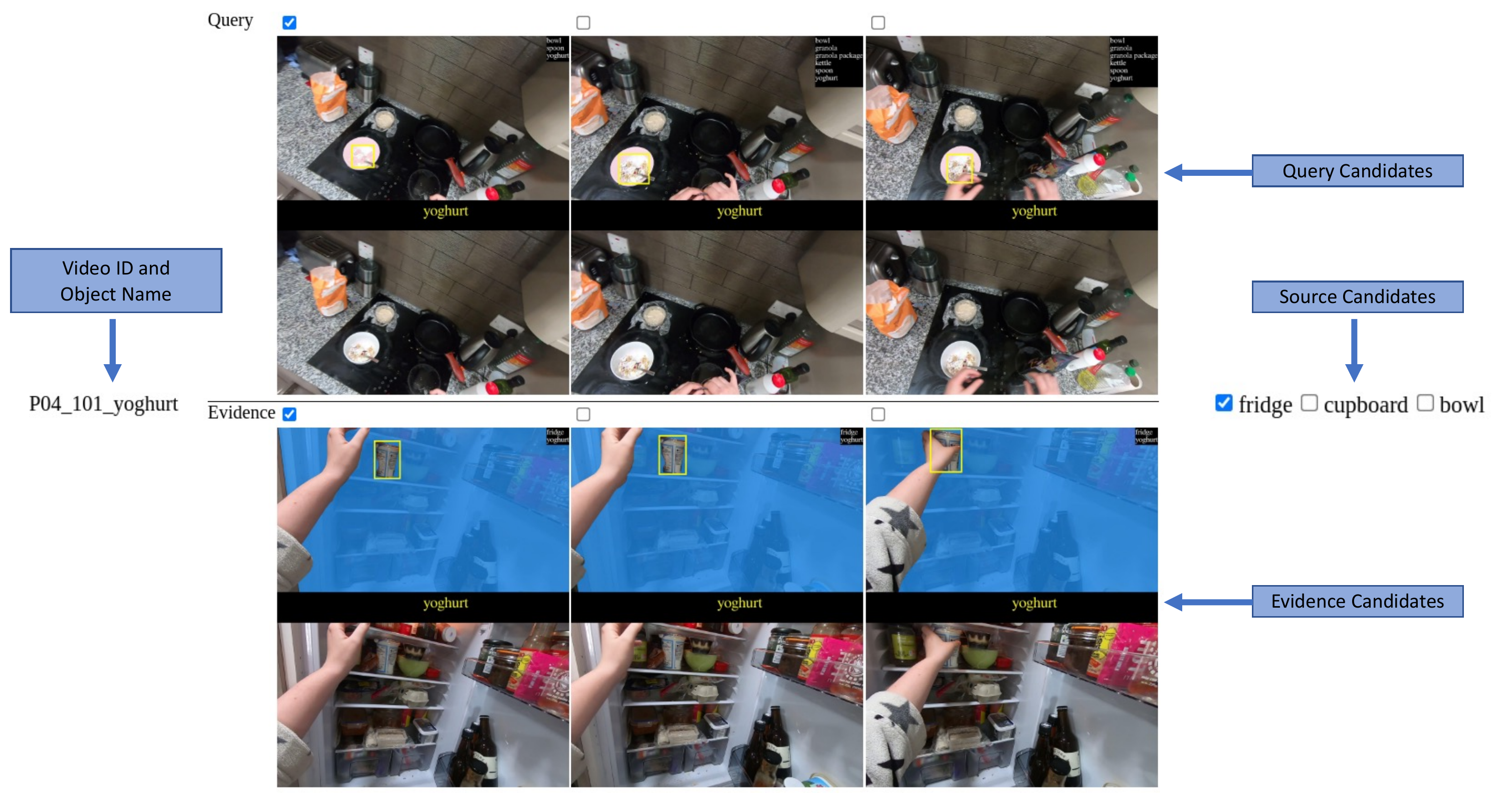}
    \caption{{\bf Annotation interface for where did this come from}. This interface is used to identify the source object in the video.}
    \label{fig:where_interface_demo}
\end{figure}

Fig.~\ref{fig:where_interface_demo} shows the interface for annotation. The key components contain the video ID, the query object name, query candidates, evidence candidates and source candidates. In this example, the query object `yogurt' can be clearly seen emerging from `fridge'. Note that although it could contain multiple evidence frames for each query object, only one is annotated for the taster challenge.

\begin{figure}[h]
    \centering
    \includegraphics[width=0.6\textwidth]{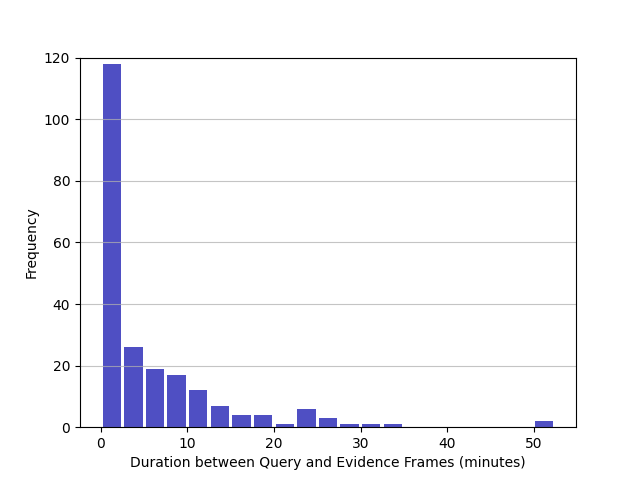}
    \caption{\textbf{Distribution of duration between query and evidence frames.} Some are close, but many are far away.}
    \label{fig:dis_Q_E}
\end{figure}

We annotate 222 examples for this test set.
Fig.~\ref{fig:dis_Q_E} shows the distribution of duration between query and evidence frames. 
While many gaps are small (within 2 minutes), the duration varies greatly, with 19.4\% longer than 10 minutes.

\subsection{Evaluation Details}
\label{sec:app:wdtcf:metrics}

Given the query segment, each method produces: (1) a class id indicating the source object; (2) an evidence frame where the query object emerges from the source object; as well as (3) segments of the source and query objects in the evidence frame. 

This challenge is defined on sparse frames. While methods are free to look at the dense frames, they are asked only to produce results and evaluated {\it only} on the sparsely annotated frames. This is important to make sure that methods are not being asked to identify the {\it precise} moment that an object emerges out of a high frame rate video. Instead, WDTCF asks the method to identify which of a number of distinct keyframe best shows the object emerging. 

Each method is evaluated on three metrics. First, {\it Source} evaluates the accuracy of the class prediction (i.e., whether the predicted class is the same as the ground-truth class). Second, {\it Query IoU} evaluates the intersection-over-union/Jaccard Index of the query object. This is zero if the evidence frame is not localized correctly. Finally {\it Source IoU} evaluates the IoU of the source object. This is also zero if the evidence frame is not localized correctly.

\subsection{Baseline Details} 
\label{sec:app:wdtcf:baselines}

We next explain how we use PointRend model to produce baseline results with oracle knowledge which is also trained with R50-FPN backbone and default learning schedule in Detectron2~\ref{refB} for 90,000 iterations using batch size 24 and base learning rate 0.02. Note that the model is trained using the Train and Val sets of VISOR, and the WDTCF examples are obtained from the Train and Val sets as well.
We predict the query object class from the best overlap with masks based on the trained model. Note that WDTCF is defined only on the sparse frames with ground truth masks, which we use for IoU evaluation. Specifically, if the query name is predicted correctly,
then the model is used to further detect the co-occurrence of the object and one of the potential 15 sources starting from the first frame of the video as evidence frame candidates.
We consider first-3 candidate frames and pick the one with the highest confidence score as evidence frame prediction.

In baselines where the evidence frame is given, the problem is simplified to predicting the source in the evidence frame, regardless of the query object. Consequently, the prediction of query and source masks of the evidence frame are directly used to compute the IoU with ground truth masks.

\begin{figure}[t]
    \centering
    \includegraphics[width=1.0\textwidth]{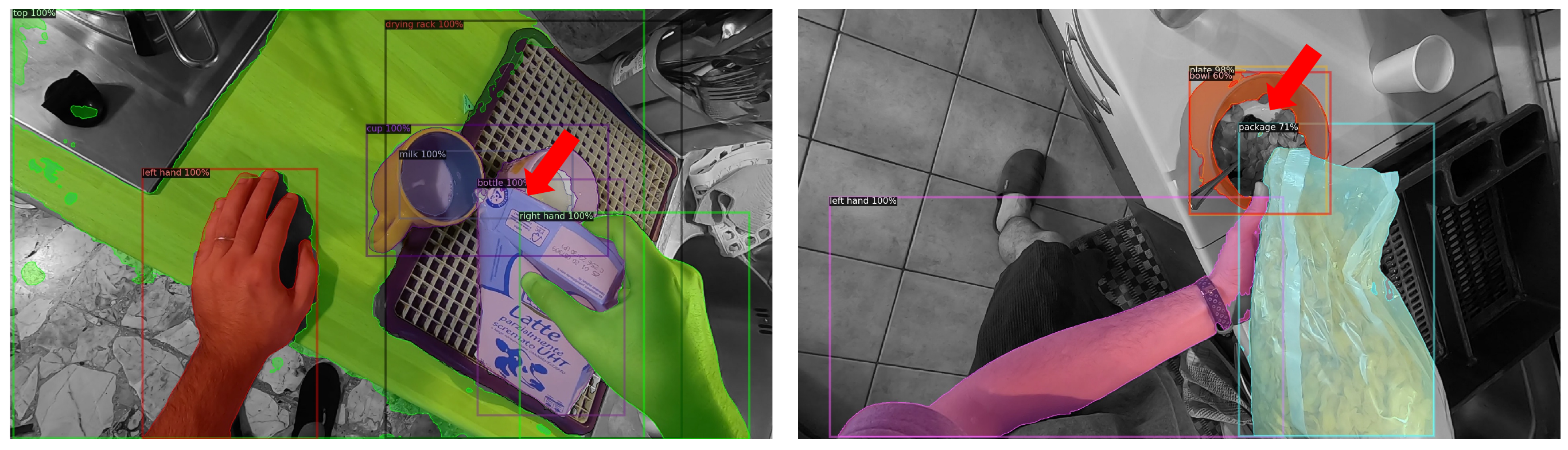}
    \caption{{\bf Two examples of PointRend prediction results on query frame}. We show  one \textcolor{greensuccess}{success} (left with query object `bottle') and one \textcolor{redfailure}{failure} (right with query object `yogurt').}
    \label{fig:pointrend_queryprediction}
\end{figure}

\parnobf{Query Object Prediction}
Fig.~\ref{fig:pointrend_queryprediction} shows two examples of PointRend prediction results on query frame. In the left example, the query object is detected and segmented perfectly. Therefore, the query object `bottle' can be predicted correctly by comparison with the GT masks of `bottle'. However, the `yogurt' in the bowl is undetected due to occlusion of cereal in the right example. For all the query masks, the overall prediction accuracy is 90.5\%, which shows that the query object segmentation is non-trivial. For instance, the tiny objects (e.g., garlic) and liquid (e.g., coffee and oil) are hardly to be detected.

\begin{figure}[t]
    \centering
    \includegraphics[width=1.0\textwidth]{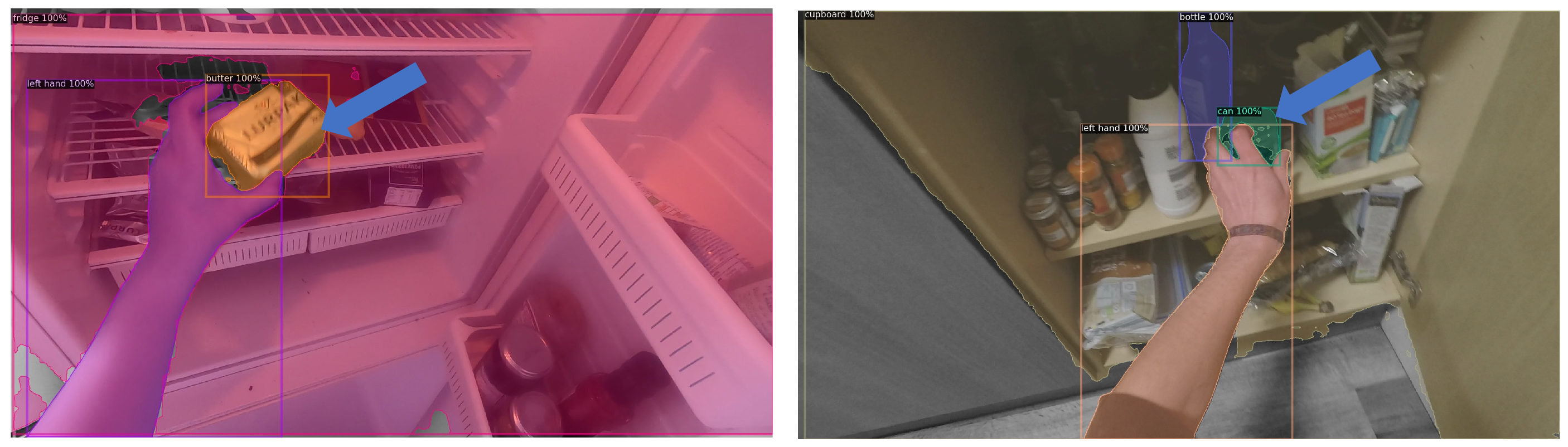}
    \caption{{\bf Two examples of PointRend prediction results on evidence frame.} The blue arrow indicates the query object.}
    \label{fig:pointrend_evidence}
\end{figure}

\parnobf{Evidence Frame Segmentation} Fig.~\ref{fig:pointrend_evidence} shows two examples of PointRend segmentation results in the evidence frame. The only predicted source in the left example is `fridge', thus it is trivial to predict the source with given evidence frame. In contrast, there are two source candidates are detected in the right example, i.e., `bottle' and `cupboard'. Without prior knowledge, it is difficult for the model to decide which one is a better choice.

\subsection{Additional Results}
\label{sec:app:wdtcf:visualizations}
\begin{figure}[t]
    \centering
    \includegraphics[width=.8\textwidth]{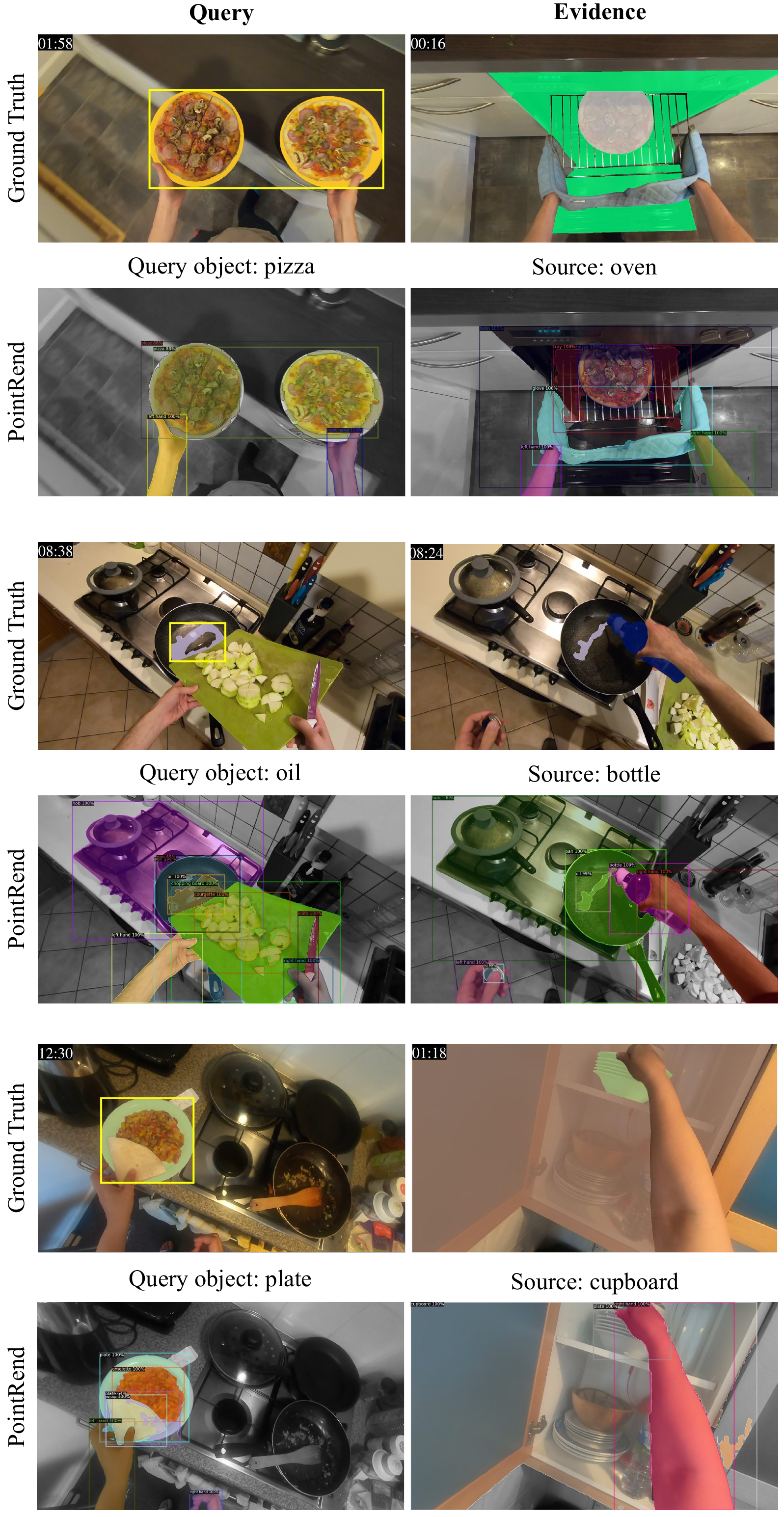}
    \caption{{\bf Visualization of ground truth along with PointRend prediction results.}}
    \label{fig:pointrend_additional}
\end{figure}

We showcase additional WDTCF qualitative results in Fig~\ref{fig:pointrend_additional}.

\clearpage
\section{Appendix - EPIC-KITCHENS VISOR - Datasheet for Dataset}
\label{sec:app:datasheet}

\newcommand{\dsq}[1]{{\vspace{1mm} \par \noindent {\bf Q. #1}}}
\newcommand{\dsqc}[2]{{\vspace{1mm} \par \noindent {\bf Q. #1} {\it #2}}}

\newcommand{\dsa}[1]{{{\bf A.} #1}}
\newcommand{\dsak}[1]{{\textcolor{blue}{{\bf (EK) A.} #1}}}

This paper introduces a new set of annotations, VISOR, for the EPIC-KITCHENS-100 dataset. Given that the VISOR annotations depend on the EPIC-KITCHENS-100 dataset, we have also answered some of the questions about ethics and consent for EPIC-KITCHENS-100. We have marked \textcolor{blue}{these answers in blue}. For instance: 

\dsq{Datasheet Question}

\dsa{This answer applies to VISOR}

\dsak{This answer applies to EPIC-KITCHENS. }

\subsection*{Motivation}

\dsqc{For what purpose was the dataset created?}{Was there a specific task in mind? Was there a
specific gap that needed to be filled? Please provide a description.
}

\dsa{The VISOR dataset introduces a large dataset of segmentations that are consistent segmented throughout long videos, as well as relations between these segments in both time and space. The goal of the dataset is to provide a setting in which the community can evaluate rich long-term video understanding.}

\dsq{Who created the dataset (e.g., which team, research group) and on behalf of which entity
(e.g., company, institution, organization)?}

\dsa{The VISOR annotations were the joint work of groups located at the Universities of Bristol, Michigan, and Toronto. The primary contributors are Ahmad Darkhalil (PhD candidate), Dandan Shan (PhD candidate), Bin Zhu (Postdoctoral researcher), Jian Ma (PhD candidate), Amlan Kar (PhD candidate), Richard Higgins (PhD candidate), Sanja Fidler (Faculty), David Fouhey (Faculty), and Dima Damen (Faculty).}

\dsqc{Who funded the creation of the dataset?}{If there is an associated grant, please provide the
name of the grantor and the grant name and number.}

\dsa{ The dataset was funded in multiple parts. The segmentation annotations themselves were funded by charitable unrestricted donation from Procter and Gamble as well as charitable unrestricted donation from DeepMind. 

Support for the dataset creation (i.e., student and PI support) was funded as follows: (a) Research at the University of Bristol is supported by UKRI Engineering and Physical Sciences Research Council (EPSRC) Doctoral Training Program (DTP), EPSRC Fellowship UMPIRE (EP/T004991/1) and EPSRC Program Grant Visual AI (EP/T028572/1); (b) Research at the University of Michigan is based upon work
supported by the National Science Foundation under Grant No. 2006619; (c) Research at the University of Toronto is in part sponsored by NSERC as well as support through the Canada CIFAR AI Chair program.  
}

\dsak{The EPIC-KITCHENS videos were funded by a charitable unrestricted donation from Nokia Technologies and the Jean Golding Institute at the University of Bristol.}

\dsq{Any other comments?}

\dsa{No}

\subsection*{Composition}

\dsqc{What do the instances that comprise the dataset represent (e.g., documents, photos, people,
countries)?}{Are there multiple types of instances (e.g., movies, users, and ratings; people and
interactions between them; nodes and edges)? Please provide a description.}

\dsa{The VISOR dataset contains segments (i.e., a collection of pixel masks) and relations between these segments. These relations consist of: (a) both open-vocabulary and mapped closed-vocabulary semantic entity names that link the segments of the same category; (b) relationships that link each hand in the dataset to at most one object that the hand is in contact with; and (c) long-term relationships between objects and the container from which they emerge in a previous frame.}

\dsak{VISOR is built upon the EPIC-KTICHENS-100 dataset. EPIC-KITCHENS-100 contains 700 variable-length videos along with extensive metadata and labelling, and VISOR has been annotated on frames from this dataset.}

\dsq{How many instances are there in total (of each type, if appropriate)?}

\dsa{VISOR contains: (a) ${\approx}$271.6K manually segmented semantic masks covering 257 object classes and ${\approx}$9.9M automatically obtained dense masks; (b) 67K hand-object relations; (c) exhaustive labels for each segment indicating whether workers believe it to be exhaustive; and (d) 222 instances of long-term tracking.}

\dsqc{Does the dataset contain all possible instances or is it a sample (not necessarily random) of
instances from a larger set?}{If the dataset is a sample, then what is the larger set? Is the sample
representative of the larger set (e.g., geographic coverage)? If so, please describe how this
representativeness was validated/verified. If it is not representative of the larger set, please
describe why not (e.g., to cover a more diverse range of instances, because instances were
withheld or unavailable). }

\dsa{VISOR is annotated on sparse frames at a rate of approximately 2 per EPIC-KITCHENS action. This subselection is needed because it is not possible to annotate every frame in the dataset. }

\dsqc{What data does each instance consist of?}{``Raw'' data (e.g., unprocessed text or images) or
features? In either case, please provide a description.}

\dsa{The VISOR dataset consists of multiple types of instance. The segments consist of: (a) a frame-level mask, (b) short-term track ID linking masks within a subsequence, (c) open-vocabulary semantic entity name, (d) closed-vocabulary grouping into one category, and (e) exhaustive flag on whether the mask covers all instances of that category in the image. For {\bf hands} you additionally have (f) hand side, (g) hand contact state and (h) ID of mask in-contact with the hand if present.}

\dsqc{Is there a label or target associated with each instance?}{ If so, please provide a description.}

\dsa{The VISOR dataset consists of labels. See the description of the instances above. The entity classes are also mapped into macro-classes for a one-level hierarchical grouping. This is inherited from EPIC-KITCHENS-100.} 

\dsqc{Is any information missing from individual instances?}{If so, please provide a description,
explaining why this information is missing (e.g., because it was unavailable). This does not
include intentionally removed information, but might include, e.g., redacted text.}

\dsa{Yes, we flag in-exhaustively annotated instances. We also only segment active objects in the frame and do not segment background objects. This not only highlights active objects but also avoids expensive annotations for objects lying around and of no relevance to the ongoing action.}

\dsqc{Are relationships between individual instances made explicit (e.g., users' movie ratings, social
network links)?}{If so, please describe how these relationships are made explicit.}

\dsa{The segments of the dataset are linked in a variety of ways: (a) segments within the same sub-sequence share a unique ID, (b) segments that have the same name are linked;  
(c) hands that are in contact with objects are linked; (d) a number of segments are linked to the segments from which they emerge earlier in the video.
}

\dsqc{Are there recommended data splits (e.g., training, development/validation, testing)?}{If so,
please provide a description of these splits, explaining the rationale behind them}

\dsa{Yes, we provide splits that are detailed in Sec of the paper. We first use the Test split from EPIC-KITCHENS-100 as the source of our test set. The action annotations in this split are hidden and only available by submitting to the evaluation server. This ensures our test annotations for VISOR are also suitable to use for a formal challenge. We provide a new train/val split from the train and val videos of EPIC-KITCHENS-100. This focuses on ensuring: (a) a number of unseen kitchens are available in Val to assess generality in the same way as Test; (b) some zero-shot classes exist; (c) an 80-20 split of masks is roughly selected per seen kitchen. We use the same Train/Val/Test splits for all the VISOR challenges.
}

\dsqc{Are there any errors, sources of noise, or redundancies in the dataset?}{If so, please provide a
description.}

\dsa{Noise and errors are inevitable in datasets. The most likely source of errors is incorrect labels or fundamental ambiguities in labeling. However, the VISOR dataset collection process has multiple quality assurance steps that aims to substantially reduce the prevalence of noise and errors.}

\dsqc{Is the dataset self-contained, or does it link to or otherwise rely on external resources (e.g.,
websites, tweets, other datasets)?}{If it links to or relies on external resources, a) are there
guarantees that they will exist, and remain constant, over time; b) are there official archival
versions of the complete dataset (i.e., including the external resources as they existed at the
time the dataset was created); c) are there any restrictions (e.g., licenses, fees) associated with
any of the external resources that might apply to a future user? Please provide descriptions of
all external resources and any restrictions associated with them, as well as links or other access
points, as appropriate.}

\dsa{The dataset relies on EPIC-KITCHENS-100. (a) EPIC-KITCHENS-100 is available via the data.bris service, which provides for long-term preservation of the dataset even in the case that PIs move institution. (b) We used \href{https://data.bris.ac.uk/data/dataset/2g1n6qdydwa9u22shpxqzp0t8m}{this} version of the data, but to prevent issues with consistent extraction of frames from videos, we will provide our copies of the frames that were used. (c) While there are licensing restrictions (non-commercial use), these licensing restrictions also apply to VISOR.

EPIC-KITCHENS-100 is accessible via \href{https://data.bris.ac.uk/data/dataset/2g1n6qdydwa9u22shpxqzp0t8m}{this data.bris} link.}


\dsqc{Does the dataset contain data that might be considered confidential (e.g., data that is
protected by legal privilege or by doctor–patient confidentiality, data that includes the
content of individuals’ non-public communications)?}{If so, please provide a description.}

\dsa{No. VISOR only contains segments and relations between them.}

\dsak{We do not believe that EPIC-KITCHENS-100 contains confidential information. Participants reviewed their footage before release. It is stored in a GDPR-compliant server.}

\dsqc{Does the dataset contain data that, if viewed directly, might be offensive, insulting,
threatening, or might otherwise cause anxiety?}{If so, please describe why}

\dsa{No. VISOR only contains segments.}

\dsak{We do not believe so. The data shows samples of cooking and other daily kitchen activities.} 

\dsqc{Does the dataset relate to people?}{If not, you may skip the remaining questions in this section.}

\dsa{The VISOR dataset relates to people since it consists of annotations for egocentric data collected by people doing daily activities in the kitchen. The footage and annotations are otherwise anonmous.}

\dsqc{Does the dataset identify any subpopulations (e.g., by age, gender)?}{If so, please describe how
these subpopulations are identified and provide a description of their respective distributions within the dataset.}

\dsa{No. The participants from the base EPIC-KITCHENS dataset are anonymous.}

\dsqc{Is it possible to identify individuals (i.e., one or more natural persons), either directly or
indirectly (i.e., in combination with other data) from the dataset?}{If so, please describe how}

\dsa{VISOR contains only segments. We do not believe this is possible from VISOR data alone}

\dsak{It is possible not possible to identify individuals in the dataset. The data has been stripped of information that would make this easy.  The consent forms linking participant IDs to their identities are not public and stored securely at the University of Bristol.}

\dsqc{Does the dataset contain data that might be considered sensitive in any way (e.g., data that
reveals racial or ethnic origins, sexual orientations, religious beliefs, political opinions or union
memberships, or locations; financial or health data; biometric or genetic data; forms of
government identification, such as social security numbers; criminal history)?}{If so, please
provide a description.}

\dsa{VISOR contains only segments. We do not believe this is possible from VISOR data alone.}

\dsak{The EPIC-KITCHENS data may reveal information about racial or ethic origin, sex, and location due to the participants visible hands and kitchen contents. We do not believe that the EPIC-KITCHENS dataset contains sensitive data. Two factors also make this less likely: participants in the dataset are anonymous, and participants collected the footage themselves and reviewed it before its inclusion in the dataset.}

\subsection*{Collection Process}

\dsqc{How was the data associated with each instance acquired?}{Was the data directly observable
(e.g., raw text, movie ratings), reported by subjects (e.g., survey responses), or indirectly
inferred/derived from other data (e.g., part-of-speech tags, model-based guesses for age or
language)? If data was reported by subjects or indirectly inferred/derived from other data, was
the data validated/verified? If so, please describe how.}

\dsa{A full description appears in the associated paper and its appendix. However, briefly:}
\begin{enumerate}
    \item {\it Frame and Entity Identification.} Frames and entities in the each frame to be labelled were identified via a mix of rules, crowdsoucing, and student work.
    \item {\it Pixel Labelling.} A freelancer annotator segmented the entity in each frame; each video was annotated by a single annotator who had the ability to move back and forth through time.
    \item {\it Correction of Pixel Labelling.} These segments were checked for consistency by researchers in our lab.
    \item {\it Extra annotation (Exhaustive Labels, Hand-Object Relations).} The segments were annotated with extra information (e.g., exhaustive annotations, hand-object relations) by a crowdsourcing company.
\end{enumerate}
Additionally, the VISOR dense annotations were created by a deep learning model.

\dsqc{What mechanisms or procedures were used to collect the data (e.g., hardware apparatus or
sensor, manual human curation, software program, software API)?}{How were these
mechanisms or procedures validated?}

\dsa{A full description appears in the associated paper and its appendix. However, briefly:}

\begin{enumerate}
    \item {\it Frame and Entity Identification.} This stage was done by a mix of the Amazon Mechanical Turk platform and a custom interface created for this project.
    \item {\it Pixel Labelling.} This stage was done via the TOronto Annotation Suite (TORAS). 
    \item {\it Correction of Pixel Labelling.} This stage was done via a custom interface created for this project.
    \item {\it Extra annotation (Exhaustive Labels, Hand-Object Relations).} This stage was done via The Hive's platform.
\end{enumerate}

\dsq{If the dataset is a sample from a larger set, what was the sampling strategy (e.g.,
deterministic, probabilistic with specific sampling probabilities)?}

\dsa{There is not a larger set of labels from which VISOR is subsampled. However, VISOR is annotated on a subset of the frames of the EPIC-KITCHENS-100 dataset. This subset was selected from with the goal of avoiding motion blur in the annotated frames.
}

\dsq{Who was involved in the data collection process (e.g., students, crowdworkers, contractors)
and how were they compensated (e.g., how much were crowdworkers paid)?}

\dsa{The annotation process for VISOR consists of multiple stages.
\begin{enumerate}
\item {\it Frame and Entity Identification.} Crowdworkers from Amazon Mechanical Turk performed this task with graduate students involved in the project providing quality control. 
We paid \$11.25 per thousand actions annotated.
We provide these as HITs of 16 consecutive actions each.
Graduate students did this as part of their normal responsibilities on the project.
\item {\it Pixel Labelling.} Contractors from Upwork performed the pixel annotation. Contractors were paid \$6-9 an hour based on experience with higher rates reserved for experienced annotators who also performed quality assurance checks. 
\item {\it Correction of Pixel Labelling.} Researchers involved in the project performed this task alongside other members from the Machine Learning and Computer Vision group at the University of Bristol. These volunteered for the task and are acknowledged in the paper.
\item {\it Extra annotation (Exhaustive Labels, Hand-Object Relations).} We paid Hive (thehive.ai) to obtain annotations for these annotations. We paid Hive \$20 per thousand tasks annotated. 
\end{enumerate}
}

\dsqc{Over what timeframe was the data collected?}{Does this timeframe match the creation
timeframe of the data associated with the instances (e.g., recent crawl of old news articles)? If
not, please describe the timeframe in which the data associated with the instances was created.}

\dsa{The annotations were collected over the a period of 22 months with the bulk of the collection done in the year from June 2021 -- June 2022.}

\dsak{The collection time of the underlying EPIC-KITCHENS data and annotations spanned Apr 2017-July 2020.}

\dsqc{Were any ethical review processes conducted (e.g., by an institutional review board)?}{If so,
please provide a description of these review processes, including the outcomes, as well as a link
or other access point to any supporting documentation}

\dsa{VISOR is new annotations, rather than new data.}

\dsak{EPIC-KITCHENS was collected with University of Bristol faculty ethics approval. These application is held at the university of Bristol. Participant consent form is \href{https://data.bris.ac.uk/data/dataset/7f36006fae1331bc247b8c3522a13083}{available here}.
} 

\dsqc{Does the dataset relate to people?}{If not, you may skip the remaining questions in this section.}

\dsa{Yes. VISOR consists of annotations for videos showing people performing daily activities in their kitchen.} 

\dsq{Did you collect the data from the individuals in question directly, or obtain it via third parties
or other sources (e.g., websites)}

\dsa{Not applicable to VISOR.}

\dsak{Yes. This data was collected directly by and with individuals in question.}

\dsqc{Were the individuals in question notified about the data collection?}{If so, please describe (or
show with screenshots or other information) how notice was provided, and provide a link or
other access point to, or otherwise reproduce, the exact language of the notification itself.}

\dsa{Not applicable to VISOR.}

\dsak{Yes. Since the data was directly collected by the participants, the participants were aware of the data collection process. All participants were given the opportunity to ask questions before participating, and they could withdraw at any time without giving a reason. Participants consented to the process and watched their footage. All participants were volunteers and were not compensated.}

\dsqc{Did the individuals in question consent to the collection and use of their data?}{If so, please
describe (or show with screenshots or other information) how consent was requested and
provided, and provide a link or other access point to, or otherwise reproduce, the exact
language to which the individuals consented.}

\dsa{Not applicable to VISOR.}

\dsak{Yes. The participants consented to data the collection and use of their data. In particular, they reviewed their footage before its use in the dataset.}

\dsqc{If consent was obtained, were the consenting individuals provided with a mechanism to
revoke their consent in the future or for certain uses?}{If so, please provide a description, as
well as a link or other access point to the mechanism (if appropriate).}

\dsa{Not applicable to VISOR.}

\dsak{Participants were able to withdraw from the process at any point until the data was published by DOI. At the moment, participants are unable to withdraw their data.}

\dsqc{Has an analysis of the potential impact of the dataset and its use on data subjects (e.g., a data
protection impact analysis) been conducted?}{If so, please provide a description of this analysis,
including the outcomes, as well as a link or other access point to any supporting documentation.}

\dsak{The university of Bristol faculty ethics committee have reviewed the protocol, and approved the dataset. They checked any potential impact and as the data is anonymous no further actions were deemed as needed.}

\dsq{Any other comments?}

\dsa{No}

\subsection*{Preprocessing/cleaning/labeling}

\dsqc{Was any preprocessing/cleaning/labeling of the data done (e.g., discretization or bucketing,
tokenization, part-of-speech tagging, SIFT feature extraction, removal of instances, processing
of missing values)?}{If so, please provide a description. If not, you may skip the remainder of the
questions in this section.}

\dsa{The data consists of labels, so naturally labelling was done. There were multiple stages of cleaning of the data. This cleaning, however, aimed to fix inconsistencies in labelling (e.g., correcting typos).}

\dsqc{Was the ``raw'' data saved in addition to the preprocessed/cleaned/labeled data (e.g., to
support unanticipated future uses)?}{If so, please provide a link or other access point to the
“raw” data.}

\dsa{There is no raw data besides incorrect earlier versions of the data, such as annotations before typographic errors were fixed.}

\dsqc{Is the software used to preprocess/clean/label the instances available?}{If so, please provide a
link or other access point.}

\dsa{Primarily no. Some of the software is proprietary (e.g., the TORAS labelling software); some of the software is one-off script that are not of interest due their simplicity and non-general purpose nature.}

\dsq{Any other comments?}

\dsa{No}

\subsection*{Uses}

\dsqc{Has the dataset been used for any tasks already?}{If so, please provide a description.}

\dsa{We use VISOR in conjunction with EPIC-KITCHENS-100 to solve three challenges: (a) video object segmentation; (b) hand-object segmentation; (c) and a {\it Where Did This Come From?} challenge. These are challenges are documented in our paper.}

\dsqc{Is there a repository that links to any or all papers or systems that use the dataset?}{If so,
please provide a link or other access point}

\dsa{No}

\dsq{What (other) tasks could the dataset be used for?}

\dsa{We anticipate that the data will be useful for many different long-term pixel-grounded video understanding tasks.}

\dsqc{Is there anything about the composition of the dataset or the way it was collected and
preprocessed/cleaned/labeled that might impact future uses?}{For example, is there anything
that a future user might need to know to avoid uses that could result in unfair treatment of
individuals or groups (e.g., stereotyping, quality of service issues) or other undesirable harms
(e.g., financial harms, legal risks) If so, please provide a description. Is there anything a future
user could do to mitigate these undesirable harms?}

\dsa{While the EPIC-KITCHENS videos were collected in 4 countries by participants from 10 nationalities, it is in no way representative of all kitchen-based activities globally, or even within the recorded countries.
 Models trained on this dataset are thus expected to be exploratory, for research and investigation purposes. Anticipating unintended consequences of data is difficult. Here are some potential issues that we see in the data. \begin{enumerate}
\item The frames that are annotated are selected to be easy to annotate, and therefore may have little motion blur. Models may require motion blur augmentation in order to generalise to all frames.
\item Due to the collection of data collection process, the data shows fewer unique individuals compared to e.g., Internet data. This may make it harder to generalise.
\item The verb and noun classes in no way cover all actions and objects present, even in the kitchens recorded.
\item The dataset is naturally-long tailed. Accordingly the models will be biased to better recognise head classes.
\end{enumerate}

}

\dsqc{Are there tasks for which the dataset should not be used?}{If so, please provide a description.}

\dsa{VISOR is available for non-commercial research purposes only. Accordingly, it should not be used for commercial purposes. A commercial license can be acquired through negotiation with the University of Bristol.}

\dsq{Any other comments?}

\dsa{No}

\subsection*{Distribution}

\dsqc{Will the dataset be distributed to third parties outside of the entity (e.g., company, institution,
organization) on behalf of which the dataset was created?}{If so, please provide a description.}

\dsa{Yes. The VISOR will be publicly available for non-commercial research purposes, just like its base data, EPIC-KITCHENS-100.}

\dsqc{How will the dataset will be distributed (e.g., tarball on website, API, GitHub)?}{Does the
dataset have a digital object identifier (DOI)?}

\dsa{The dataset will be released via the University of Bristol data.bris data repository. This repository  assigns unique DOIs upon deposit.}

\dsq{When will the dataset be distributed?}

\dsa{The dataset will be publicly released on (or before) 1 August 2022.}

\dsqc{Will the dataset be distributed under a copyright or other intellectual property (IP) license,
and/or under applicable terms of use (ToU)?}{If so, please describe this license and/or ToU, and
provide a link or other access point to, or otherwise reproduce, any relevant licensing terms or
ToU, as well as any fees associated with these restrictions.}

\dsa{VISOR will be released under a \href{https://creativecommons.org/licenses/by-nc/4.0/}{Creative Commons BY-NC 4.0} license, which restricts commercial use of the data.}

\dsqc{Have any third parties imposed IP-based or other restrictions on the data associated with the
instances?}{If so, please describe these restrictions, and provide a link or other access point to, or
otherwise reproduce, any relevant licensing terms, as well as any fees associated with these
restrictions. }

\dsa{No third parties have imposed restrictions on VISOR.}

\dsqc{Do any export controls or other regulatory restrictions apply to the dataset or to individual
instances?}{If so, please describe these restrictions, and provide a link or other access point to, or
otherwise reproduce, any supporting documentation. }

\dsa{No. There are no restrictions beyond following the non-commercial license.}

\dsq{Any other comments?}

\dsa{No}

\subsection*{Maintenance}

\dsq{Who will be supporting/hosting/maintaining the dataset?}

\dsa{The dataset will be released via the University of Bristol data.bris data repository. This enables long-term preservation of the data even if the PIs change institutions.}

\dsq{How can the owner/curator/manager of the dataset be contacted (e.g., email address)?}

\dsa{The creators of the dataset are listed in this document and can be contacted via email.} 

\dsqc{Is there an erratum?}{If so, please provide a link or other access point.}

\dsa{Not at the time of release. If there are errata or updates, we will provide them on the dataset website.}

\dsqc{Will the dataset be updated (e.g., to correct labeling errors, add new instances, delete
instances)?}{If so, please describe how often, by whom, and how updates will be communicated
to users (e.g., mailing list, GitHub)?}

\dsa{We do not have concrete plans as of yet; we will announce any updates on the dataset website.}

\dsqc{If the dataset relates to people, are there applicable limits on the retention of the data
associated with the instances (e.g., were individuals in question told that their data would be
retained for a fixed period of time and then deleted)?}{If so, please describe these limits and
explain how they will be enforced}

\dsa{This does not apply to VISOR.}

\dsak{There are no limits on the retention of data.}

\dsqc{Will older versions of the dataset continue to be supported/hosted/maintained?}{If so, please
describe how. If not, please describe how its obsolescence will be communicated to users.}

\dsa{Yes. The version released is fixed by the DOI \url{https://doi.org/10.5523/bris.2v6cgv1x04ol22qp9rm9x2j6a7} and will not be changed. The university of Bristol is committed to storage and maintenance of the dataset for 20 years.}

\dsqc{If others want to extend/augment/build on/contribute to the dataset, is there a mechanism
for them to do so?}{If so, please provide a description. Will these contributions be
validated/verified? If so, please describe how. If not, why not? Is there a process for
communicating/distributing these contributions to other users? If so, please provide a
description.}

\dsa{Users are free to extend the dataset on their own and create derivative works, so long as they follow the license agreement. This was also the case for EPIC-KITCHENS-100. There is, however, no official mechanism to integrate user contributions into a new version of the dataset.}

\dsq{Any other comments?}

\dsa{No}

\end{document}